\documentclass[openany]{now} 

\usepackage{soul}
\usepackage{etoolbox}
\usepackage{amssymb,amsmath} 
\usepackage{dsfont}
\usepackage{appendix}
\usepackage{natbib}  
\usepackage{mathrsfs}  
\usepackage{breakcites}     
\usepackage{tikz}    
\usepackage{pgf}  
\usepackage{caption}
\usepackage{subcaption}   
\usepackage{pgfplots} 
\usepackage{pgffor}     
\usepackage{bm}
\usepackage{xcolor} 
\usepackage{pdflscape}  
\usepackage[chapter]{algorithm}  
\usepackage{algorithmic} 
\usepackage{adjustbox}
\usepackage{wasysym} 
\usepackage{pifont}     
\usepackage{enumitem}  
\usepackage{glossaries} 
\usepackage{booktabs} 
\usepackage{longtable} 
\usepackage[framemethod=TikZ]{mdframed} 
\newcommand{\cmark}{\ding{51}}%
\newcommand{\xmark}{\ding{55}}%
 
\usepackage{framed}  
  
\mdfsetup{%
  middlelinewidth = 1pt 
}  
    
\allowdisplaybreaks  
   
\usepgfmodule{shapes}
\usepgfmodule{plot}
\usetikzlibrary{decorations}
\usetikzlibrary{arrows}
\usetikzlibrary{shadows}  
\usetikzlibrary{snakes} 
\usetikzlibrary{shapes}
\usetikzlibrary{matrix} 
\usetikzlibrary{positioning} 
\usetikzlibrary{backgrounds}
 
\hypersetup{
  colorlinks,
  linkcolor=black,
  citecolor=blue!50!black 
} 
 
\newmdenv[roundcorner=6pt,shadow=true]{shadowbox}

\robustify{\gls} 
\newacronym[plural={RKHSes}]{rkhs}{RKHS}{reproducing kernel {H}ilbert space}
 
\newcommand{\ie}{\emph{i.e.}}
\newcommand{\eg}{\emph{e.g.}}

\newcommand{\wrt}{w.r.t.~}
 
\newcommand{\fspace}{\ensuremath{\mathcal{F}}}    
\newcommand{\inspace}{\ensuremath{\mathcal{X}}}   
\newcommand{\inx}{\ensuremath{\mathcal{X}}}
\newcommand{\iny}{\ensuremath{\mathcal{Y}}}
\newcommand{\inz}{\ensuremath{\mathcal{Z}}}
\newcommand{\outspace}{\ensuremath{\mathcal{Y}}}  
\newcommand{\pp}[1]{\ensuremath{\mathbb{#1}}}     
\newcommand{\pspace}{\ensuremath{M_+^1}}   

\newcommand{\hbspace}{\ensuremath{\mathscr{H}}}   
\newcommand{\hbspg}{\ensuremath{\mathscr{G}}} 
\newcommand{\hbspf}{\ensuremath{\mathscr{F}}}

\newcommand{\muv}{\ensuremath{\mu}}
\newcommand{\muh}{\ensuremath{\hat{\muv}}}
\newcommand{\borelms}{\mathrm{\Lambda}}           
\newcommand{\cf}[1]{\varphi_{#1}}                 
\newcommand{\rd}[1]{\mathbb{R}^{#1}}              
\newcommand{\rr}{\mathbb{R}} 		          
\newcommand{\cc}{\mathbb{C}}                   
\newcommand{\ep}{\mathbb{E}}                      
\newcommand{\kmat}{\mathbf{K}}                  
\newcommand{\lmat}{\mathbf{L}}                  
\newcommand{\hmat}{\mathbf{H}}                  
\newcommand{\avec}{\bm{\alpha}}                  
\newcommand{\bvec}{\bm{\beta}}                  
\newcommand{\id}{\mathbf{I}}
\newcommand{\covx}{\ensuremath{\mathcal{C}_{\mathit{XX}}}} 
\newcommand{\covy}{\ensuremath{\mathcal{C}_{\mathit{YY}}}}
\newcommand{\covxy}{\ensuremath{\mathcal{C}_{\mathit{XY}}}}
\newcommand{\covyx}{\ensuremath{\mathcal{C}_{\mathit{YX}}}}
\newcommand{\ecovx}{\ensuremath{\widehat{\mathcal{C}}_{\mathit{XX}}}}
\newcommand{\ecovy}{\ensuremath{\widehat{\mathcal{C}}_{\mathit{YY}}}}
\newcommand{\ecovxy}{\ensuremath{\widehat{\mathcal{C}}_{\mathit{XY}}}}
\newcommand{\ecovyx}{\ensuremath{\widehat{\mathcal{C}}_{\mathit{YX}}}}
\newcommand{\dd}{\, \mathrm{d}}

\definecolor{krikcolor}{rgb}{0.2, 0.6, 0.8}

\newcommand{\m}{\ensuremath{\mathbf{m}}}
\newcommand{\w}{\ensuremath{\mathbf{w}}}
\newcommand{\x}{\ensuremath{\mathbf{x}}}
\newcommand{\y}{\ensuremath{\mathbf{y}}}
\newcommand{\z}{\ensuremath{\mathbf{z}}}  
\newcommand{\X}{\ensuremath{\mathbf{X}}} 
\newcommand{\Y}{\ensuremath{\mathbf{Y}}}

\def\ci{\perp\!\!\!\perp} 
 
\bibpunct[; ]{(}{)}{,}{a}{}{;}

\newcommand{\hiddensection}[1]{
    \stepcounter{section}
    \section*{\arabic{chapter}.\arabic{section}\hspace{1em}{#1}}
}

\title{Kernel Mean Embedding of Distributions: \\ A Review and Beyond}
  
\author{
Krikamol Muandet \\
Department of Mathematics \\
Faculty of Science, Mahidol University\\ 
272 Rama VI Road, Ratchathewi, Bangkok 10400, Thailand \\
\and
Empirical Inference Department \\ 
Max Planck Institute for Intelligent Systems \\
Spemannstra\ss e 38, T\"ubingen 72076, Germany \\
\href{mailto:krikamol.mua@mahidol.ac.th}{krikamol.mua@mahidol.ac.th}
\and
Kenji Fukumizu \\ 
Institute of Statistical Mathematics \\
10-3 Midoricho, Tachikawa, Tokyo 190-8562 Japan \\ 
\href{mailto:fukumizu@ism.ac.jp}{fukumizu@ism.ac.jp}
\and
Bharath Sriperumbudur \\  
Department of Statistics, Pennsylvania State University \\
University Park, PA 16802, USA \\  
\href{mailto:bks18@psu.edu}{bks18@psu.edu} 
\and
Bernhard Sch\"olkopf \\ 
Max Planck Institute for Intelligent Systems \\
Spemannstra\ss e 38, T\"ubingen 72076, Germany \\
\href{mailto:bs@tuebingen.mpg.de}{bs@tuebingen.mpg.de}
}

\begin{document} 



\frontmatter  

\maketitle
  
\tableofcontents   
   
\mainmatter   

\begin{abstract}
 
  A Hilbert space embedding of a distribution---in short, a kernel mean embedding---has recently emerged as a
  powerful tool for machine learning and statistical inference. The basic idea
  behind this framework is to map distributions into a reproducing
  kernel Hilbert space (RKHS) in which the whole arsenal of kernel
  methods can be extended to probability measures. It can be viewed as a generalization of the original ``feature map'' common to support vector machines (SVMs) and other kernel methods. 
In addition to the classical applications of kernel methods, the kernel mean embedding has found novel applications in fields ranging from probabilistic modeling to statistical
  inference, causal discovery, and deep learning.
 
The goal of this survey is to give a comprehensive
  review of existing work and recent advances in this research area,
  and to discuss some of the most challenging issues and open problems
  that could potentially lead to new research directions. The survey
  begins with a brief introduction to the RKHS and positive definite
  kernels which forms the backbone of this survey, followed by a
  thorough discussion of the Hilbert space embedding of marginal
  distributions, theoretical guarantees, and a review of its
  applications. The embedding of distributions enables us to apply 
  RKHS methods to probability measures which prompts a wide range of
  applications such as kernel two-sample testing, independent testing,
  group anomaly detection, and learning on distributional data.  Next,
  we discuss the Hilbert space embedding for conditional
  distributions, give theoretical insights, and review some
  applications. The conditional mean embedding enables us to perform
  sum, product, and Bayes' rules---which are ubiquitous in graphical
  model, probabilistic inference, and reinforcement learning---in a
  non-parametric way using the new representation of distributions in
  RKHS. We then discuss relationships between this framework and other related areas. Lastly, we give some suggestions on future research directions. The targeted audience includes graduate students and researchers in machine learning and statistics who are interested in the theory and applications of kernel mean embeddings.
 
\end{abstract} 


\chapter{Introduction} 

This work aims to provide a comprehensive review 
of kernel mean embeddings of distributions and, in the course of doing so, discusses some challenging issues that could potentially lead to new research directions. 
To the best of our knowledge, there is no comparable review in this area so far; however, the short review paper of \citet{Song2013} on Hilbert space 
embedding of conditional distributions and its applications in nonparametric inference in graphical models may be of interest to some readers.

The kernel mean embedding owes its success to a positive definite function commonly known as the \emph{kernel function}. The kernel function has become popular in the machine learning community for more than 20 years. Initially, it arises as an effortless way to perform an inner product $\langle \x,\y \rangle$ in a 
high-dimensional feature space $\hbspace$ for some data points $\x,\y\in\inspace$. The positive definiteness of the kernel function guarantees the existence of a dot product space $\hbspace$ and a mapping $\phi :\inspace\rightarrow\hbspace$ such that 
$k(\x,\y) = \langle\phi(\x),\phi(\y)\rangle_{\hbspace}$ \citep{aronszajn50reproducing} without needing to compute $\phi$ explicitly
\citep{BosGuyVap92,Cortes1995,Vapnik00,Scholkopf01:LKS}. The kernel function can be applied to any learning algorithm as long as the latter can be expressed entirely in terms of a dot product $\langle \x,\y \rangle$ \citep{Scholkopf98:KPCA}. This trick is commonly known 
as the \emph{kernel trick} (see Section \ref{sec:background} for a more detailed account). Many kernel functions have been proposed for various kinds of data structures including non-vectorial data such as graphs, text documents, semi-groups, and probability distributions \citep{Scholkopf01:LKS,Gartner2003}. Many well-known learning algorithms have already been \emph{kernelized} and have proven successful in scientific disciplines such as bioinformatics, natural language processing, computer vision, robotics, and causal inference.


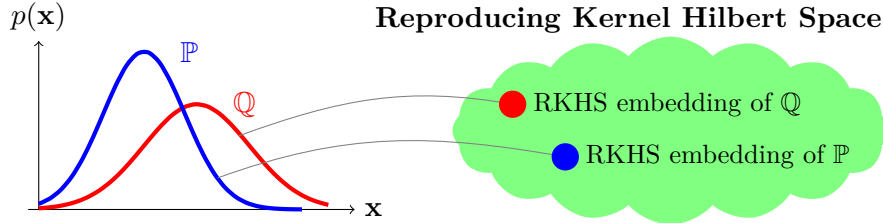
\begin{figure}[t!]
  \centering
  \begin{tikzpicture}[domain=0:6,scale=0.7]
      \coordinate (P1) at (3.4,1.6);
      \coordinate (Q1) at (3.8,2.4);
      \coordinate (P2) at (10,2);
      \coordinate (Q2) at (9,3);

      \draw[->] (-0.2,1) -- (6,1) node[right] {$\x$};
      \draw[->] (0,1) -- (0,4.2) node[above] {$p(\x)$};
      \draw[color=red,smooth,domain=0:5.5,ultra thick] plot (\x,{1+2*exp(-(\x-3)*(\x-3)*0.5/1)});
      \draw[color=blue,smooth,domain=0:5,ultra thick] plot (\x,{1+3*exp(-(\x-2)*(\x-2)*0.5/0.6)});

      \node [cloud, fill=green!50, cloud puffs=16, cloud puff arc= 100,
      minimum width=5.8cm, minimum height=2.5cm, aspect=1] at (12,2.5) {};
      \node [] at (11.2,4.6) {\textbf{Reproducing Kernel Hilbert Space}};

      \draw[color=gray] (P1) to[out=20,in=170] (P2);
      \draw[color=gray] (Q1) to[out=20,in=170] (Q2);

      \node [fill=blue,circle,thick,minimum width=0.1cm] at (P2) {};
      \node [right] at (10.2,2) {{\small RKHS embedding of $\pp{P}$}};
      \node [fill=red,circle,thick,,minimum width=0.1cm] at (Q2) {};
      \node [right] at (9.2,3) {{\small RKHS embedding of $\pp{Q}$}};

      \node [right] at (2.5,4) {$\color{blue} \pp{P}$};
      \node [right] at (3.5,3) {$\color{red} \pp{Q}$};
    \end{tikzpicture}
  \caption{Embedding of marginal distributions: each distribution is mapped into a reproducing kernel Hilbert space via an expectation operation.}
  \label{fig:marginal-embedding}
\end{figure}

Figures \ref{fig:marginal-embedding} and \ref{fig:conditional-embedding-summary} depict schematic illustrations of the kernel mean embedding framework. In words, the idea of 
\emph{kernel mean embedding} is to extend the feature map $\phi$ to the space of probability distributions by representing each distribution $\pp{P}$ as a mean function
\begin{equation}
  \label{eq:mean-representation}
  \phi(\pp{P})=\mu_{\pp{P}} := \int_{\mathcal{X}} k(\x,\cdot) \dd\pp{P}(\x), 
\end{equation}
\noindent where $k:\mathcal{X}\times\mathcal{X}\rightarrow\mathbb{R}$ is a symmetric and positive definite kernel function \citep{Berlinet04:RKHS,Smola07Hilbert}. Since $k(\x,\cdot)$ takes values in the feature space $\hbspace$, the integral in \eqref{eq:mean-representation} should be interpreted as 
a Bochner integral (see, \eg, \citealp[Chapter 2]{Diestel-77} and \citealp[Chapter 1]{Dinculeanu:2000} for a definition of the Bochner integral). 
Conditions ensuring the existence of such an integral will be discussed in Section \ref{sec:marginal-embedding}, but in this case we essentially transform the distribution $\pp{P}$ to an element in the feature space $\hbspace$, which is nothing but a reproducing kernel Hilbert space (RKHS) corresponding to the kernel $k$.
Through \eqref{eq:mean-representation}, most RKHS methods can therefore be extended to probability measures. There are several reasons why this representation may be beneficial.

First of all, for a class of kernel functions known as \emph{characteristic kernels}, the kernel mean representation captures all information about the distribution $\pp{P}$ \citep{Fukumizu04:DRS,Sriperumbudur08injectivehilbert,Sriperumbudur10:Metrics}. In other words, 
the mean map $\pp{P} \mapsto \mu_{\pp{P}}$ is injective, implying that $\|\mu_{\pp{P}} - \mu_{\pp{Q}}\|_{\hbspace} = 0$ if and only if $\pp{P}=\pp{Q}$, \ie, $\pp{P}$ and $\pp{Q}$ are the same distribution. Consequently, the kernel mean representation can be used to define a metric over the space of probability distributions \citep{Sriperumbudur10:Metrics}. Since $\|\mu_{\pp{P}} - \mu_{\pp{Q}}\|_{\hbspace}$ can be bounded from above by some popular probability metrics such as the Wasserstein distance and the total variation distance, it follows that if $\pp{P}$ and $\pp{Q}$ are close in these distances, then $\muv_{\pp{P}}$ is also close to $\muv_{\pp{Q}}$ in the $\|\cdot\|_{\hbspace}$ norm (see \S\ref{sec:MMD}).
Injectivity of $\pp{P} \mapsto \mu_{\pp{P}}$ makes it suitable for applications that require a unique characterization of distributions such as two-sample homogeneity tests \citep{Gretton12:KTT,Fukumizu2008,Zhang2011,Doran2014}. Moreover, using the kernel mean representation, most learning algorithms can be extended to the space of probability distributions with minimal assumptions on the underlying data generating process \citep{Gomez-Chova2010,Muandet12:SMM,Guevara2014,Lopez-Paz15:Towards}. See \S\ref{sec:embedding-theory} for details.

Secondly, several elementary operations on distributions (and associated random variables) can be performed directly by means of this representation. For example, by the reproducing property of $\hbspace$, we have 
\begin{align*}
  \mathbb{E}_{\pp{P}}[f(\x)] &= \langle f,\mu_{\pp{P}}\rangle_{\hbspace}, \quad \forall f\in\hbspace .
  \intertext{That is, an expected value of any function $f \in\hbspace$ w.r.t. $\pp{P}$ is nothing but an inner product in $\hbspace$ between $f$ and $\mu_{\pp{P}}$. Likewise, for an RKHS $\hbspg$ over some input space $\outspace$, we have}
  \mathbb{E}_{Y|\x}[g(Y)\,|\,X=\x] &= \langle g, \mathcal{U}_{Y|\x}\rangle_{\hbspg}, \quad \forall g\in\hbspg ,
\end{align*}
\noindent where $\mathcal{U}_{Y|\x}$ denotes the embedding of the conditional distribution $\pp{P}(Y|X=\x)$. That is, we can compute a conditional expected value of any function $g\in\hbspg$ w.r.t. $\pp{P}(Y|X=\x)$ by taking an inner product in $\hbspg$ between the function $g$ and the embedding of $\pp{P}(Y|X=\x)$ (see Section \ref{sec:conditional-embedding} for further details). These operations only require knowledge of the empirical estimates of $\mu_{\pp{P}}$ and $\mathcal{U}_{Y|\x}$. Hence, the kernel mean representation allows us to implement these operations in \emph{non-parametric} probabilistic inference, \eg, filtering for dynamical systems \citep{Song10:KCOND}, kernel belief propagation \citep{Song11:KBP}, kernel Monte Carlo filter \citep{Kanagawa2013}, kernel Bayes' rule \citep{Fukumizu11:KBR}, often with strong theoretical guarantees. Moreover, it can be used to perform functional operations $f(X,Y)$ on random variables $X$ and $Y$ \citep{Schoelkopf15:KPP,Carl-johann16:ConstKME}.

\begin{figure}[t!]
  \begin{tikzpicture}[scale=0.7]
      \draw[->] (-0.2,0.5) -- (7.5,0.5) node[right] {$X$};
      \draw[->] (0,0.5) -- (0,3.7) node[left,yshift=-10] {$Y$};
      \draw[rotate=-90,color=blue!10,smooth,domain=-3.5:-0.5,ultra thick] plot (\x,{1+exp(-(3+\x)*(3+\x)*0.5/0.2)});
      \draw[rotate=-90,color=blue!15,smooth,domain=-3.5:-0.5,ultra thick] plot (\x,{1.6+exp(-(2.8+\x)*(2.8+\x)*0.5/0.3)});
      \draw[rotate=-90,color=blue!20,smooth,domain=-3.5:-0.5,ultra thick] plot (\x,{2.2+exp(-(2.6+\x)*(2.6+\x)*0.5/0.3)});
      \draw[rotate=-90,color=blue!25,smooth,domain=-3.5:-0.5,ultra thick] plot (\x,{2.8+exp(-(2.4+\x)*(2.4+\x)*0.5/0.3)});
      \draw[rotate=-90,color=blue!30,smooth,domain=-3.5:-0.5,ultra thick] plot (\x,{3.4+exp(-(2.2+\x)*(2.2+\x)*0.5/0.3)});
      \draw[rotate=-90,color=blue!35,smooth,domain=-3.5:-0.5,ultra thick] plot (\x,{4+exp(-(2.4+\x)*(2.4+\x)*0.5/0.3)});
      \draw[rotate=-90,color=blue!40,smooth,domain=-3.5:-0.5,ultra thick] plot (\x,{4.6+exp(-(2.5+\x)*(2.5+\x)*0.5/0.5)});
      \draw[rotate=-90,color=blue!45,smooth,domain=-3.5:-0.5,ultra thick] plot (\x,{5.2+exp(-(2.6+\x)*(2.6+\x)*0.5/0.5)});
      \draw[rotate=-90,color=blue!50,smooth,domain=-3.5:-0.5,ultra thick] plot (\x,{5.8+exp(-(2.9+\x)*(2.9+\x)*0.5/0.6)});
      \node [rotate=-90] at (0.6,2.6) {$\pp{P}(Y|X)$};


      \coordinate (P1) at (-2.3,5.7);
      \coordinate (P2) at (9,5.7);



      \node [cloud, fill=green!50, cloud puffs=16, cloud puff arc= 100,
      minimum width=3.4cm, minimum height=2.2cm, aspect=1] at (-2.5,5) {};
      \node [] at (-2.5,7) {\textbf{RKHS} $\hbspace$};

      \node [cloud, fill=green!50, cloud puffs=16, cloud puff arc= 100,
      minimum width=3.4cm, minimum height=2.2cm, aspect=1] at (9.5,5) {};
      \node [] at (9.5,7) {\textbf{RKHS} $\hbspg$};

      \draw[->,color=gray,dashed,very thick] (P1) to[out=45,in=135] (P2);

      \node [fill=blue,circle,minimum width=0.05cm] at (P2) {};
      \node [below,yshift=-0.15cm,xshift=0.6cm, text width=3cm] at (P2) {{\small RKHS embedding of $\pp{P}(Y|X=\x)$}};
      \node [fill=red,circle,minimum width=0.05cm] at (P1) {};
      \node [below,yshift=-0.15cm] at (P1) {{\small feature map of $\x$}};

      \coordinate (n1) at (0,5);
      \coordinate (n2) at (7,5);
      \path (n1) -- node[sloped, fill=yellow, rounded corners=.3cm, text width=3cm] (text) {{\small RKHS embedding of $\pp{P}(Y|X)$}} (n2);
      \draw[->, very thick] (n1)--(text)--(n2) node[right] {};


      \draw[->] (-0.2,9) -- (4,9) node[right] {$\y$};
      \draw[->] (0,9) -- (0,11.2) node[above] {$p(\y|\x)$};
      \draw[color=red,smooth,domain=0:4,ultra thick] plot (\x,{9.2+1.8*exp(-(\x-2)*(\x-2)*0.5/0.3)});
      \node [right] at (3,10) {$\pp{P}(Y|X=\x)$};
    \end{tikzpicture}
  \caption{From marginal distribution to conditional distribution: unlike the embeddings shown in Figure \ref{fig:marginal-embedding}, the embedding of 
  conditional distribution $\pp{P}(Y|X)$ is not a single element in the RKHS. Instead, it may be viewed as a family of Hilbert space embeddings of conditional distributions $\pp{P}(Y|X=\x)$ indexed by the conditioning variable $X$. In other words, the conditional mean embedding can be viewed as an operator from $\hbspace$ to $\hbspg$(cf. \S\ref{sec:regression-view}).}
  \label{fig:conditional-embedding-summary}
\end{figure}


In some applications such as testing for homogeneity from finite samples, representing the distribution $\pp{P}$ by $\mu_{\pp{P}}$ bypasses an intermediate density estimation, which is known to be difficult in the high-dimensional setting \citep[Section 6.5]{Wasserman2006}. Moreover, we can extend the applications of kernel mean embedding straightforwardly to non-vectorial data such as graphs, strings, and semi-groups, thanks to the kernel function. As a result, statistical inference---such as two-sample testing and independence testing---can be adapted directly to distributions over complex objects \citep{Gretton12:KTT}.

Under additional assumptions, we can generalize the principle underlying \eqref{eq:mean-representation} to conditional distributions $\pp{P}(Y|X)$ and $\pp{P}(Y|X=\x)$. Essentially, the latter two objects are represented as an operator that maps the feature space $\hbspace$ to $\hbspg$, and as an object in the feature space $\hbspg$, respectively, where $\hbspace$ and $\hbspg$ denote the RKHS for $X$ and $Y$, respectively (see Figure \ref{fig:conditional-embedding-summary}). These representations allow us to develop a powerful language for algebraic manipulation of probability distributions in an analogous way to the sum rule, product rule, and Bayes' rule---which are ubiquitous in graphical models and probabilistic inference---without making assumption on parametric forms of the underlying distributions. The details of conditional mean embeddings will be given in Section \ref{sec:conditional-embedding}.



\begin{paragraph}{A Synopsis.}
As a result of the aforementioned advantages, the kernel mean embedding has made widespread contributions in various directions. Firstly, most tasks in machine learning and statistics involve estimation of the data-generating process whose success depends critically on the accuracy and the reliability of this estimation. It is known that estimating the kernel mean embedding is easier than estimating the distribution itself, which helps improve many statistical inference methods. These include, for example, two-sample testing \citep{Gretton12:KTT}, independence and conditional independence tests \citep{Fukumizu2008,Zhang2011,Doran2014}, causal inference \citep{Sgouritsa2013,Chen2014}, adaptive MCMC \citep{Sejdinovic2014}, and approximate Bayesian computation \citep{Fukumizu13:KBR}.

Secondly, several attempts have been made in using kernel mean embedding as a representation in the predictive learning on distributions \citep{Muandet12:SMM,Zoltan15:DistReg,Muandet13:OCSMM,Guevara2014,Lopez-Paz15:Towards}. As opposed to the classical setting where training and test examples are data points, many applications call for a learning framework in which training and test examples are probability distributions. This is ubiquitous in, for example, multiple-instance learning \citep{Doran2013}, learning with noisy and uncertain input, learning from missing data, group anomaly detection \citep{Muandet13:OCSMM,Guevara2014}, dataset squishing, and bag-of-words data \citep{Yoshikawa14:LSMM,Yoshikawa15:GPLVM}. The kernel mean representation equipped with the RKHS methods enables classification, regression, and anomaly detection to be performed on such distributions.

Finally, the kernel mean embedding also allows one to perform complex approximate inference without making strong parametric assumption on the form of underlying distribution. The idea is to represent all relevant probabilistic quantities as a kernel mean embedding. Then, basic operations such as \emph{sum rule} and \emph{product rule} can be formulated in terms of the expectation and inner product in feature space. Examples of algorithms in this class include kernel belief propagation (KBP), kernel embedding of latent tree model, kernel Bayes rule, and predictive-state representation \citep{Song10:NTGM,Song10:KCOND,Song11:KBP,Song2013,Fukumizu13:KBR}. Recently, the kernel mean representation has become one of the prominent tools in causal inference and discovery \citep{Lopez-Paz15:Towards,Sgouritsa2013,Chen2014,Schoelkopf15:KPP}.
\end{paragraph}

The aforementioned examples represent only a handful of successful applications of kernel mean embedding. More examples and details will be provided throughout the survey.

\section{Purpose and Scope}

The purpose of this survey is to give a comprehensive review of kernel mean embedding of distributions, to present important theoretical results and practical applications, and to draw connections to related areas. We restrict the scope of this survey to key theoretical results and new applications of kernel mean embedding with references to related work. We focus primarily on basic intuition and sketches for proofs, leaving the full proofs to the papers cited.

All materials presented in this paper should be accessible to a wide audience. In particular, we hope that this survey will be most useful to readers who are not at all familiar with the idea of kernel mean embedding, but already have some background knowledge in machine learning. To ease the reading, we suggest non-expert readers to also consult elementary machine learning textbooks such as \citet{Bishop2006}, \citet{Scholkopf01:LKS}, \citet{Mohri12:FML}, and \citet{Murphy12:MLP}. Experienced machine learners who are interested in applying the idea of kernel mean embedding to their work are also encouraged to read this survey. Lastly, we will also provide some practical considerations that could be useful to practitioners who are interested in implementing the idea in real-world applications.

\section{Outline of the Survey}

\begin{figure}[t!]
\begin{tikzpicture}[node distance = 1cm,auto]
  \tikzstyle{cbox} = [rectangle, rounded corners, minimum width=1cm, text centered, draw=white, fill=blue!30]
  \node [cbox, drop shadow, yshift=-1cm,text width=2.5cm] (ch2) {\textbf{Section 2} Background};
  \node [cbox, drop shadow, xshift=4cm,yshift=1cm, text width=4cm] (ch3) {\textbf{Section 3} \\ Embedding of marginal distributions};
  \node [cbox, drop shadow, xshift=4cm,yshift=-1cm, text width=4cm] (ch4) {\textbf{Section 4} \\ Embedding of conditional distributions};
  \node [cbox, drop shadow, xshift=4cm,yshift=-3cm, text width=4cm] (ch5) {\textbf{Section 5} \\ Relationships to other methods};
  \node [cbox, drop shadow, xshift=9cm,yshift=-1cm, text width=2.9cm] (ch6) {\textbf{Section 6} Future directions};
  \node [cbox, drop shadow, xshift=9cm,yshift=-3cm, text width=2.9cm] (ch7) {\textbf{Section 7} Conclusions};

  \draw[->,very thick,rounded corners=3pt] (ch2) |- (ch3.west);
  \path[->,very thick] (ch3) edge (ch4);
  \path[->,very thick] (ch4) edge (ch5);
  \draw[->,very thick,rounded corners=3pt] (ch5.east) -- +(0.5,0) |- (ch6.west);
  \path[->,very thick] (ch6) edge (ch7);
\end{tikzpicture}
\caption{Schematic outline of this survey.} 
\label{fig:survey-outline}
\end{figure}
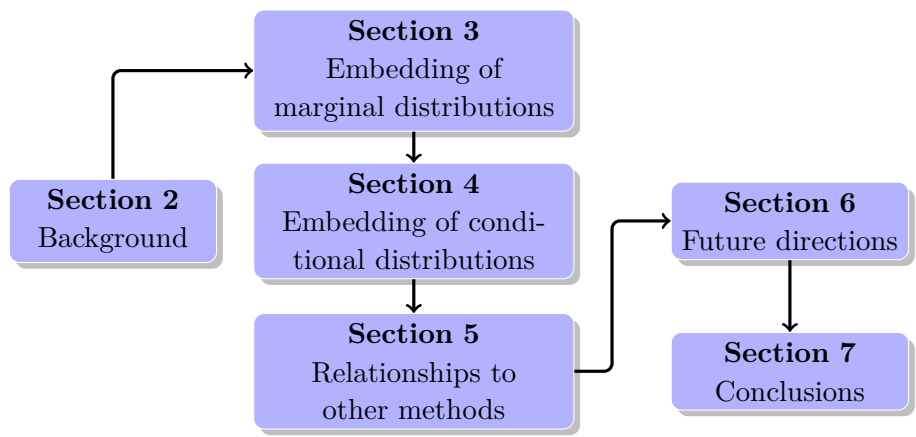

The schematic outline of this survey is depicted in Figure \ref{fig:survey-outline} and can be summarized as follows.

\begin{paragraph}{Section \ref{sec:background}} introduces notation and the basic idea of a positive definite kernel and reproducing kernel Hilbert space (RKHS) (\S\ref{sec:kernel} and \S\ref{sec:rkhs}). It also presents general theoretical results such as the reproducing property (Prop \ref{thm:reproducing-property}), the Riesz representation theorem (Thm \ref{thm:riesz-representation}), Mercer's theorem (Thm \ref{thm:mercer}), Bochner's theorem (Thm \ref{thm:bochner}), and Schoenberg's characterization (Thm \ref{thm:schoenberg}). In addition, it contains a brief discussion about Hilbert-Schmidt operators on RKHS (\S\ref{sec:HSO}).
\end{paragraph}

\begin{paragraph}{Section \ref{sec:marginal-embedding}} conveys the idea of Hilbert space embedding of marginal distributions (\S\ref{sec:point2measure}) as well as covariance operators (\S\ref{sec:cross-covariance}), presents essential properties of mean embedding (\S\ref{sec:embedding-theory}), discusses its estimation and approximation procedures (\S\ref{sec:kme}), and reviews important applications, notably maximum mean discrepancy (MMD) (\S\ref{sec:MMD}), kernel dependence measure (\S\ref{sec:dependency}), learning on distributional data (\S\ref{sec:distributional-data}), and how to recover information from the embedding of distributions (\S\ref{sec:dist-preimage}).
\end{paragraph}

\begin{paragraph}{Section \ref{sec:conditional-embedding}} generalizes the idea of kernel mean embedding to the space of conditional distributions, called \emph{conditional mean embedding} (\S\ref{sec:marginal2conditional}), presents regression perspective (\S\ref{sec:regression-view}), and describes basic operations---namely sum rule, product rule, and Bayes' rule---in terms of marginal and conditional mean embeddings (\S\ref{sec:basic-operations}). We review applications in graphical models, probabilistic inference (\S\ref{sec:graphical-models}), reinforcement learning (\S\ref{sec:mdp-reinforcement}), conditional dependence measures (\S\ref{sec:cond-dependency}), and causal discovery (\S\ref{sec:causal}). Estimating the conditional mean embedding is challenging both theoretically and empirically. We discuss some of the key challenges as well as some applications.
\end{paragraph}

\begin{paragraph}{Section \ref{sec:connections}}
  draws connections between the kernel mean embedding framework and other methods including kernel density estimation, empirical characteristic function, divergence methods and probabilistic modeling.
\end{paragraph}

\begin{paragraph}{Section \ref{sec:future}} provides suggestions for future research.
\end{paragraph}
  
\begin{paragraph}{Section \ref{sec:conclusions}} concludes the survey.
  \end{paragraph}  

\section{Notation} 

Table \ref{tab:notations} summarizes a collection of the commonly used notation and symbols used throughout the survey. 

\begin{longtable}[l]{p{50pt} p{270pt}}
  \caption{Notation and symbols} \label{tab:notations} \\
  \toprule 
  \textbf{Symbol} & \textbf{Description} \\
  \midrule \endhead
  $x$ & A scalar quantity \\
  $\x$ & A vector \\
  $\mathbf{X}$ & A matrix \\
  $X$ & A random variable \\
  $\rr$ & Real line or the field of real numbers \\
  $\rr^d$ & Euclidean $d$-space \\
  $\cc$ & Complex plane or the field of complex numbers \\ 
  $\cc^d$ & Complex $d$-space \\
  $\langle\cdot,\cdot \rangle$ & An inner product \\
  $\|\cdot\|$ & A norm \\
  $\inspace,\; \outspace$ & Non-empty spaces in which $X$ and $Y$ take values\\
  $\rr^{\inspace}$ & A vector space of functions from $\inspace$ to $\rr$ \\ 
  $\hbspace$ & Reproducing kernel Hilbert spaces (RKHS) of functions from $\inspace$ to $\rr$ \\ 
  $\hbspg$ & Reproducing kernel Hilbert spaces (RKHS) of functions from $\iny$ to $\rr$ \\ 
  $\hbspg\otimes\hbspace$ & Tensor product space \\
  $k(\cdot,\cdot)$ & Positive definite kernel function on $\inspace\times\inspace$\\
  $l(\cdot,\cdot)$ & Positive definite kernel function on $\iny\times\iny$\\
  $\phi(\cdot)$ & Feature map from $\inspace$ to $\hbspace$ associated with the kernel $k$ \\
  $\varphi(\cdot)$ & Feature map from $\iny$ to $\hbspg$ associated with the kernel $l$ \\
  $\kmat$ & Gram matrix with $\kmat_{ij} = k(\x_i,\x_j)$\\ 
  $\lmat$ & Gram matrix with $\lmat_{ij} = l(\y_i,\y_j)$\\ 
  $\mathbf{H}$ & Centering matrix \\ 
  $\kmat\circ\lmat$ & Hadamard product of matrices $\kmat$ and $\lmat$ \\
  $\kmat\otimes\lmat$ & Kronecker product of matrices $\kmat$ and $\lmat$ \\ 
  $\covx, \; \covy$ & Covariance operators in $\hbspace$ and $\hbspg$, respectively \\
  $\covxy$ & Cross-covariance operators from $\hbspg$ to $\hbspace$ \\
  $\mathcal{C}_{\mathit{XY}|Z}$ & Conditional cross-covariance operator \\
  $\mathcal{V}_{\mathit{YX}}$ & Normalized cross-covariance operator from $\hbspace$ to $\hbspg$ \\
  $\mathcal{V}_{\mathit{YX}|Z}$ & Normalized conditional cross-covariance operator \\
  $C_b(\inspace)$ & Space of bounded continuous functions on $\inspace$ \\
  $L_2[a,b]$ & Space of square-integrable functions on $[a,b]$ \\
  $L_2(\inspace,\mu)$ & Space of square $\mu$-integrable functions on $\inspace$ \\
$M^1_+(\mathcal{X})$ & Space of probability measures on $\inspace$\\
  $\ell^1$ & Space of sequences whose series is absolutely convergent \\
  $\ell^2$ & Space of square summable sequences \\
  $\ell^{\infty}$ & Space of bounded sequences \\
  $H^r_2(\rr^d)$ & Sobolev space of $r$-times differentiable functions \\
  $\text{HS}(\hbspg,\hbspace)$ &  Hilbert space of Hilbert-Schmidt operators mapping from $\hbspg$ to $\hbspace$ \\
$\mathfrak{F}g$ & Fourier transform of $g$\\
  $\mathrm{Id}$ & Identity operator \\
  $\Lambda$ & Spectral density \\
  $\mathcal{N}(\mathcal{C})$ & Null space of an operator $\mathcal{C}$\\
  $\mathcal{R}(\mathcal{C})$ & Range of an operator $\mathcal{C}$\\
  $S^{\bot}$ & Orthogonal complement of a closed subspace $S$ \\
  $(h_i)_{i\in I}$ & Orthonormal basis \\
  $\pp{P},\,\pp{Q}$ & Probability distributions \\  
  $\cf{\pp{P}}$ & Characteristic function of the distribution $\pp{P}$ \\
  $O(n)$ & Order $n$ time complexity of an algorithm \\
  $O_p(n)$ & Order $n$ in probability (or stochastic boundedness) \\
  \bottomrule
\end{longtable}


\chapter{Background}
\label{sec:background}
 
This section introduces the kernel methods and the concept of reproducing kernel Hilbert space (RKHS) which form the backbone of the survey. Readers who are familiar with 
probability theory and the concept of RKHS may skip this section entirely or return to it later. More detailed accounts on this topic can be found, in \citet{Scholkopf01:LKS}, \citet{Berlinet04:RKHS}, and \citet{Hofmann2008}, for example. Additional materials can also be found in \citet{Gretton16:RKHS}. Readers who are interested particularly in the typical applications of kernels in machine learning, \eg, support vector machines (SVMs) and Gaussian processes (GPs), are encouraged to read \citet{Cucker02:Learning}, \citet{Burges1998,Rasmussen2005} and references therein.

\section{Learning with Kernels}
\label{sec:kernel}

Many classical learning algorithms---such as the perceptron \citep{Rosenblatt1958:Perceptron}, support vector machine (SVM) \citep{Cortes1995}, and principal component analysis (PCA) \citep{Pearson01:PCA,Hotelling33:PCA}---employ data instances, e.g., $\x,\x'\in\inspace$, only through an inner product $\langle \x,\x' \rangle$, which basically is a similarity measure between $\x$ and $\x'$. However, the class of linear functions induced by this inner product may be too restrictive for many real-world problems. Kernel methods aim to build more flexible and powerful learning algorithms by replacing $\langle \x,\x' \rangle$ with some other, possibly non-linear, similarity measures.

The most natural extension of $\langle\x,\x'\rangle$ is to explicitly apply a non-linear transformation:
\begin{eqnarray}
  \label{eq:feature-map-1} 
  \phi\,:\, \inspace &\longrightarrow& \mathcal{F}, \nonumber \\
  \x &\longmapsto& \phi(\x),
\end{eqnarray}
\noindent into a possibly high-dimensional \emph{feature space} $\mathcal{F}$ and subsequently evaluate the inner product in $\mathcal{F}$, \ie,
\begin{equation}
  \label{eq:kernel-inner}
  k(\x,\x') := \langle \phi(\x),\phi(\x') \rangle_\mathcal{F}, 
\end{equation}
where $\langle\cdot,\cdot\rangle_\mathcal{F}$ denotes the inner product of $\mathcal{F}$. 
We will refer to $\phi$ and $k$ as a \emph{feature map} and a \emph{kernel function}, respectively. Likewise, we can interpret $k(\x,\x')$ as a non-linear similarity measure 
between $\x$ and $\x'$. Since most algorithms depend on the data set only through the inner product $\langle \x,\x'\rangle$, we can obtain a non-linear extension of these algorithms 
by simply substituting $\langle \x,\x'\rangle$ with $\langle \phi(\x),\phi(\x')\rangle_\mathcal{F}$. Note that the learning algorithm remains the same: we only change the space in which these algorithms operate. As \eqref{eq:feature-map-1} is non-linear, a linear algorithm in $\mathcal{F}$ corresponds to the non-linear counterpart in the original space $\inspace$.

For example, consider a polynomial feature map $\phi(\x) = (x_1^2,x_2^2,\sqrt{2}x_1x_2)$ when $\x\in\rr^2$. Then we have
\begin{equation}
  \label{eq:poly-feature}
  \langle \phi(\x),\phi(\x') \rangle_{\mathcal{F}} = x_1^2x_1'^2 +x_2^2x_2'^2+2x_1x_2x'_1x'_2 = \langle \x,\x'\rangle^2.
\end{equation}
Put differently, the new similarity measure is simply the square of the inner product in $\inspace$. The equality \eqref{eq:poly-feature} holds more generally for a $d$-degree polynomial, \ie, $\phi$ maps $\x\in\rr^N$ to the vector $\phi(\x)$ whose entries are all possible $d$th degree ordered products of the entries of $\x$. In that case, we have $k(\x,\x') = \langle \phi(\x),\phi(\x')\rangle_{\mathcal{F}} = \langle \x,\x' \rangle^d$. Thus, the complexity of the learning algorithm is controlled by the complexity of $\phi$ and by increasing the degree $d$, one would expect the resulting algorithm to become more complex. Additional examples of how to construct an explicit feature map can be found in \citet[Chapter 2]{Scholkopf01:LKS}.

Unfortunately, evaluating $k(\x,\x')$ as above requires a two-step procedure: i) construct the feature maps $\phi(\x)$ and $\phi(\x')$ explicitly, and ii) then evaluate $\langle\phi(\x),\phi(\x')\rangle_{\mathcal{F}}$. These two steps can be computationally expensive if $\phi(\x)$ lives in a high-dimensional feature space, \eg, when the degree $d$ of the polynomial is large. Fortunately, \eqref{eq:poly-feature} implies that there is an alternative way to evaluate $\langle\phi(\x),\phi(\x')\rangle_{\mathcal{F}}$ without resorting to constructing $\phi(\x)$ explicitly if all we need is an inner product $\langle \phi(\x),\phi(\x')\rangle_{\mathcal{F}}$. That is, it is sufficient to consider $k(\x,\x') = \langle \x,\x' \rangle^2$ directly. 
This is an essential aspect of kernel methods, often referred to 
as the \emph{kernel trick} in the machine learning community.

When is such trick possible? It turns out that if $k$ is \emph{positive definite} (cf. Definition \ref{def:psd}) there always exists $\phi:\inspace\rightarrow\mathcal{F}$ for which $k(\x,\x') = \langle \phi(\x),\phi(\x')\rangle_{\mathcal{F}}$. Since $\langle \cdot,\cdot\rangle$ is positive definite, it follows that $k$ defined as in \eqref{eq:kernel-inner} is positive definite for any choice of explicit feature map $\phi$.

\begin{shadowbox}
\begin{definition}[positive definite kernel]
  \label{def:psd}
  A function $k:\inspace\times\inspace\rightarrow\rr$ is a positive definite kernel 
 if it is symmetric, \ie, $k(\x,\y)=k(\y,\x)$, and the Gram matrix is positive definite:
  \begin{equation}
    \label{eq:positive-definite}
    \sum_{i,j=1}^n c_ic_jk(\x_i,\x_j) \geq 0 ,
  \end{equation}
  \noindent for any $n\in\mathbb{N}$, any choice of $\x_1,\ldots,\x_n\in\inspace$ and any $c_1,\ldots,c_n\in\rr$. It is said to be \emph{strictly} positive definite if equality in \eqref{eq:positive-definite} implies $c_1 = c_2 = \cdots = c_n = 0$.
\end{definition}
\end{shadowbox}


Moreover, a positive definite kernel in Definition \ref{def:psd} defines a space of functions from $\inspace$ to $\rr$ called a \emph{reproducing kernel Hilbert space} (RKHS) $\hbspace$, hence it is also called a \emph{reproducing kernel} \citep{aronszajn50reproducing}. We defer the details of this link to Section \ref{sec:rkhs}. An RKHS has two important properties: (i) for any $\x\in\inspace$, the function $k(\x,\cdot): \y\mapsto k(\x,\y)$ is an element of $\hbspace$. Whenever we use the kernel $k$, the feature space $\mathcal{F}$ is the RKHS $\hbspace$ associated with this kernel and we often consider the \emph{canonical feature map}
\begin{eqnarray}
  \label{eq:kernel-map}
  k \,:\, \inspace &\rightarrow& \hbspace \subset \rr^{\inspace}, \nonumber \\*
  \x &\mapsto& k(\x,\cdot),
\end{eqnarray}
\noindent where $\rr^\inspace$ denotes the vector space of functions from $\inspace$ to $\rr$; (ii) the kernel $k$ represents point evaluation, \ie, for all $f\in\hbspace$ and $\x\in\inspace$,
\begin{equation}
  \label{eq:reproducing}
  f(\x) = \langle f, k(\x,\cdot) \rangle_{\hbspace} .
\end{equation}
If $f(\x')=k(\x',\cdot)$ for some $\x'\in\inspace$, we obtain the \emph{reproducing property}, \ie,  $\langle k(\x,\cdot),k(\x',\cdot) \rangle_{\hbspace} = k(\x,\x') $, as a special case. Although we do not need to know $\phi(\x) = k(\x,\cdot)$ explicitly, it can be derived directly from the kernel $k$ (see, \eg, \citealp{Scholkopf01:LKS} for concrete examples). Further 
details of RKHS will be given in Section \ref{sec:rkhs}.

The capability of the kernel trick not only results in powerful learning algorithms, but also allows domain experts to easily invent domain-specific kernel functions which are suitable for particular applications. This leads to a number of kernel functions in various application domains \citep{Genton02:Kernel}. In machine learning, commonly used kernels include the Gaussian and Laplace kernels, \ie,
\begin{equation}
  k(\x,\x') = \exp\left(-\frac{\|\x-\x'\|_2^2}{2\sigma^2}\right), \qquad
  k(\x,\x') = \exp\left(-\frac{\|\x-\x'\|_1}{\sigma}\right),
\end{equation}
\noindent where $\sigma > 0$ is a bandwidth parameter. These kernels belong to a class of kernel functions called \emph{radial basis functions} (RBF). Both these kernels are \emph{translation invariant} on $\pp{R}^d$ which form an important class of kernel functions with essential properties (cf. Theorem \ref{thm:bochner}).\footnote{A kernel $k$ is said to be translation invariant on $\pp{R}^d$ if $k(\x,\x')=\psi(\x-\x'),\,\x,\x'\in\pp{R}^d$ for some positive definite function $\psi$.}

Another essential property of the kernel trick is that it can be applied not only to Euclidean data, but also to non-Euclidean structured data, functional data, and other domains on which positive definite kernels may be defined \citep{Scholkopf01:LKS,Gartner2003}. A review of several classes of kernel functions can be found in \citet{Genton02:Kernel}. A review of kernels for vector-valued functions can be found in \citet{Alvarez12:KVF}. \citet{Hofmann2008} also provides a general review of kernel methods in machine learning.

Next, we give an important characterization of a continuous positive definite kernel $k$ on a compact set, known as Mercer's theorem \citep{Mercer09:Mercer}. For the proof, see, \eg, \citet[pp. 59]{Felipe07:LearningTheory}. 
\begin{shadowbox} 
\begin{theorem}[Mercer's theorem]
  \label{thm:mercer}
Let $\inspace$ be a compact Hausdorff space and $\mu$ be a finite Borel measure with support $\inspace$.\footnote{see, \eg, \citet{Kreyszig78:FA} and \citet{Rudin91:FA} for the details of a compact Hausdorff space and Borel measures.}  Suppose $k$ is a continuous positive definite kernel on $\inspace$, and define the integral operator $\mathcal{T}_k: L_2(\inspace,\mu)\rightarrow L_2(\inspace,\mu)$ by
  \begin{equation*}
    (\mathcal{T}_kf)(\cdot) = \int_\inspace k(\cdot,\x)f(\x)\dd\mu(\x),
  \end{equation*}
  \noindent which is positive definite, \ie, $\forall f\in L_2(\inspace,\mu)$,
  \begin{equation} 
    \label{eq:mercer-condition}
    \int_\inspace k(\mathbf{u},\mathbf{v})f(\mathbf{u})f(\mathbf{v}) \dd \mathbf{u}\dd \mathbf{v} \geq 0 .
  \end{equation}
  Then, there is an orthonormal basis $\{\psi_i\}$ of $L_2(\inspace,\mu)$ consisting of eigenfunctions of $\mathcal{T}_k$ such that the corresponding sequence of eigenvalues $\{\lambda_i\}$ are non-negative. The eigenfunctions corresponding to non-zero eigenvalues can be taken as continuous functions on $\inspace$ and $k(\mathbf{u},\mathbf{v})$ has the representation
  \begin{equation}
    \label{eq:mercer-rep}
    k(\mathbf{u},\mathbf{v}) = \sum_{i=1}^\infty \lambda_i\psi_i(\mathbf{u})\psi_i(\mathbf{v}) ,
  \end{equation}
  \noindent where the convergence is absolute and uniform.
\end{theorem}
\end{shadowbox}

The condition \eqref{eq:mercer-condition} is known as the \emph{Mercer's condition} and the kernel functions that satisfy this condition are often referred to as \emph{Mercer kernels}. An important consequence of Mercer's theorem is that it is possible to derive the explicit form of the feature map $\phi(\x)$ from \eqref{eq:mercer-rep}, \ie, $\phi(\x) = [\sqrt{\lambda_1}\psi_1(\x),\sqrt{\lambda_2}\psi_2(\x),\ldots]$. \citet{Steinwart12:Mercer} studied Mercer's theorem in general domains by relaxing the compactness assumption on $\inx$. Moreover, there is an intrinsic connection between the integral operator $\mathcal{T}_k$, covariance operator (\S\ref{sec:cross-covariance}), and Gram matrix $\kmat$, see, \eg, \citet{Rosasco10:Integral} for further details. 

When the kernel $k$ is translation invariant on $\pp{R}^d$, \ie, $k(\x,\x') = \psi(\x-\x'), \,\x,\x'\in\rr^d$ where $\psi$ is called a positive definite function, the kernel can be characterized by Bochner's theorem \citep{Bochner33}. It states that any bounded continuous kernel $k$ is the inverse Fourier transform of some finite non-negative Borel measure.
 
\begin{shadowbox}
\begin{theorem}[Bochner's theorem]
  \label{thm:bochner}
  A complex-valued bounded continuous kernel $k(\x,\x') = \psi(\x-\x')$ on $\mathbb{R}^d$ is positive definite if and only if there exists a finite non-negative Borel measure $\Lambda$ on $\rr^d$ such that
  \begin{equation}
    \label{eq:bochner}
    \psi(\x-\x') = \int_{\rr^d} e^{\sqrt{-1}\bm{\omega}^\top(\x-\x')}\dd\Lambda(\bm{\omega}).
  \end{equation} 
\end{theorem}
\end{shadowbox}
\noindent Put differently, the continuous positive definite functions form a convex cone generated
by the Fourier kernels $\{e^{\sqrt{-1}\bm{\omega}^\top(\x-\x')}\,|\, \bm{\omega}\in\rr^d\}$.

\begin{table}[t!]
  \centering
  \caption{Well-known kernel functions and their corresponding spectral densities. More examples can be found in \citet{Rasmussen2005,Rahimi2007,Fukumizu2009,Kar2012,Pham2013}. $K_{\nu}$ is the modified Bessel function of the second kind of order $\nu$ and $\Gamma$ is the Gamma function. We also define $h(\nu,d\sigma) = \frac{2^d\pi^{d/2}\Gamma(\nu+d/2)(2\nu)^\nu}{\Gamma(\nu)\sigma^{2\nu}}$.}
  \label{tab:kernel-fourier}
  \resizebox{\textwidth}{!}{
  \begin{tabular}{lll} 
    \toprule
    Kernel & $k(\x,\x')$ & $\Lambda(\bm{\bm{\omega}})$ \\
    \midrule
    Gaussian & $\exp\left(-\frac{\|\x-\x'\|^2_2}{2\sigma^2}\right),\; \sigma > 0$ & $\frac{1}{(2\pi/\sigma^2)^{d/2}}\exp\left(-\frac{-\sigma^2\|\bm{\bm{\omega}}\|^2_2}{2}\right)$ \\
    Laplace & $\exp\left(-\frac{\|\x-\x'\|_1}{\sigma}\right) ,\; \sigma > 0$ & $\prod_{i=1}^d\frac{\sigma}{\pi(1+\omega_i^2)}$ \\
Cauchy & $\prod_{i=1}^d \frac{2}{1 + (x_i-x'_i)^2}$ & $\exp\left(-\|\bm{\omega}\|_1\right)$ \\
    Mat\'ern & $\frac{2^{1-\nu}}{\Gamma(\nu)}\left(\frac{\sqrt{2\nu}\|\x-\x'\|_2}{\sigma}\right)K_\nu\left(\frac{\sqrt{2\nu}\|\x-\x'\|_2}{\sigma}\right)$ & $h(\nu,d\sigma)\left(\frac{2\nu}{\sigma^2}+4\pi^2\|\bm{\omega}\|^2_2\right)^{\nu+d/2}$ \\
    & \; $\sigma >0,\; \nu > 0$ &  \\
    \bottomrule
  \end{tabular}}
\vspace{-1em}
\end{table}

By virtue of Theorem \ref{thm:bochner}, we may interpret the similarity measure $k(\x,\x') = \psi(\x-\x')$ in the Fourier domain. That is, the measure $\Lambda$ determines which frequency component occurs in the kernel by putting a non-negative power on each frequency $\bm{\omega}$. Note that we may normalize $k$ such that $\psi(\mathbf{0}) = 1$, in which case $\Lambda$ will be a probability measure and $k$ corresponds to its characteristic function. For example, the measure $\Lambda$ that corresponds to the Gaussian kernel $k(\x-\x') = e^{-\|\x-\x'\|^2_2/(2\sigma^2)}$ is a Gaussian distribution of the form $(2\pi/\sigma^2)^{-d/2}e^{-\sigma^2\|\bm{\omega}\|^2_2/2}$. For the Laplacian kernel $k(\x,\x') = e^{-\|\x-\x'\|_1/\sigma}$, the corresponding measure is a Cauchy distribution, \ie, $\Lambda(\bm{\omega}) = \prod_{i=1}^d\frac{\sigma}{\pi(1+\omega_i^2)}$. Table \ref{tab:kernel-fourier} summarizes the Fourier transform of some well-known kernel functions.

As we will see later, Bochner's theorem also plays a central role in providing a powerful characterization of the kernel mean embedding. That is, the measure $\Lambda(\bm{\omega})$ determines which frequency component $\bm{\omega}$ of the characteristic function of $\pp{P}$ appears in the embedding $\muv_{\pp{P}} = \ep_{\x\sim\pp{P}}[k(\x,\cdot)]$. Hence, if the $\Lambda$ is supported on the entire $\rr^d$, we have by the uniqueness of the characteristic function that $\muv_{\pp{P}}$ 
uniquely determine $\pp{P}$ \citep{Sriperumbudur08injectivehilbert,Sriperumbudur10:Metrics,Sriperumbudur2011}. Specifically, if $k$ is translation invariant on $\rr^d$ and $\cf{\pp{P}}$ denotes the characteristic function of $\pp{P}$, it follows that
\begin{align}
  \label{eq:bochner-mean-map}
  \langle \muv_{\pp{P}},\muv_{\pp{Q}}\rangle_{\hbspace} &= \left\langle \int k(\x,\cdot) \dd\pp{P}(\x), \int k(\x',\cdot) \dd\pp{Q}(\x') \right\rangle_{\hbspace}  \nonumber \\ 
  &= \int\int k(\x,\x') \dd\pp{P}(\x)\dd\pp{Q}(\x') \nonumber \\ 
  &= \int\int \left[\int e^{\sqrt{-1}\bm{\omega}^\top(\x-\x')}\dd\Lambda(\bm{\omega}) \right] \dd\pp{P}(\x)\dd\pp{Q}(\x') \nonumber \\
  &= \int \left[\int e^{\sqrt{-1}\bm{\omega}^\top \x}\dd\pp{P}(\x)\right] \left[\int e^{-\sqrt{-1}\bm{\omega}^\top \x'}\dd\pp{Q}(\x')\right] \dd\Lambda(\bm{\omega}) \nonumber \\
  &= \langle \cf{\pp{P}},\cf{\pp{Q}}\rangle_{L^2(\rr^d,\Lambda)},
\end{align}
\noindent where we invoked the reproducing property of $\hbspace$ in the second equation and Theorem \ref{thm:bochner} in the third equation, respectively. Hence, we may think of $\Lambda$ as a filter that selects certain properties when computing the similarity measure between probability distributions using $\langle \muv_{\pp{P}},\muv_{\pp{Q}}\rangle_{\hbspace}$ w.r.t. a certain class of distributions (more below).

Another promising application of Bochner's theorem is the \emph{approximation of the kernel function}, which is useful in speeding up kernel methods. The feature map $\phi$ of many kernel functions such as the Gaussian kernel is infinite dimensional. To avoid the need to construct the Gram matrix $\kmat_{ij} = k(\x_i,\x_j)$---which scales at least $O(n^2)$ in terms of both computation and memory usage---\citet{Rahimi2007} proposes to approximate the integral in \eqref{eq:bochner} based on a Monte Carlo sample $\bm{\omega}\sim\Lambda$. The resulting approximation is known as random Fourier features (RFFs). See also \citet{Kar2012,Vedaldi12:AdditiveKernels,Le2013,Pham2013} and references therein for a generalization of this idea. Theoretical results on the preservation of kernel evaluation can be found in, \eg, \citet{SutherlandS15:RF} and \citet{Sriperumbudur15:RF}. Other approaches to approximating the Gram matrix $\kmat$ are low-rank approximation via incomplete Cholesky decomposition \citep{Fine01:LowRank} and the Nystr\"{o}m method \citep{Williams01:Nystrom, Bach13:LowRank}.


Finally, we briefly mention Schoenberg's characterization \citep{Schoenberg38:Metr} for a class of \emph{radial} kernels. 

\begin{shadowbox}
  \begin{definition}[Radial kernels]
    The kernel $k$ is said to be \emph{radial} if $k(\x,\x')=\psi(\|\x-\x'\|)$ for some positive definite function $\psi : [0,\infty)\rightarrow \rr$. 
  \end{definition}
\end{shadowbox}

\noindent That is, a radial kernel is a translation-invariant kernel that depends only on the distance between two instances. This class of kernel functions is characterized by Schoenberg's theorem.
  
\begin{shadowbox}
\begin{theorem}[Schoenberg's theorem]
  \label{thm:schoenberg}
  A continuous function $\psi: [0,\infty)\rightarrow\rr$ is positive definite and radial on $\rr^d$ for all $d$ if and only if it is of the form
    \begin{equation}
      \psi(r) = \int_{0}^{\infty}e^{-r^2t^2}\dd\mu(t) ,
    \end{equation}
    where $\mu$ is a finite non-negative Borel measure on $[0,\infty)$.
\end{theorem}
\end{shadowbox}
\noindent In other words, this class of kernels can be expressed as a scaled Gaussian mixture. Examples of well-known radial kernels include the Gaussian RBF kernel $k(\x,\y)=\exp(-\|\x-\y\|^2/2\sigma^2)$, the mixture-of-Gaussians kernel $k(\x,\y)=\sum_{j=1}^K\beta_j\exp(-\|\x-\y\|^2/2\sigma^2_j)$ where $\bm{\beta} \geq \mathbf{0}$ and $\sum_{j=1}^K\beta_j=1$, the inverse multiquadratic kernel $k(\x,\y) = (c^2 + \|\x-\y\|^2)^{-\gamma}$ where $c,\gamma > 0$, and the Mat\'{e}rn kernel
\begin{equation}
  \label{eq:matern}
  k(\x,\y) =  \frac{2^{1-\nu}}{\Gamma(\nu)}\left(\frac{\sqrt{2\nu}\|\x-\x'\|_2}{\sigma}\right)K_\nu\left(\frac{\sqrt{2\nu}\|\x-\x'\|_2}{\sigma}\right) ,
\end{equation}
\noindent where $\sigma>0$, $\nu>0$, $K_\nu$ is the modified Bessel function of the second kind of order $\nu$, and $\Gamma$ is the Gamma function. \citet{Pennington15:Poly} apply Schoenberg's theorem to build \emph{spherical random Fourier} (SRF) features of polynomial kernels, which are generally unbounded, for data on the unit sphere. For a detailed exposition on radial kernels and Schoenberg's theorem, see, \eg, \citet{Bavaud11:Schoenberg}, \citet[Ch. 7]{Wendland-05}, and \citet{Rasmussen2005}.

\section{Reproducing Kernel Hilbert Spaces}
\label{sec:rkhs} 
 
A reproducing kernel Hilbert space (RKHS) $\hbspace$ is a Hilbert space where all evaluation functionals in $\hbspace$ are bounded. First, we give a formal definition of a Hilbert space \citep[Chapter 3]{Kreyszig78:FA}.

\begin{shadowbox}
\begin{definition}[Hilbert spaces]
  Let $\langle \cdot,\cdot \rangle : \mathcal{V}\times\mathcal{V}\rightarrow\rr$ be an inner product on a real or complex vector space $\mathcal{V}$ and let $\|\cdot\| : \mathcal{V}\rightarrow\rr$ be the associated norm defined as
    $$\|\mathbf{v}\| = \sqrt{\langle \mathbf{v},\mathbf{v}\rangle}, \qquad \mathbf{v}\in\mathcal{V}.$$
    The vector space $\mathcal{V}$ is a Hilbert space if it is complete with respect to this norm, \ie, every Cauchy sequence in $\mathcal{V}$ has a limit which is an element of $\mathcal{V}$.\footnote{A sequence $\{\mathbf{v}_n\}_{n=1}^\infty$ of elements of a normed space $\mathcal{V}$ is a Cauchy sequence if for every $\varepsilon > 0$, there exist $N = N(\varepsilon)\in\mathbb{N}$, such that $\| \mathbf{v}_n - \mathbf{v}_m\|_{\mathcal{V}} < \varepsilon$ for all $m,n \geq N$.}
\end{definition}
\end{shadowbox}

Examples of Hilbert spaces include the standard Euclidean space $\rr^d$ with $\langle \x,\y\rangle$ being the dot product of $\x$ and $\y$, the space of square summable sequences $\ell^2$ of $\mathbf{x}=(x_1,x_2,\ldots)$ with an inner product $\langle \x,\y \rangle = \sum_{i=1}^\infty x_iy_i$, and the space of square-integrable functions $L_2[a,b]$ with inner product $\langle f,g \rangle = \int_a^b f(x)g(x)\dd x$, after identifying functions having different values only on measure zero sets. Every Hilbert space is a \emph{Banach space}, but the converse need not hold because a Banach space may have a norm that is not given by an inner product, \eg, the supremum norm $\|f\| = \sup_{x\in\inspace} |f(x)|$ \citep{Megginson98:Banach,Folland99:Real}. This survey will deal mostly with the embedding of distributions in a Hilbert space, while extensions to Banach spaces are investigated in \citet{Zhang09:RKB} and \citet{Sriperumbudur2011a}.

We are now in a position to give the definition of a reproducing kernel Hilbert space.
\begin{shadowbox}
\begin{definition}
  A Hilbert space $\hbspace$ of functions is \emph{a reproducing kernel Hilbert space (RKHS)} if the evaluation functionals $\mathcal{F}_{\x}[f]$ defined as $\mathcal{F}_{\x}[f] = f(\x)$ are bounded, \ie, for all $\x\in\inspace$ there exists some $C > 0$ such that
  \begin{equation}
    \label{eq:bounded-functional}
    \left| \mathcal{F}_{\x}[f]\right| = \left| f(\x) \right| \leq C\|f\|_{\hbspace}, \qquad \forall f\in\hbspace .
  \end{equation}
\end{definition}
\end{shadowbox}
\noindent Intuitively speaking, functions in the RKHS are \emph{smooth} in the sense of \eqref{eq:bounded-functional}. This smoothness property ensures that the solution in the RKHS obtained from learning algorithms will be well-behaved, since regularization on the RKHS norm leads to regularization on function values. For example, in classification and regression problems, it is ensured that by minimizing the empirical risk on the training data w.r.t. the functions 
in the RKHS, we obtain a solution $\hat{f}$ that is \emph{close} to the true solution $f$ and also generalizes well to unseen test data. This does not necessarily hold for functions in 
arbitrary Hilbert spaces. The space of square-integrable functions $L_2[a,b]$ does not have this property. It is very easy to find a function in $L_2[a,b]$ that attains zero risk on the training data, but performs poorly on unseen data, \ie, \emph{overfitting}.

The next theorem known as Riesz representation theorem provides a characterization of a bounded linear operator in $\hbspace$ (see, \eg,  \citealp{TLOHP93} for the proof).
\begin{shadowbox}
\begin{theorem}[Riesz representation]
  \label{thm:riesz-representation}
  If $\mathcal{A}:\hbspace\rightarrow\rr$ is a bounded linear operator in a Hilbert space $\hbspace$, there exists some $g_{\mathcal{A}}\in\hbspace$ such that
  \begin{equation}
    \label{eq:riesz}
    \mathcal{A} [f] = \langle f, g_{\mathcal{A}}\rangle_{\hbspace}, \qquad \forall f\in\hbspace .
  \end{equation}
\end{theorem} 
\end{shadowbox}
\noindent The Riesz representation theorem will be used to prove a sufficient condition for the existence of the kernel mean embedding in an RKHS (see Lemma \ref{lem:kme-existence}). By the definition of a RKHS, the evaluation functional $\mathcal{F}_{\x}[f] = f(\x)$ is a bounded linear operator in $\hbspace$. Therefore, the Riesz representation theorem ensures that for any $\x\in\inspace$ we can find an element in $\hbspace$ that is a \emph{representer} of the evaluation $f(\x)$. Proposition \ref{thm:reproducing-property} states this result, which is called the \emph{point evaluation property}.

\begin{shadowbox}
\begin{proposition}[point evaluation property] 
  \label{thm:reproducing-property}
  For each $\x\in\inspace$, there exists a function $k_{\x}\in\hbspace$ such that
  \begin{equation}
    \label{eq:reproducing-property}
    \mathcal{F}_{\x}[f] = \langle k_{\x},f \rangle_{\hbspace} = f(\x).
  \end{equation}
\end{proposition}
\end{shadowbox} 

The function $k_{\x}$ is called the reproducing kernel for the point $\x$. The \emph{reproducing property} is the special case of Proposition \ref{thm:reproducing-property} when $f=k(\x',\cdot)$. Let $k:\inspace\times\inspace\rightarrow\rr$ be a two-variable function defined by $k(\x,\y) := k_{\y}(\x)$. Then, it follows from the reproducing property that
\begin{equation}
  k(\x,\y) = k_{\y}(\x) = \langle k_{\x},k_{\y}\rangle_{\hbspace} = \langle \phi(\x),\phi(\y) \rangle_{\mathcal{\hbspace}} ,
\end{equation}
\noindent where $\phi(\x) := k_{\x}$ is the feature map $\phi:\inspace\rightarrow\mathcal{\hbspace}$. As mentioned earlier, we call $\phi$ a \emph{canonical feature map} associated with $\hbspace$ essentially because when we apply the function $k(\x,\y)$ in the learning algorithms, the data points are implicitly represented by a function $k_{\x}$ in the feature space. As we will see in Section \ref{sec:marginal-embedding}, the kernel mean embedding is defined by means of $k_{\x}$ and can itself be viewed as a canonical feature map of the probability distribution.

The RKHS $\hbspace$ is fully characterized by the reproducing kernel $k$. In fact, the RKHS uniquely determines $k$, and vice versa, as stated in the following theorem which is due to \citet{aronszajn50reproducing}:
\begin{shadowbox}
\begin{theorem}
  For every positive definite function $k(\cdot,\cdot)$ on $\inx\times\inx$ there exists a unique (up to isometric isomorphism)  RKHS with $k$ as its reproducing kernel. Conversely, the reproducing kernel of an RKHS is
  unique and positive definite.
\end{theorem}
\end{shadowbox}
\noindent 
It can be shown that the kernel $k$ generates the RKHS, \ie, $\hbspace = \overline{\text{span}\{k(\x,\cdot)\; | \, \x\in\inspace\}}$, 
where the closure is taken w.r.t. the RKHS norm \citep[Chapter 1, Theorem 3]{Berlinet04:RKHS}. For example, the RKHS of the translation invariant kernel $k(\x,\y)=\psi(\x-\y)$ on $\rr^d$ consists of functions whose smoothness is determined by the decay rate of the Fourier transform of $\psi$. 
The Mat\'{e}rn kernel \eqref{eq:matern} generates spaces of differentiable functions known as the \emph{Sobolev spaces} \citep{Adams03:Sobolev}. Detailed exposition on RKHSs can be found in \citet{Scholkopf01:LKS} and \citet{Berlinet04:RKHS}, for example.

\section{Hilbert-Schmidt Operators}
\label{sec:HSO}

We conclude this part by describing the notion of the \emph{Hilbert-Schmidt operator} which is ubiquitous in many applications of RKHS. Before proceeding to the detail, we first define a \emph{separable} Hilbert space.
\begin{shadowbox}
\begin{definition}[separable Hilbert spaces]
  A Hilbert space $\hbspace$ is said to be separable if it has a countable basis.
\end{definition}
\end{shadowbox}
\noindent Separability of $\hbspace$ ensures that there is an orthonormal basis for it and every orthonormal system of vectors in $\hbspace$ consists of a countable number of elements. It follows that every separable, infinite-dimensional Hilbert space is isometric to the space $\ell^2$ of square-summable sequences. 

Let $\hbspace$ and $\hbspf$ be separable Hilbert spaces and $(h_i)_{i\in I}$ and $(f_j)_{j\in J}$ are orthonormal basis for $\hbspace$ and $\hbspf$, respectively. A Hilbert-Schmidt operator is a bounded operator $\mathcal{A}:\hbspf \rightarrow \hbspace$ whose Hilbert-Schmidt norm
\begin{equation}
  \|\mathcal{A}\|^2_{\text{HS}} = \sum_{j\in J}\|\mathcal{A} f_j\|^2_{\hbspace}
  = \sum_{i\in I}\sum_{j\in J}|\langle \mathcal{A} f_j,h_i\rangle_{\hbspace}|^2 
\end{equation}
\noindent is finite. It can be easily shown that the right-hand side does not depend on the choice of orthonormal bases \citep[Lemma 5.5.1]{Davies07:LinOpt}. An example of a Hilbert-Schmidt operator is the covariance operator defined later in Section \ref{sec:cross-covariance}. Moreover, the identity operator $\mathrm{Id} : \hbspace\rightarrow\hbspace$ is Hilbert-Schmidt if and only if $\hbspace$ is finite dimensional as $\|\mathrm{Id}\|^2_{\text{HS}} = \sum_{j\in J}\|f_j\|^2_{\hbspace} = \sum_{j\in J} 1$ clearly diverges for an infinite dimensional space. The Hilbert-Schmidt operators mapping from $\hbspf$ to $\hbspace$ form a Hilbert space $\text{HS}(\hbspf,\hbspace)$ with inner product $\langle \mathcal{A},\mathcal{B} \rangle_{\text{HS}} = \sum_{j\in J}\langle \mathcal{A} f_j,\mathcal{B} f_j\rangle_{\hbspace}$. The Hilbert space of Hilbert-Schmidt operators is beyond the scope of this survey; see, \eg, \citet[Chapter 12]{Rudin91:FA} or \citet[Chapter 6]{ReedSimon81:FA} for further details.

Recently, the Hilbert-Schmidt operators have received much attention in the machine learning community and form a backbone of many modern applications of kernel methods. For instance, 
\citet{Gretton05:MSD} uses a Hilbert-Schmidt norm of the cross-covariance 
operator as a measure of statistical dependence between two random variables; see also \citet{Chwialkowski2014} and \citet{ChwialkowskiSG14:WILD} for an extension 
to random processes. Likewise, these notions have also been applied in sufficient dimension reduction \citep{Fukumizu04:DRS}, kernel CCA \citep{Fukumizu07:KCCA}, kernel PCA \citep{Zwald04:KPCA}, and kernel Bayes' rule \citep{Fukumizu13:KBR}. \citet{Quang14:LogHS} proposes a \emph{Log-Hilbert-Schmidt metric} between positive definite operators on a Hilbert space. It is applied in particular to compute the distance between covariance operators in a RKHS with applications in multi-category image classification.

\chapter{Hilbert Space Embedding of Marginal Distributions}
\label{sec:marginal-embedding}

This section introduces the idea of Hilbert-space embedding of distributions by generalizing the standard viewpoint of the kernel feature map of random sample to Dirac measures (see \eqref{eq:dirac-measure}), and then to more general probability measures. The presentation here is similar to that of \citet[Chapter 4]{Berlinet04:RKHS} with some modifications. We summarize this generalization in Figure \ref{fig:from-point-to-measure}.

\section{From Data Point to Probability Measure}
\label{sec:point2measure} 
 
Let $\inspace$ be a fixed non-empty set and $k:\inspace\times\inspace\rightarrow\rr$ be a real-valued positive definite kernel function associated with the Hilbert space $\hbspace$. 
For all functions $f\in\hbspace$, it follows from the reproducing property that $\langle k(\x,\cdot),f\rangle_\hbspace = f(\x)$. In particular, 
$k(\x,\y) = \langle k(\x,\cdot),k(\y,\cdot)\rangle_\hbspace$. By virtue of this property, we may view the kernel evaluation as an inner product in $\hbspace$ induced by a map from $\inspace$ into $\hbspace$
\begin{equation}
  \label{eq:feature-map}
  \x \longmapsto k(\x,\cdot) .
\end{equation}
In other words, $k(\x,\cdot)$ is a high-dimensional \emph{representer} of $\x$. Moreover, by the reproducing property it also acts as a \emph{representer of evaluation} of any function in $\hbspace$ on the data point $\x$. These properties of kernels are imperative in practical applications because if an algorithm can be formulated in terms of an inner product $\langle \x,\y\rangle$, one can construct an alternative algorithm by replacing the inner product by a positive definite kernel $k(\x,\y)$ without building the mapping of $\x$ and $\y$ explicitly (a.k.a. the \emph{kernel trick}). Well-known examples of kernelizable learning algorithms include the \emph{support vector machine (SVM)} \citep{Cortes1995} and \emph{principle component analysis (PCA)} \citep{Hotelling33:PCA}.

\begin{figure}[t!]
  \centering
  \begin{subfigure}[b]{0.32\textwidth}
    \includegraphics[width=\textwidth]{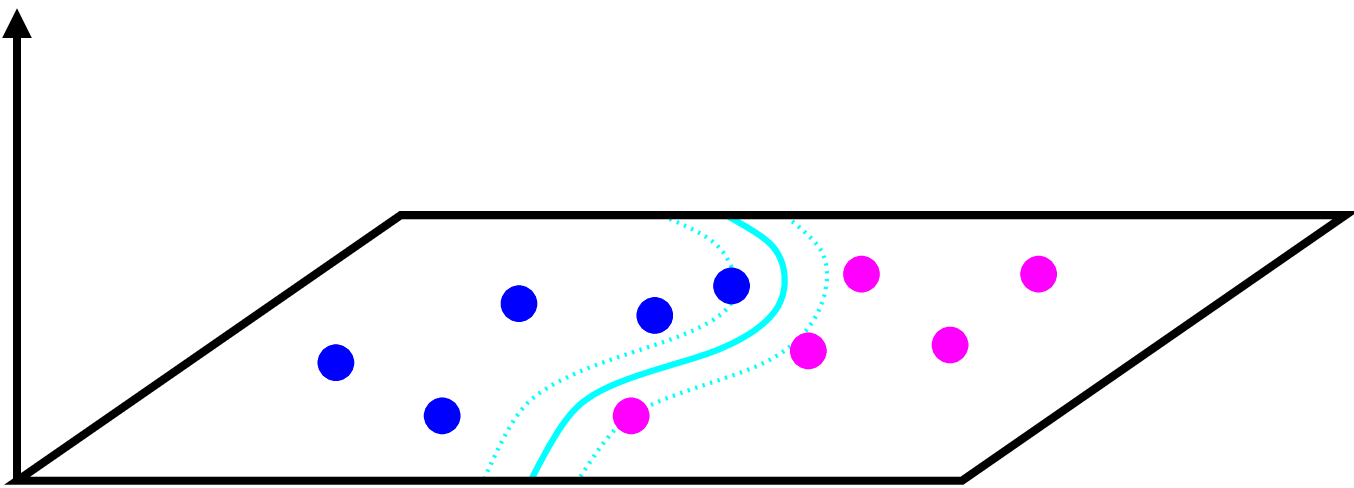}
    \caption{$\x \mapsto k(\x,\cdot)$}
    \label{subfig:data-point}
  \end{subfigure}
  \hfill
  \begin{subfigure}[b]{0.33\textwidth}
      \includegraphics[width=\textwidth]{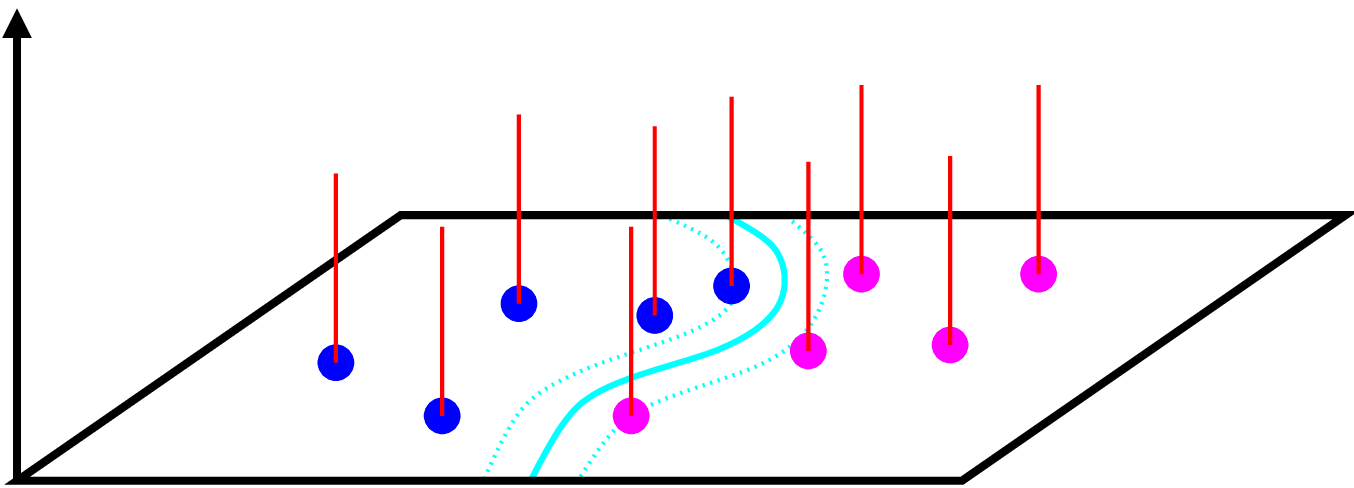}
      \caption{$\delta_{\x} \mapsto \int k(\y,\cdot) \dd\delta_{\x}(\y)$}
      \label{subfig:dirac-measure}
  \end{subfigure}
  \hfill
  \begin{subfigure}[b]{0.33\textwidth}
    \includegraphics[width=\textwidth]{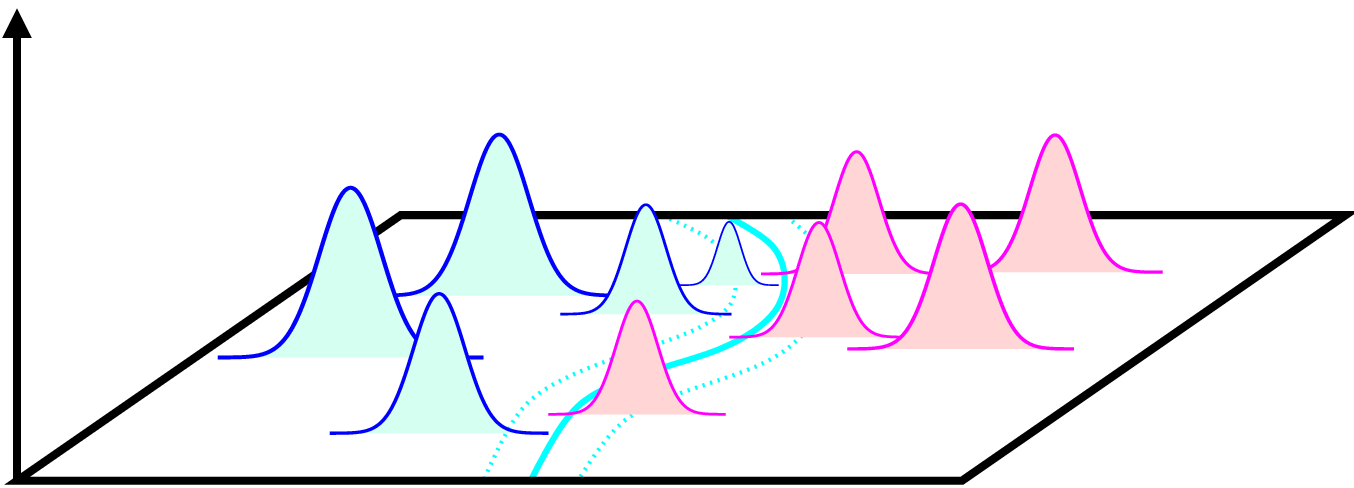}
    \caption{$\pp{P} \mapsto \int k(\x,\cdot) \dd\pp{P}(\x)$}
    \label{subfig:prob-measure}
  \end{subfigure}
  \caption{From data points to probability measures: (\subref{subfig:data-point}) An illustration of a  typical application of the kernel as a high-dimensional feature map of an individual data point. (\subref{subfig:dirac-measure}) A measure-theoretic view of high-dimensional feature map. An embedding of data point into a high-dimensional feature space can be equivalently viewed as an embedding of a Dirac measure assigning the mass 1 to each data point. (\subref{subfig:prob-measure}) Generalizing the Dirac measure point of view, we can generally extend the concept of a high-dimensional feature map to the class of probability measures.}
  \label{fig:from-point-to-measure}
\end{figure}

We can generalize the concept of a high-dimensional feature map of data points $\x\in\inspace$ to measures on a measurable space $(\inspace,\Sigma)$ where $\Sigma$ is a $\sigma$-algebra of subsets of $\inspace$. The simplest example of a measure is the Dirac measure $\delta_{\x}$ defined for $\x$ in $\inspace$ by
\begin{equation}
  \label{eq:dirac-measure}
  \delta_{\x}(A) = \begin{cases}
    1 & \text{if } \x\in A \\
    0 & \text{if } \x\notin A ,
    \end{cases}
\end{equation}
\noindent where $A\in\Sigma$. Since any measurable function $f$ on $\inspace$ is integrable \wrt $\delta_{\x}$, we have
\begin{equation}
  \label{eq:dirac-evaluation}
  \int f(\mathbf{t}) \dd\delta_{\x}(\mathbf{t}) = f(\x) .
\end{equation}
When $f$ belongs to the Hilbert space $\hbspace$ of functions on $\inspace$ with reproducing kernel $k$, we can rewrite \eqref{eq:dirac-evaluation} using the reproducing property of $\hbspace$ as
\begin{eqnarray}
  \label{eq:dirac-evaluation-2}
  \int f(\mathbf{t}) \dd\delta_{\x}(\mathbf{t}) &=& \int \langle f,k(\mathbf{t},\cdot)\rangle_\hbspace \dd\delta_{\x}(\mathbf{t}) \nonumber \\
  &=& \left\langle f,\int k(\mathbf{t},\cdot) \dd\delta_{\x}(\mathbf{t})\right\rangle_\hbspace
  = \langle f, k(\x,\cdot) \rangle_\hbspace.
\end{eqnarray}
As in the case of input space $\inspace$, the function $\int k(\mathbf{t},\cdot) \dd\delta_{\x}(\mathbf{t})$ acts as a representer of the measure $\delta_{\x}$ in the Hilbert space. Also, it may be viewed as a representer of evaluation of the following functional:
\begin{equation}
  f \longmapsto \int f(\mathbf{t}) \dd\delta_{\x}(\mathbf{t}),
\end{equation}
\noindent namely, the expectation of $f$ \wrt the Dirac measure $\delta_{\x}$. Although integrating $f$ \wrt $\delta_{\x}$ or evaluating $\langle f,k(\x,\cdot)\rangle_\hbspace$ gives 
the same result $f(\x)$, \ie, the value of $f$ at point $\x$, the
former gives a measure-theoretic point of view of the latter (see also
Figure \ref{fig:from-point-to-measure}). Consequently, we can define a feature map from the space of Dirac measures to $\hbspace$ as
\begin{equation}
  \label{eq:dirac-map}
  \delta_{\x} \longmapsto \int_{\inspace} k(\y,\cdot) \dd\delta_{\x}(\y) .
\end{equation}

Intuitively, the Dirac measure $\delta_{\x}$ is a probability measure
on $(\inspace,\Sigma)$ assigning the mass 1 to the set
$\{\x\}$. By virtue of \eqref{eq:dirac-map}, any kernel-based learning
algorithm such as SVMs and GPs can then be extended to operate on
a set of probability measures $\delta_{\x_1},\ldots,\delta_{\x_n}$
\citep{Muandet12:SMM}, \eg, by employing the kernel $K(\delta_{\x},\delta_{\x'}) = \langle  \int  k(\y,\cdot) \dd\delta_{\x}(\y),  \int k(\y,\cdot) \dd\delta_{\x'}(\y) \rangle_{\hbspace} = k(\x,\x')$. However, as we can see in
\eqref{eq:dirac-evaluation-2} this extension is not useful in
practice because \eqref{eq:feature-map} and \eqref{eq:dirac-map} are in fact equivalent. In what
follows, we will consider more interesting cases of non-trivial probability measures.

More generally, if $\x_1,\ldots,\x_n$ are $n$ distinct points in $\inspace$ and $a_1,\ldots,a_n$ are $n$ non-zero real numbers, we consider a linear combination
\begin{equation}
  \label{eq:finite-measure}
  \sum_{i=1}^na_i\delta_{\x_i}
\end{equation}
\noindent of Dirac measures putting the mass $a_i$ at the point $\x_i$. This is a \emph{signed} measure which constitutes a class of measures with finite support. A measure of the form \eqref{eq:finite-measure} is ubiquitous in the machine learning community, especially in Bayesian probabilistic inference \citep{Adams09:Thesis}. For example, if $a_i=1/n$ for all $i$, we obtain an \emph{empirical measure} associated with a sample $\x_1,\ldots,\x_n$. A Donsker measure is obtained when $a_i$ is also a random variable \citep{Berlinet04:RKHS}. Lastly, if $a_i=1$ for all $i$, the measure of the form \eqref{eq:finite-measure} represents an instance of a \emph{point process} on $\inspace$ which has numerous applications in Bayesian nonparametric inference and neural coding \citep{Dayan05:TNC}. For example, a determinantal point process (DPP)---a point process with a repulsive property---has recently gained popularity in the machine learning community \citep{Kulesza12:DPPML}.

Likewise, for any measurable function $f$ we have
\begin{equation}
  \int f \dd\left(\sum_{i=1}^na_i\delta_{\x_i}\right) = \sum_{i=1}^n a_i\int f \dd\delta_{\x_i} = \sum_{i=1}^na_if(\x_i) .
\end{equation}
This extends our previous remark on Dirac measures to measures with finite support, and if $f$ belongs to $\hbspace$, we obtain similar results to the case of Dirac measure. That is, the mapping
\begin{equation}
  \sum_{i=1}^na_i\delta_{\x_i} \longmapsto \sum_{i=1}^na_ik(\x_i,\cdot)
\end{equation}
\noindent gives a representer in $\hbspace$ of a measure with finite support. Furthermore, it is a representer of expectation \wrt the measure, \ie, if $\muv:=\sum_{i=1}^na_i\delta_{\x_i}$, we have for any $f$ in $\hbspace$
\begin{equation}
  \left\langle f,\sum_{i=1}^na_ik(\x_i,\cdot)\right\rangle_\hbspace = \sum_{i=1}^na_if(\x_i) = \int f \dd\muv .
\end{equation}
In particular, for any Hilbert space $\hbspace$ of functions on $\inspace$ with reproducing kernel $k$, a linear combination $\sum_{i=1}^na_ik(\x_i,\cdot)$ forms a dense subset of $\hbspace$. Here some readers may have concerns regarding the measurability of $f$. 
This is easily seen, however, since it is known that point-wise convergence of measurable functions gives a measurable function.


\begin{figure}[t!]
  \centering
  \begin{tikzpicture}[domain=0:6,scale=0.7]
      \coordinate (P1) at (3.4,1.6);
      \coordinate (Q1) at (3.8,2.4);
      \coordinate (P2) at (12,2);
      \coordinate (Q2) at (11,3);

      \draw[->] (-0.2,1) -- (6,1) node[right] {$\x$};
      \draw[->] (0,1) -- (0,4.2) node[above] {$p(\x)$};
      \draw[color=red,smooth,domain=0:5.5,ultra thick] plot (\x,{1+2*exp(-(\x-3)*(\x-3)*0.5/1)});
      \draw[color=blue,smooth,domain=0:5,ultra thick] plot (\x,{1+3*exp(-(\x-2)*(\x-2)*0.5/0.6)});

      \node [cloud, fill=green!50, cloud puffs=16, cloud puff arc= 100,
      minimum width=4cm, minimum height=2.5cm, aspect=1] at (12.5,2.5) {};
      \node [] at (12.5,4.6) {\textbf{RKHS} $\hbspace$};

      \draw[color=gray] (P1) to[out=20,in=170] (P2);
      \draw[color=gray] (Q1) to[out=20,in=170] (Q2);

      \node [fill=blue,circle,thick,minimum width=0.1cm] at (P2) {};
      \node [right] at (12.2,2) {$\muv_{\pp{P}}$};
      \node [fill=red,circle,thick,,minimum width=0.1cm] at (Q2) {};
      \node [right] at (11.2,3) {$\muv_{\pp{Q}}$};

      \node [right] at (2.5,4) {$\color{blue} \pp{P}$};
      \node [right] at (3.5,3) {$\color{red} \pp{Q}$};
    \end{tikzpicture}
  \caption{Embedding of marginal distributions: Each distribution is mapped into a \gls{rkhs} via an expectation operation.}
  \label{fig:kme-summary}
\end{figure}
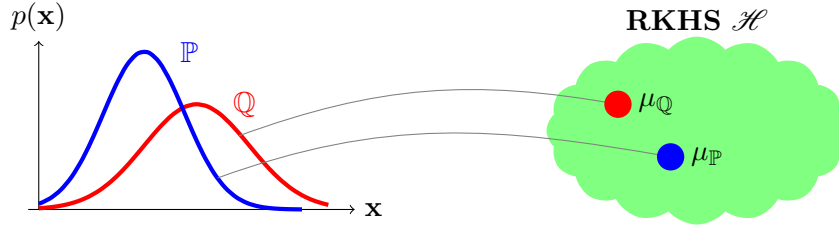

In what follows, we use $M_+^1(\inspace)$ to denote the space of probability measures over a measurable space $\inspace$. Then, we can define the representer in $\hbspace$ of any probability measure $\pp{P}$ through the mapping
\begin{equation}
  \label{eq:mean-embedding}
  \muv : M_+^1(\inspace) \longrightarrow \hbspace, \quad \pp{P} \longmapsto \int k(\x,\cdot) \dd\pp{P}(\x) ,
\end{equation}
\noindent which will be denoted by $\muv_{\pp{P}}$. Here the integral should be interpreted as a Bochner integral, which is in general an integral for Banach space valued functions.  See \citet[Chapter 2]{Diestel-77} and \citet[Chapter 1]{Dinculeanu:2000} for details. The above mapping is essentially the kernel mean embedding we consider throughout the survey. 

\begin{shadowbox}
\begin{definition}[{\citealp{Berlinet04:RKHS,Smola07Hilbert}}]
The kernel mean embedding of probability measures in $M_+^1(\inspace)$ into an RKHS $\hbspace$ endowed with a reproducing kernel $k:\inspace\times\inspace\rightarrow\rr$ is defined by a mapping
  \begin{equation*}
  \muv : M_+^1(\inspace) \longrightarrow \hbspace, \quad \pp{P} \longmapsto \int k(\x,\cdot) \dd\pp{P}(\x).
  \end{equation*}
\end{definition}
\end{shadowbox}

Next, we provide the conditions under which the embedding $\muv_{\pp{P}}$ exists and belongs to $\hbspace$.

\begin{shadowbox}
\begin{lemma}[\citealp{Smola07Hilbert}]
  \label{lem:kme-existence}
  If $\ep_{X\sim \pp{P}}[\sqrt{k(X,X)}] < \infty$, then $\muv_{\pp{P}}\in\hbspace$ and $\ep_{\pp{P}}[f(X)] = \langle f,\muv_{\pp{P}}\rangle_{\hbspace}$.
\end{lemma} 
\end{shadowbox}

\begin{proof}
  Let $\mathbf{L}_{\pp{P}}$ be a linear operator defined as $\mathbf{L}_{\pp{P}} f := \ep_{X\sim \pp{P}}[f(X)]$. We assume $\mathbf{L}_{\pp{P}}$ is bounded for all $f\in\hbspace$, \ie,
  \begin{eqnarray*}
    |\mathbf{L}_{\pp{P}} f| = |\ep_{X\sim \pp{P}}[f(X)]| &\stackrel{(*)}{\leq}& \ep_{X\sim \pp{P}}[|f(X)|] \\
    & = & \ep_{X\sim \pp{P}}[|\langle f, k(X,\cdot)\rangle_{\hbspace}|] \\
    & \leq & \ep_{X\sim \pp{P}}\left[\sqrt{k(X,X)}\|f\|_{\hbspace}\right] ,
  \end{eqnarray*}
  \noindent where we use Jensen's inequality in $(*)$. Hence, by the Riesz representation theorem (see Theorem \ref{thm:riesz-representation}), there exists $h \in\hbspace$ such that $\mathbf{L}_{\pp{P}}f = \langle f,h\rangle_{\hbspace}$. Choose $f=k(\x,\cdot)$ for some $\x\in\inspace$. Then, $h(\x) = \mathbf{L}_{\pp{P}}k(\x,\cdot) = \int k(\x,\x')\dd\pp{P}(\x')$ which means $h = \int k(\cdot,\x')\dd\pp{P}(\x') = \muv_{\pp{P}}$. 
\end{proof}



From the proof of Lemma \ref{lem:kme-existence}, $\ep_{X\sim \pp{P}}[f(X)] = \langle f, \muv_{\pp{P}}\rangle_{\hbspace}$ for any $f\in\hbspace$. This equality can essentially be viewed as a \emph{reproducing property} of the expectation operation in the \gls{rkhs}. That is, it allows us to compute the expectation of a function $f$ in the \gls{rkhs} \wrt the distribution $\pp{P}$ by means of an inner product between the function $f$ and the embedding $\muv_{\pp{P}}$. This property has proven useful in certain applications such as graphical models and probabilistic inference that require an evaluation of expectation \wrt the model \citep{Song10:HMM,Song11:KBP,Boots2013,McCalman2013}. This property can be 
extended to conditional distributions as well (see \S\ref{sec:conditional-embedding}).

\subsection{Explicit Representation of Mean Embeddings}

It is important to understand what information of the distribution is retained by the kernel mean embedding. For a linear kernel $k(\x,\x')=\langle\x,\x'\rangle$, it is clear that $\muv_{\pp{P}}$ becomes just the first moment of $\pp{P}$, whereas for the polynomial kernel $k(\x,\x')=(\langle\x,\x'\rangle+1)^2$ the mean map retains both the first and the second moments of $\pp{P}$. Below we provide some explicit examples which can also be found in, \eg, \citet{Smola07Hilbert,Fukumizu2008,Sriperumbudur10:Metrics,Gretton12:KTT,Schoelkopf15:KPP}.

\begin{shadowbox} 
\begin{example}[inhomogeneous polynomial kernel]
  \label{exp:polynomial}
  Consider the inhomogeneous polynomial kernel $k(\x,\x) =
  (\langle\x,\x'\rangle + 1)^p, \,\x,\x'\in\rr^d$ of degree $p$. Using 
  \begin{eqnarray*}
    (\langle\x,\y\rangle + 1)^p &=& 1 + \binom{p}{1}\langle\x,\y\rangle + \binom{p}{2}\langle\x,\y\rangle^2 + \binom{p}{3}\langle\x,\y\rangle^3 +  \cdots \\
    &=& 1 + \binom{p}{1}\langle\x,\y\rangle + \binom{p}{2}\langle\x^{(2)},\y^{(2)}\rangle \\
    && + \binom{p}{3}\langle\x^{(3)},\y^{(3)}\rangle +  \cdots
  \end{eqnarray*}
  \noindent where $\x^{(i)} := \bigotimes_{k=1}^i \x$ denotes the $i$th-order tensor product
  \citep[Proposition 2.1]{Scholkopf01:LKS}, the kernel mean embedding
  can be written explicitly as 
  \begin{eqnarray*}
    \muv_{\pp{P}}(\mathbf{t}) &=& \int (\langle\x,\mathbf{t}\rangle + 1)^p \dd \pp{P}(\x) \\
    &=& 1 + \binom{p}{1}\langle \mathbf{m}_{\pp{P}}(1),\mathbf{t}\rangle + \binom{p}{2}\langle \mathbf{m}_{\pp{P}}(2),\mathbf{t}^{(2)}\rangle \\
    && + \binom{p}{3}\langle \mathbf{m}_{\pp{P}}(3),\mathbf{t}^{(3)}\rangle +  \cdots ,
  \end{eqnarray*}
\noindent where $\mathbf{m}_{\pp{P}}(i)$ denotes the $i$th moment of the distribution $\pp{P}$. 
\end{example}
\end{shadowbox}

\vspace{10pt}
As we can see from Example \ref{exp:polynomial}, the embedding incorporates up to the $m$th moment of $\pp{P}$. As we increase $p$, more information about $\pp{P}$ is stored in the kernel mean embedding.

\begin{shadowbox}
\begin{example}[moment-generating function]
  \label{exp:mgf}
  Consider $k(\x,\x') = \exp(\langle \x,\x'\rangle)$. Hence, we can write the kernel mean embedding as
  \begin{equation*}
    \muv_{\pp{P}} = \ep_{X\sim\pp{P}}\left[ e^{\langle X, \cdot\rangle}\right] ,
  \end{equation*}
  \noindent which is essentially the \emph{moment-generating function} (MGF)
  of a random variable $X$ with distribution $\pp{P}$, given that the
  MGF exists.
\end{example}
\end{shadowbox}

The MGF is essentially the Laplace transformation of a random
variable $X$, which does not need to exist as it requires in particular the existence
of moments of any order. On the other hand, the characteristic function---which is
the Fourier transformation---always exists for any probability distribution $\pp{P}$.

\begin{shadowbox}
\begin{example}[characteristic function]
  \label{exp:characteristic-func}
  First, consider the Fourier kernel $k(\x,\x') = \exp(\sqrt{-1}\x^\top\x')$ using which we can express $\muv_{\pp{P}}$ as
  \begin{equation*}
    \muv_{\pp{P}}(\mathbf{t}) = \ep_{X\sim\pp{P}}[k(X,\mathbf{t})] = \ep_{X\sim\pp{P}}[\exp(\sqrt{-1} X^\top\mathbf{t})].
  \end{equation*}
  It is essentially the characteristic function of $\pp{P}$, which we
  denote by $\cf{\pp{P}}$. It is important to note that the Fourier kernel is not positive definite. As we see below, using $\cf{\pp{P}}$, it is
possible to represent the kernel mean embedding of $\pp{P}$ for any translation invariant kernel on $\mathbb{R}^d$.
  %

  Consider any translation invariant kernel $k(\x,\x') = \psi(\x-\x'), \,\x,\x'\in\rr^d$ where $\psi$ is a positive definite function. Bochner's theorem (see Theorem \ref{thm:bochner}) allows us to express the kernel mean embedding as
  \begin{eqnarray*}
    \muv_{\pp{P}}(\mathbf{t}) &=& \int_{\rr^d} \psi(\x - \mathbf{t}) \dd \pp{P}(\x) \\
    &=& \iint_{\rr^d}\exp\left( \sqrt{-1}\bm{\omega}^\top(\x-\mathbf{t})\right) \dd\Lambda(\bm{\omega})\dd \pp{P}(\x) \\
    &=& \iint_{\rr^d}\exp\left( \sqrt{-1}\bm{\omega}^\top \x \right) \dd \pp{P}(\x) \exp\left( -\sqrt{-1}\bm{\omega}^\top \mathbf{t} \right) \dd\Lambda( \bm{\omega}) \\
    &=& \int_{\rr^d} \cf{\pp{P}}(\bm{\omega})\exp\left( -\sqrt{-1}\bm{\omega}^\top \mathbf{t} \right) \dd\Lambda(\bm{\omega})
  \end{eqnarray*}
\noindent for some non-negative finite measure $\Lambda$ \citep{Sriperumbudur10:Metrics}. As shown in \eqref{eq:bochner-mean-map}, for $k(\x,\x') = \psi(\x-\x')$ we have $\langle\muv_{\pp{P}},\muv_{\pp{Q}}\rangle_{\hbspace} = \langle \cf{\pp{P}}, \cf{\pp{Q}} \rangle_{L_2(\rd{d},\borelms)}$.
\end{example}
\end{shadowbox}



\subsection{Empirical Estimation of Mean Embeddings}

In practice, we do not have access to the true distribution $\pp{P}$, and thereby cannot compute $\muv_{\pp{P}}$. Instead, we must rely entirely on the sample from this distribution. Given an i.i.d.~sample $\{\x_1,\ldots,\x_n\}$, the most common empirical estimate, denoted by $\muh_{\pp{P}}$ of the kernel mean $\muv_{\pp{P}}$ is
\begin{equation}
  \label{eq:emp-kmean}
  \muh_{\pp{P}} := \frac{1}{n}\sum_{i=1}^nk(\x_i,\cdot) . 
\end{equation}
Clearly, $\muh_{\pp{P}}$ is an unbiased estimate of $\muv_{\pp{P}}$,
and by the law of large numbers, $\muh_{\pp{P}}$ converges to 
$\muv_{\pp{P}}$ as $n\rightarrow\infty$. \citet{Sriperumbudur2012}
provides a thorough discussion on several properties of this
estimator. We will defer the details of kernel mean estimation and further discussion on some other estimators, \eg, shrinkage
estimators, to \S\ref{sec:kme}.

Some authors might also consider explicitly the mean map of the sample $\X=\{\x_1,\ldots,\x_n\}$. In contrast, one can also think of it as a special case of the mean embedding when the distribution is given by an empirical distribution associated with the sample $\X$, \eg, $\widehat{\pp{P}} := \frac{1}{n}\sum_{i=1}^n\delta_{\x_i}$. In this case, the mean embedding takes the form of an empirical estimate $\muh_{\pp{P}}$.

In summary, under suitable assumptions on the kernel $k$, the Hilbert-space embedding of distributions allows us to apply RKHS methods to probability measures. Throughout this section, we restrict our attention 
to the space of marginal distributions $\pp{P}(X)$, and defer an extension to the space of conditional distributions $\pp{P}(Y|X)$ to \S\ref{sec:conditional-embedding}.

\section{Covariance Operators}
\label{sec:cross-covariance}

In addition to the mean element, the \emph{covariance} and \emph{cross-covariance} operators on RKHSs are important concepts for modern applications of Hilbert space embedding of distributions. In principle, they are generalizations of covariance and cross-covariance matrices in Euclidean space to the infinite-dimensional elements in RKHSs. We give a brief review here; see \citet{Baker1973,Fukumizu04:DRS,Gretton16:RKHS} for further details.

Cross-covariance operators were introduced in \citet{Baker70:XCov} and then treated more extensively in \citet{Baker1973}. Let $(X,Y)$ be a random variable taking values on $\inx\times\iny$ and $(\hbspace,k)$ and $(\hbspg,l)$ be RKHSs with measurable kernels on $\inx$ and $\iny$, respectively. Throughout we assume 
\begin{equation}
\label{eq:asumption_cov}
  \ep_X[k(X,X)] \leq \infty, \quad   \ep_Y[l(Y,Y)] \leq \infty ,
\end{equation}
\noindent which ensures that $\hbspace\subset L^2(\pp{P}_X)$ and $\hbspg\subset L^2(\pp{P}_Y)$, since $\int f^2(x)\dd\pp{P}_X(x)=\int \langle f,k(\cdot,x)\rangle^2 \dd\pp{P}_X(x)\leq \int \|f\|^2 \|k(\cdot,X)\|^2 \dd\pp{P}_X \leq \|f\|^2 \ep_X[k(X,X)]$. A similar result also holds for $\int g(y)^2 \dd\pp{P}_Y(y)$. The (uncentered) \emph{cross-covariance operator} $\covyx:\hbspace\rightarrow\hbspg$ can be defined in terms of the tensor product $\varphi(Y)\otimes\phi(X)$ in a tensor product feature space $\hbspg\otimes\hbspace$ as
\begin{equation} 
  \label{eq:xcov-optr}
  \covyx := \ep_{\mathit{YX}}[\varphi(Y)\otimes\phi(X)] = \muv_{\pp{P}_{\mathit{YX}}},
\end{equation}
\noindent where $\pp{P}_{\mathit{YX}}$ denotes the joint distribution of $(X,Y)$ and $\phi$ (resp. $\varphi$) is the canonical feature map corresponding to $k$ (resp. $l$). The corresponding centered version of $\covyx$ is given by $\tilde{\mathcal{C}}_{\mathit{YX}} := \ep_{\mathit{YX}}[\varphi(Y)\otimes\phi(X)] - \muv_{\pp{P}_Y}\otimes\muv_{\pp{P}_X} = \muv_{\pp{P}_{\mathit{YX}}} - \muv_{\pp{P}_Y\otimes \pp{P}_X}$.

In the above definition, one needs to note that an operator can be identified with an element in the product space.  More precisely, the space of Hilbert-Schmidt operators ${\rm HS}(\hbspace,\hbspg)$ (Section \ref{sec:HSO}) forms Hilbert spaces which are isomorphic to the product space $\hbspace\otimes\hbspg$ given by the product kernel.  The isomorphism is defined by
\begin{eqnarray}
 \hbspace\otimes\hbspg & \to & {\rm HS}(\hbspace,\hbspg) ,\nonumber \\
\sum_i f_i\otimes g_i & \mapsto & \left[h\mapsto \sum_i \langle h,f_i\rangle_\hbspace g_i \right]. 
\label{eq:cov_HS}  
\end{eqnarray}  
Note that, given an orthonormal basis $\{\phi_a\}_a$ of $\hbspace$ and $\{\varphi_b\}_b$ of $\hbspg$, we have $\| \sum_i \langle \cdot,f_i\rangle_\hbspace g_i\|_{{\rm HS}}^2 = \sum_a\sum_b \{ \sum_i \langle \phi_a,f_i\rangle_\hbspace \langle \varphi_b,g_i\rangle_\hbspg\}^2 = \sum_a\sum_b \langle \sum_i f_i\otimes g_i, \phi_a\otimes \varphi_b \rangle_{\hbspace\otimes\hbspg}^2=\|\sum_i f_i\otimes g_i\|_{\hbspace\otimes\hbspg}^2$, where the last equality is based on the fact that $\{\phi_a\otimes\varphi_b\}_{a,b}$ is an orthonormal basis of $\hbspace\otimes\hbspg$.  This implies the above map is an isometry. 

To see that the cross-covariance operator $\covyx$ is well-defined as an Hilbert-Schmidt operator, it suffices from \eqref{eq:cov_HS} that $\| \ep_{\mathit{YX}}[\varphi(Y)\otimes\phi(X)] \|_{\hbspace\otimes\hbspg}<\infty$ is shown.  From
$\| \ep_{\mathit{YX}}[\varphi(Y)\otimes\phi(X)] \|_{\hbspace\otimes\hbspg}\leq \ep_{\mathit{YX}}\|\varphi(Y)\otimes\phi(X) \|_{\hbspace\otimes\hbspg} \leq \ep_{\mathit{YX}}[\sqrt{k(X,X)l(Y,Y)}]\leq \{ \ep_{\mathit{X}}[k(X,X)]\ep_{\mathit{Y}}[l(Y,Y)]\}^{1/2}$, the assumption \eqref{eq:asumption_cov} guarantees the existence.

Alternatively, we may define an operator $\covyx$ as a unique bounded operator that satisfies
\begin{equation*}
\langle g,\covyx f\rangle_{\hbspg} = \mathrm{Cov}[g(Y),f(X)]
\end{equation*}
\noindent for all $f\in\hbspace$ and $g\in\hbspg$. It can be shown using the Hilbert-Schmidt theory (see \S\ref{sec:HSO}) that---under the stated assumptions---this operator is of the form \eqref{eq:xcov-optr}. These two equivalent definitions stem from the relations between the covariance operator and mean element of the joint measure $\pp{P}_{\mathit{XY}}$ \citep{Baker1973}. To see that \eqref{eq:xcov-optr} leads to the result above, note that
\begin{eqnarray}
  \langle g,\covyx f\rangle_{\hbspg} &=& \langle \ep_{\mathit{YX}}[\varphi(Y)\otimes\phi(X)], g\otimes f\rangle_{\hbspg\otimes\hbspace} \nonumber \\
  &=& \ep_{\mathit{YX}}[\langle g\otimes f, \varphi(Y)\otimes \phi(X)\rangle_{\hbspg\otimes\hbspace}] \nonumber \\
  &=& \ep_{\mathit{YX}}[\langle g,\varphi(Y)\rangle_{\hbspg} \langle\phi(X),f\rangle_{\hbspace}] \nonumber \\
  &=& \ep_{\mathit{YX}}[g(Y)f(X)] =: \text{Cov}[g(Y),f(X)]. \label{eq:covxy2}
\end{eqnarray}
\noindent If $X=Y$, we call $\covx$ the \emph{covariance operator}, which is positive and self-adjoint, \ie, for all $f,h\in\hbspace$, $\langle \covx f, h\rangle_{\hbspace} = \langle f,\covx h\rangle_{\hbspace}$.

For any $f\in\hbspace$, we can also write the integral expressions for $\covyx$ and $\covx$, respectively, as
\begin{eqnarray*}
  (\covyx f)(\cdot) &=& \int_{\inspace\times\outspace} l(\cdot,\y)f(x)\dd\pp{P}_{\mathit{XY}}(\x,\y) \\
  (\covx f)(\cdot) &=& \int_{\inspace} k(\cdot,\x)f(\x)\dd\pp{P}_X(\x),
\end{eqnarray*}
\noindent where $\pp{P}_X$ denotes the marginal distribution of $X$ and $\pp{P}_{\mathit{XY}}$ denotes the joint distribution of $(X,Y)$. The first and second equations can be confirmed by plugging 
$g=l(\cdot,\y')$ into \eqref{eq:covxy2}.  Note that $\langle g,\covyx f\rangle_{\hbspg} = \langle \covxy g,f\rangle_{\hbspace}$. Hence, $\covxy$ is the adjoint of $\covyx$.

Below we outline a basic result that will become crucial for the definition of conditional mean embedding presented in Section \ref{sec:conditional-embedding}.

\begin{shadowbox}
\begin{theorem}[{\citealp{Fukumizu04:DRS}, \citealp[Theorem 1]{Fukumizu13:KBR}}]
  \label{thm:fukumizu:thm}
  If $\ep_{\mathit{YX}}[g(Y)|X=\cdot]\in\hbspace$ for $g\in\hbspg$, then
  \begin{equation*}
    \covx\ep_{\mathit{YX}}[g(Y)|X=\cdot] = \covxy g .
  \end{equation*}
\end{theorem}
\end{shadowbox} 

Given an i.i.d. sample $(\x_1,\y_1),\ldots,(\x_n,\y_n)$ on $\inx\times\iny$ drawn from $\pp{P}_{XY}$, an empirical estimate of the \emph{centered} $\covyx$ can be obtained as
\begin{eqnarray}
  \label{eq:emp-xcov}
  \ecovyx &=& \frac{1}{n}\sum_{i=1}^n\left\{l(\y_i,\cdot) - \muh_{\pp{P}_Y}\right\}\otimes\left\{k(\x_i,\cdot) - \muh_{\pp{P}_X}\right\} \nonumber \\
  &=& \frac{1}{n}\Psi\mathbf{H}\Phi^\top,
\end{eqnarray}
\noindent where $\mathbf{H}=\id_n-n^{-1}\mathds{1}_n$ is the centering matrix with $\mathds{1}_n$ an $n\times n$ matrix of ones, $\Phi = (\phi(\x_1),\ldots,\phi(\x_n))^\top$ and $\Psi=(\varphi(\y_1),\ldots,\varphi(\y_n))^\top$. The empirical covariance operator $\ecovx$ can be obtained in a similar way, \ie, $\ecovx = \frac{1}{n}\Phi\mathbf{H}\Phi^\top$. It has been shown that $\| \ecovyx - \covyx\|_{\text{HS}} = O_p(1/\sqrt{n})$ as $n\rightarrow\infty$ \citep{Berlinet04:RKHS}. Since $\covyx$ can be seen as an element in $\hbspace\otimes\hbspg$, this result is a consequence of the $\sqrt{n}$-consistency of the empirical mean $\muh_{\pp{P}}$ (cf. Theorem \ref{thm:kme-bound}) because the Hilbert-Schmidt norm of an operator from $\hbspace$ to $\hbspg$ corresponds to the norm in $\hbspace\otimes\hbspg$.

Let $\mathcal{N}(\mathcal{C})$ and $\mathcal{R}(\mathcal{C})$ denote the null space and the range of an operator $\mathcal{C}$. The following result due to \citet{Baker1973} states that the cross-covariance operator can be decomposed into the covariance of the marginals and the correlation.
\begin{shadowbox}
\begin{theorem}[{\citealp[Theorem 1]{Baker1973}}] 
  \label{thm:baker-thm}
  There exists a unique bounded operator $\mathcal{V}_{\mathit{YX}} : \hbspace\rightarrow\hbspf$, $\|\mathcal{V}\|\leq 1$, such that
  \begin{equation}
    \covyx = \covy^{1/2}\mathcal{V}_{\mathit{YX}}\covx^{1/2} ,
  \end{equation}
  \noindent where $\mathcal{R}(\mathcal{V}_{\mathit{YX}}) \subset \overline{\mathcal{R}(\covy)}$ and $\mathcal{N}(\mathcal{V}_{\mathit{YX}})^\perp \subset \overline{\mathcal{R}(\covx)}$.
\end{theorem}
\end{shadowbox}
\noindent The operator $\mathcal{V}_{\mathit{YX}}$ is often referred to as the \emph{normalized cross-covariance operator} and has been used as a basis for the conditional dependence measure. It is also  related to the canonical correlation.  See, \eg, \citet{Fukumizu07:KCCA,Fukumizu2008}. Compared to $\covyx$, $\mathcal{V}_{\mathit{YX}}$ captures the same information about the dependence of $X$ and $Y$, but with less influence of the marginal distributions.

The covariance operator serves as a basic building block in classical kernel-based methods such as kernel PCA \citep{Scholkopf98:KPCA,Zwald04:KPCA}, kernel Fisher discriminant, kernel CCA \citep{Fukumizu07:KCCA}, and kernel ICA \citep{Bach03:KICA}. More recent applications of the covariance operator include independence and conditional independence measures \citep{Gretton05:KIND,Zhang08:Dependence,Zhang2011,Doran2014}. See \S\ref{sec:dependency} and \S\ref{sec:cond-dependency} for more details.

Lastly, it is instructive to point out the connection between covariance operator $\covx$ and integral operator $\mathcal{T}_k$ defined in Theorem \ref{thm:mercer} w.r.t.~the measure $\pp{P}_X$. 
Since $\mathcal{T}_k^{1/2}$ is an isometry from $L_2(\inspace,\pp{P}_X)$ to $\hbspace$, one may define $\covx$ directly on $L_2(\inspace,\pp{P}_X)$ which happens to coincide with the operator $\mathcal{T}_k$. Hence, they have the same eigenvalues \citep{Hein04:Kernel,Rosasco10:Integral}. A detailed exposition on this connection is also given in Section 2.1 of \citet{Bach15:Quadrature}.

\section{Properties of the Mean Embedding} 
\label{sec:embedding-theory}

The following result, which appears in \citet[Proposition A.1]{Tolstikhin16:Minimax} which establishes the convergence of the empirical mean embedding $\muh_{\pp{P}}$ to the embedding of its population counterpart $\muv_{\pp{P}}$ in the \gls{rkhs} norm:\footnote{Similar result has also appeared in \citet[Theorem 2]{Smola07Hilbert}, \citet{Gretton12:KTT}, \citet{Lopez-Paz15:Towards} and  \citet[Theorem 27]{Song08:Thesis} but with worse constants.}

\begin{shadowbox}
\begin{theorem}
  \label{thm:kme-bound}
Let $k:\mathcal{X}\times\mathcal{X}\rightarrow\mathbb{R}$ be a continuous positive definite kernel on a separable topological space $\mathcal{X}$ with $\sup_{x\in\mathcal{X}} k(x,x)\le C_k < \infty$. Then for any $\delta\in(0,1)$ with probability atleast $1-\delta$,
  \begin{equation}
    \label{eq:kme-bound}
   \left\| \muh_{\pp{P}} - \muv_{\pp{P}} \right\|_{\hbspace} \leq
    \sqrt{\frac{C_k}{n}} + \sqrt{\frac{2C_k\log\frac{1}{\delta}}{n}} .
  \end{equation}
\end{theorem}
\end{shadowbox}
\noindent In other words, the convergence happens at the rate
$n^{-1/2}$. While various estimators have been studied for
$\muv_{\pp{P}}$ \citep{Muandet15:KMSE,Sriperumbudur16:Optimal}. \citet{Tolstikhin16:Minimax} recently showed that this rate is
minimax optimal and so the empirical estimator $\muh_{\pp{P}}$ is a minimax optimal
estimator of $\muv_{\pp{P}}$. Moreover,
$\sqrt{n}(\muh_{\pp{P}}-\muv_{\pp{P}})$ converges to a zero mean
Gaussian process on $\hbspace$ \citep[Section 9.1]{Berlinet04:RKHS}. 


\subsection{Characteristic and Universal Kernels}
\label{sec:universal-characteristic}


In this section we discuss the notion of the \emph{characteristic kernel}
which can be formally defined as follows.

\begin{shadowbox}
\begin{definition}
  \label{def:characteristic-kernel}
  A kernel $k$ is said to be characteristic if the map $\muv : \pp{P}
  \mapsto \muv_{\pp{P}}$ is injective. The \gls{rkhs} $\hbspace$ is said to be characteristic if its reproducing kernel is characteristic.
\end{definition}
\end{shadowbox}

A characteristic kernel is essential for kernel mean embedding because it ensures that $\|\muv_{\pp{P}} - 
\muv_{\pp{Q}}\|_{\hbspace}=0$ if and only if $\pp{P} = \pp{Q}$. In
other words, there is no information loss when mapping the
distribution into the Hilbert space. They were first introduced in \citet{Fukumizu04:DRS} being the kernel that 
satisfy Definition \ref{def:characteristic-kernel}.\footnote{The
  terminology \emph{probability determining} is used there.} \citet{Fukumizu2008} shows that Gaussian and Laplacian kernels are characteristic on $\rr^d$. The properties of characteristic 
kernels were explored further in \citet{Sriperumbudur08injectivehilbert,Sriperumbudur10:Metrics,Sriperumbudur2011}. 


A closely related notion is that of a universal kernel which we discuss its connection to a characteristic kernel below.

\begin{shadowbox}
\begin{definition}[Universal kernels, \citealp{Steinwart:2002:IKC}]
  \label{def:universal-kernel}
A continuous positive definite kernel $k$ on a compact metric space $\inspace$ is said to be universal if the corresponding RKHS $\hbspace$ is dense in $C_b(\inspace)$, \ie, for any $f\in C_b(\inspace)$ and $\varepsilon > 0$, there exists a function $h\in\hbspace$ such that $\| f-h \|_{\infty} < \varepsilon$.
\end{definition}
\end{shadowbox} 

\subsubsection{Characterization of Characteristic Kernels}

There are several characterizations of characteristic kernels. Intuitively, for the map $\pp{P}\mapsto\mu_{\pp{P}}$ to be injective, the RKHS endowed with the kernel $k$ should 
contain a sufficiently rich class of functions to represent all higher order moments of $\pp{P}$. \citet[Lemma 1]{Fukumizu2008}---see also \citet[Prop. 5]{Fukumizu2009:KDR}---shows 
that, for $q\geq 1$, if the sum\footnote{The sum of two RKHS's corresponds to the positive definite kernel given by the sum of the respective kernels.} $\hbspace_k+\rr$ is dense in 
$L^q(\inspace,\pp{P})$ for any probability $\pp{P}$ on
$(\inspace,\mathcal{A})$, the kernel $k$ is characteristic. As a
result, universal kernels on compact domains \citep{Steinwart:2002:IKC}
are characteristic as the associated RKHS is dense in $L^2(\mathcal{X},\pp{P})$ for
any $\pp{P}$; see also \citet[Theorem 3]{Gretton07:MMD}. For non-compact
spaces, \citet[Theorem 2]{Fukumizu2008} proves that the kernel
$k(\x,\y)=\psi(\x-\y)$ is characteristic, \ie, the RKHS associated with $k$ is dense in $L^2(\pp{P})$ for any Borel probability $\pp{P}$ on $\rr^d$, if for any $\xi\in\rr^d$ there exists $\tau_0$ such 
that its Fourier transform $\mathfrak{F}\psi$ satisfies
$\int\frac{(\mathfrak{F}\psi)(\tau(t+\xi))^2}{(\mathfrak{F}\psi)(t)}\dd t <
\infty$ for all $\tau > \tau_0$. Gaussian and Laplacian kernels
satisfy this condition, and hence are characteristic. As we can see,
it is required that $(\mathfrak{F}\varphi)(t) > 0$ for all $t$ for such an
integral to be finite. This property was studied further in
\citet{Sriperumbudur08injectivehilbert,Sriperumbudur10:Metrics,Sriperumbudur2011}
who showed that any translation-invariant kernel is characteristic if
the support of the Fourier transform of the kernel is the entire
$\rr^d$. Specifically, for $k(\x,\y) = \psi(\x-\y)$, it follows that
\begin{equation}
  \label{eq:fourier-shift}  
  \mu_{\pp{P}} = \ep_{\x\sim\pp{P}}[\psi(\x-\cdot)]= \psi\ast\pp{P} ,
\end{equation}
\noindent where $\ast$ denotes the convolution. Hence, $\mathfrak{F}(\psi\ast\pp{P}) =
\varphi_\pp{P} \cdot\mathfrak{F}\psi$ where $\varphi_\pp{P}$ denotes the characteristic function of $\pp{P}$.\footnote{Generally speaking, we may view 
\eqref{eq:fourier-shift} as an \emph{integral transform} of the distribution, \eg, Fourier and Laplace transforms. Thus, from the continuity of $\varphi_\pp{P}$, if the transform is 
everywhere positive, there will be a one-to-one correspondence between the distributions and their mean embeddings.} Since $\cf{\pp{P}}$ uniquely determines $\pp{P}$, so does $\mu_{\pp{P}}$ if $\mathfrak{F}\psi$ is everywhere positive. Later, it was shown 
in \citet{Sriperumbudur2009} that \emph{integrally strictly positive
  definite} kernels are characteristic. A measurable and 
  bounded kernel $k$ is said to be strictly positive definite on a 
  topological space $\mathcal{X}$ if and only if $\iint_{\mathcal{X}} 
  k(\x,\y) \dd\nu(\x)\dd\nu(\y) > 0$ for all finite non-zero signed
  Borel measures $\nu$ on $\mathcal{X}$. \citet{Fukumizu2009} also provides numerous examples of characteristic kernels on groups and semi-groups. Table \ref{tab:characteristic-kernels} summarizes various characterizations of well-known characteristic kernels.

Given a characteristic kernel $k$, we can generate new 
characteristic kernels through a \emph{conformal mapping}---\ie, a transformation that preserves angles locally---given by
$\tilde{k}(\x,\y) := f(\x)k(\x,\y)f(\y)$ for any bounded continuous
function $f$ such that $f(\x) > 0$ for all $\x\in\inspace$ and
$k(\x,\x)|f(\x)|^2$ is bounded \citep[Lemma 2]{Fukumizu2009}, see, also \citet{Wu02:Conformal}.

\begin{landscape}
\begin{table}  
  \centering  
  \caption{Various characterizations of well-known kernel functions. The columns marked
    \lq U\rq, \lq C\rq, \lq TI\rq, and \lq SPD\rq \; indicate whether the kernels are universal,
  characteristic, translation-invariant, and strictly positive definite, respectively, w.r.t. the domain $\inspace$. For the discrete kernel, $\#_s(\x)$ is the number of times substrings $s$ occurs in a string $\x$. $K_{\nu}$ is the modified Bessel function of the second kind of order $\nu$ and $\Gamma$ is the Gamma function.} 
  \label{tab:characteristic-kernels}
  \begin{adjustbox}{width=1.5\textwidth,totalheight=\textheight,keepaspectratio}
  \begin{tabular}{lllcccc} 
    \toprule 
    \textbf{Kernel Function} & $k(\x,\y)$ & \textbf{Domain} $\inspace$ & \textbf{U} & \textbf{C} & \textbf{TI} & \textbf{SPD} \\
    \midrule  
    Dirac & $\mathds{1}_{\x=\y}$ & $\{1,2,\ldots,m\}$ & \cmark & \cmark & \xmark & \cmark \\
    Discrete & $\sum_{s\in\inspace}w_s\#_s(\x)\#_s(\y)$ with $w_s > 0$ for all $s$ & $\{s_1,s_2,\ldots,s_m\}$ & \cmark & \cmark & \xmark & \cmark \\
    Linear & $\langle \x,\y\rangle$ & $\rr^d$ & \xmark & \xmark & \xmark & \xmark \\
    Polynomial & $(\langle \x,\y\rangle + c)^p$ & $\rr^d$ & \xmark & \xmark & \xmark & \xmark \\
    Gaussian & $\exp(-\sigma\|\x-\y\|^2_2), \;\sigma > 0$ & $\rr^d$ & \cmark & \cmark & \cmark & \cmark \\
    Laplacian & $\exp(-\sigma\|\x-\y\|_1), \;\sigma > 0$ & $\rr^d$ & \cmark & \cmark & \cmark & \cmark \\
    Rational quadratic & $(\|\x-\y\|_2^2 + c^2)^{-\beta}, \;\beta > 0, c > 0$ & $\rr^d$ & \cmark & \cmark & \cmark & \cmark \\
    $B_{2l+1}$-splines & $B_{2l+1}(\x - \y)$ where $l\in\mathbb{N}$ with $B_i := B_i\otimes B_0$ & $[-1,1]$ & \cmark & \cmark & \cmark & \cmark \\
    Exponential& $\exp(\sigma\langle\x,\y\rangle), \;\sigma>0$ & compact sets of $\rr^d$ & \xmark & \cmark & \xmark & \cmark \\
    Mat\'ern &  $\frac{2^{1-\nu}}{\Gamma(\nu)}\left(\frac{\sqrt{2\nu}\|\x-\x'\|_2}{\sigma}\right)K_\nu\left(\frac{\sqrt{2\nu}\|\x-\x'\|_2}{\sigma}\right)$ & $\rr^d$ & \cmark & \cmark & \cmark & \cmark \\ 
    Poisson  & $1/(1-2\alpha\cos(\x-\y) + \alpha^2), \; 0<\alpha<1$ & $([0,2\pi),+)$ & \cmark & \cmark & \cmark & \cmark \\
    \bottomrule  
  \end{tabular}
  \end{adjustbox} 
\end{table} 
\end{landscape} 

The richness of RKHS was previously studied through the so-called \emph{universal} kernels \citep{Steinwart:2002:IKC}.  Recall that a continuous positive definite kernel $k$ on a compact metric space 
$\inspace$ is said to be universal if the corresponding \gls{rkhs} $\hbspace$ is dense in the space of bounded continuous 
functions on $\inspace$ (Definition \ref{def:universal-kernel}). It implies 
that any kernel-based learning algorithms with universal kernels can in principle approximate any bounded continuous function $f$ arbitrarily well. Note that approximation in the 
RKHS norm implies the approximation in the sup norm by the continuity of the evaluation functional. It follows from Definition \ref{def:characteristic-kernel} and \citet[Theorem
8]{Gretton12:KTT} that all universal kernels are characteristic. Examples of
universal kernels on a compact domain are Gaussian and Laplace
kernels. On a discrete domain $\inspace=\{\x_1,\ldots,\x_n\}$, any
strictly positive definite kernel, \eg, $k(\x,\x') = \mathds{1}_{\{\x=\x'\}}$, is
universal \citep[Section 2.3]{Borgwardt06:MMD}. Overall, universality is
stronger than the characteristic property, \ie, all universal kernels are characteristic, but not vice versa. Nevertheless, they match if the
kernel is translation invariant, continuous, and decays to zero at
infinity. \citet{Sriperumbudur2011} provides a comprehensive insight into this connection between
universal and characteristic kernels. 


In non-Euclidean spaces, sufficient and necessary conditions for the characteristic RKHS on groups and semi-groups---\eg, periodic domains, rotation matrices, and $\rr^d_+$---are established in \citet{Fukumizu2009}. \citet{Christmann10:Kernels} studies universal kernels on non-Euclidean spaces and discusses the connection to characteristic kernels. Later, \citet{Nishiyama14:Chrac} establishes a connection between translation invariant kernels on $\rr^d$ and the infinitely divisible distributions. That is, the kernel is characteristic if it is generated by the bounded continuous density of a symmetric and infinitely divisible distribution.

For applications involving statistical hypothesis testing, a characteristic kernel is essential because it ensures that we obtain the right test statistics asymptotically, although in practice we only have access to a finite sample. However, there are other application domains---for instance, predictive learning on distributional data \citep{Muandet12:SMM,Muandet13:OCSMM,Oliva14:FastD2R,Zoltan15:DistReg}---in which the uniqueness of the embeddings is not strictly required, and thereby the choice of kernel functions can be less restrictive. In such applications, it is better to interpret kernel $k$ as a \emph{filter function} which determines which frequency component of the corresponding characteristic functions appears in the embeddings, see, \eg, Example \ref{exp:characteristic-func}. As a result, the shape of the kernel function in the Fourier domain can sometimes be more informative when choosing kernel functions.

 
\begin{paragraph}{Non-characteristic kernels.}
  In general, if the kernel $k$ is \emph{non}-characteristic, the embedding $\mu$ forms an equivalence class of distributions that correspond to the same mean embedding $\mu_{\pp{P}}$. Nevertheless, $k$ may be characteristic for a more restricted class of distributions. Consider, for example, any translation-invariant kernel with the corresponding $\Lambda$ whose support has a non-empty interior. Then, we may conclude from \eqref{eq:fourier-shift} that $k$ will be ``characteristic'' for any class of probability measures whose characteristic functions only agree outside the support of $\Lambda$. \citet[Theorem 12]{Sriperumbudur10:Metrics} and \citet[Proposition 3]{Harmeling2013} consider a more interesting class, namely, a class of probability measures with compact support on $\rr^d$. It follows from the Paley-Wiener theorem \citep{Rudin91:FA} that the characteristic functions of such measures are entire functions on $\mathbb{C}^d$. As a result, if $\Lambda$ has support with non-empty interior, the corresponding kernel will be characteristic for probability measures with compact support. Examples of kernels with this property include the sinc kernel $k(\x,\x')=\psi(\x-\x')=\frac{\sin \sigma(\x-\x')}{\x-\x'}$ with $\Lambda(\bm{\omega}) = \sqrt{\frac{\pi}{2}}\mathds{1}_{[-\sigma,\sigma]}(\bm{\omega})$.
\end{paragraph}
   
\section{Kernel Mean Estimation and Approximation}
\label{sec:kme}

In practice, the distribution $\pp{P}$ is generally unknown, and we must rely entirely on the sample drawn from $\pp{P}$. Given an independent and identically distributed (i.i.d.) sample $\x_1,\ldots,\x_n$ from $\pp{P}$, the standard estimator of the kernel mean is an empirical average
\begin{equation}
  \label{eq:emp-estimator}
  \hat{\mu}_{\pp{P}} := \frac{1}{n}\sum_{i=1}^nk(\x_i,\cdot) \,.
\end{equation}
By the weak law of large numbers, the empirical estimator \eqref{eq:emp-estimator} is guaranteed to converge to the true mean embedding. \citet{Berlinet04:RKHS} shows that the convergence happens at a rate $O_p(n^{-1/2})$. This result has also been reported in \citet{Smola07Hilbert}, \citet[Chapter 4]{Shawe04:KMPA}, \citet[Theorem 27]{Song08:Thesis}, \citet{Gretton12:KTT}, \citet[Theorem 1]{Lopez-Paz15:Towards} and \citet[Proposition A.1]{Tolstikhin16:Minimax}, in slightly different forms.

One may argue that the estimator \eqref{eq:emp-estimator} is the ``best'' possible estimator of $\mu_{\pp{P}}$ if nothing is known about the underlying distribution $\pp{P}$. In fact, \eqref{eq:emp-estimator} is minimax in the sense of \citet[Theorem 25.21, Example 25.24]{Vaart-98}. 
Nevertheless, given that a kernel mean is central to kernel methods in that it is used by many classical algorithms such as kernel principal component analysis (PCA), and it also forms the core inference step of modern kernel methods that rely on embedding probability distributions in RKHSs, it is compelling to ask whether the estimation of $\mu_{\pp{P}}$ can be 
improved.



\subsection{Kernel Mean Shrinkage Estimators}

Inspired by the James-Stein estimator of the mean of multivariate Gaussian distribution \citep{Stein55:Inadmissible,Stein61:JSE}, \citet{Muandet2014,Muandet15:KMSE} proposed a shrinkage estimator of the kernel mean which has the form
\begin{equation}
  \label{eq:shrinkage-est}
  \hat{\mu}_{\alpha} := \alpha f^* + (1-\alpha)\hat{\mu}_{\pp{P}},  
\end{equation}
\noindent for some $f^*\in\hbspace$ which is independent of the sample. The shrinkage parameter $\alpha$ specifies the amount by which the estimator $\muh_{\pp{P}}$ is shrunk towards $f^*$. 
Note that the works of \citet{Muandet2014,Muandet15:KMSE} differ from the Stein's seminal work and those along this line. That is, the setting of \citet{Muandet15:KMSE} 
involves a non-linear feature map $\phi$ associated with the kernel $k$. Consequently, the resulting kernel mean $\mu_{\pp{P}}$ may incorporate higher moments of $\pp{P}$. When $k$ is 
a linear kernel, $\mu_{\pp{P}}$ becomes a mean of $\pp{P}$ and this setting coincides with that of Stein. A direct generalization of the James-Stein estimator to an infinite dimensional 
Hilbert space has been considered in
\citet{Berger83:GP-Stein,Privault08:GP-malliavin} and \citet{Mandelbaum87:admissibility}.\footnote{The effect
  of shrinkage in the context of kernel mean embedding has previously
  been observed in the experimental study of \citet{Huszar2012} that
  investigates the connection between kernel Herding and Bayesian
  quadrature. In Bayesian quadrature, there is an indication that the
  Bayesian weight obtained by minimizing the posterior variance
  exhibits shrinkage when the sample size is small.} 


To understand the effect of the shrinkage estimator in \eqref{eq:shrinkage-est}, consider the bias-variance decomposition
\begin{equation}
  \label{eq:bias-variance}
  \ep\left\| \muh_{\alpha} - \muv_{\pp{P}}\right\|^2_{\hbspace} =
  \alpha^2\|f^*-\muv_{\pp{P}}\|^2_{\hbspace} + (1-\alpha)^2\ep[\| \muh_{\pp{P}}
  - \muv_{\pp{P}} \|^2_{\hbspace}], 
\end{equation} 
\noindent where the expectation is taken w.r.t. the i.i.d. sample of size $n$ from $\pp{P}$. The first term on the r.h.s. of \eqref{eq:bias-variance} represents the squared-bias, 
whereas the second term represents the variance. Note that 
for $\alpha \in (0,2)$, 
$(1-\alpha)^2 < 1$, which means 
the variance 
of $\muh_{\alpha}$ is always smaller than that of $\muh_{\pp{P}}$ at the expense of the increased bias. Hence, $\alpha$ which controls the bias-variance trade-off can be chosen to be the minimizer
of the mean-squared error. 
However, since this minimizer will depend on the unknown $\muv_{\pp{P}}$, \citet{Muandet15:KMSE} consider an estimate of $\alpha$ which when plugged in (\ref{eq:emp-estimator}) yields 
an estimate of $\muv_{\pp{P}}$. 
\citet{Muandet15:KMSE} also propose the positive-part version of \eqref{eq:shrinkage-est}, \ie, $\hat{\mu}_{\alpha} := \alpha f^* + (1-\alpha)_+\hat{\mu}_{\pp{P}}$, which is similar in spirit to the positive-part James-Stein estimator, where $(a)_+:=\max(0,a)$.

It is known that the empirical estimator \eqref{eq:shrinkage-est} can
be considered as an \emph{M-estimator} \citep{Shawe04:KMPA,Kim12:RKDE}, \ie, it can be obtained as
\begin{equation}
  \label{eq:regression-pers}
  \hat{\mu}_{\pp{P}} = {\arg\min}_{g\in\hbspace} \; \frac{1}{n}\sum_{i=1}^n\| g - k(\x_i,\cdot)\|_{\hbspace}^2 .
\end{equation}
The population counterpart $\mu_{\pp{P}}$ can be obtained by replacing the sum with the expectation. Owing to this interpretation, \citet{Muandet15:KMSE} constructs a shrinkage estimator by adding a regularizer 
$\Omega(\|g\|_\hbspace) = \lambda\|g\|^2_{\hbspace}$ to \eqref{eq:regression-pers} whose minimizer is given by $\left(\frac{1}{1+\lambda}\right)\muh_{\pp{P}}$.
It can be seen that this estimator is the same as the one in 
\eqref{eq:shrinkage-est} when $\alpha = \frac{\lambda}{1+\lambda}$ and $f^*=0$. \citet{Muandet15:KMSE} propose a data-dependent choice for $\lambda$ using leave-one-out cross-validation, 
resulting in a shrinkage estimator for $\muv_{\pp{P}}$. As an aside, \citet{Kim12:RKDE} also exploit this interpretation to robustify the kernel density estimator (KDE) by replacing the squared loss $L(k(\x_i,\cdot),g) = \|g-k(\x_i,\cdot)\|^2_{\hbspace}$ in \eqref{eq:regression-pers} by the loss which is less sensitive to outliers such as Huber loss \citep{Huber1964} and $k$ is assumed to be non-negative and to integrate to one, \eg, Gaussian kernel and Student's-\emph{t} kernel.

Later, \citet{Muandet2014a} provides a \emph{non-linear} extension of the estimator in \eqref{eq:shrinkage-est} by means of spectral filtering algorithms \citep{Bauer07:Regularization,Vito061:Spectral}. By the representer theorem \citep{Scholkopf01:GRT}, any minimizer of \eqref{eq:regression-pers} can be written as $g=\sum_{i=1}^n\beta_ik(\x_i,\cdot)$ for some $\bvec\in\rr^d$. Hence, finding $g$ amounts to solving a system of equations $\kmat\bvec = \kmat\mathbf{1}_n$. Using spectral filtering algorithms, \citet{Muandet2014a} proposes the following estimators:
\begin{equation}
  \label{eq:spectral-kmse} 
  \muh_{\lambda} := \sum_{i=1}^n\beta_i k(\x_i,\cdot), 
  \qquad \bvec := g_{\lambda}(\kmat)\kmat\mathbf{1}_n,
\end{equation}
\noindent where $g_\lambda$ is a matrix-valued filter function.  This function can be described in terms of a scalar function on the eigenspectrum of $\kmat$. That is, let $\kmat = \mathbf{U}\mathbf{D}\mathbf{U}^\top$ be the eigendecomposition where $\mathbf{D} = \mathrm{diag}(\gamma_1,\ldots,\gamma_n)$ is a diagonal matrix whose diagonal entries are eigenvalues $\gamma_i$ and $\mathbf{U} = [\mathbf{u}_1,\ldots,\mathbf{u}_n]$ is a matrix whose columns consist of eigenvectors $\mathbf{u}_i$. Then, we have $g_\lambda(\mathbf{D}) = \mathrm{diag}(g_{\lambda}(\gamma_1),\ldots,g_{\lambda}(\gamma_n))$ and $g_\lambda(\kmat) = \mathbf{U}g_{\lambda}(\mathbf{D})\mathbf{U}^\top$. Table \ref{tab:filter-func} shows examples of well-known filter functions. There exist efficient algorithms, \eg, Landweber iteration and iterated Tikhonov,  for solving \eqref{eq:spectral-kmse} which do not require an explicit eigendecomposition of $\kmat$. See, \eg, \citet{Heinz96:RegInverse,Bauer07:Regularization,Vito061:Spectral,Muandet2014a} for detail of each algorithm.

It was shown that the estimators of \citet{Muandet2014a} performs shrinkage by first projecting data onto the KPCA basis, and then shrinking each component independently according to a pre-defined shrinkage rule, which is specified by the filter function $g_{\lambda}$ \citep[Proposition 3]{Muandet2014a}. The shrinkage estimate is then reconstructed as a superposition of the resulting components. Unlike \eqref{eq:shrinkage-est}, the spectral shrinkage estimators \eqref{eq:spectral-kmse} also take the eigenspectrum of the kernel $\kmat$ into account. The idea has been extended to estimate the covariance operator \citep{Muandet15:KMSE,Wehbe15:ShrinkageHSIC} which is ubiquitous in kernel independence and conditional independence tests (cf. \S\ref{sec:dependency} and \S\ref{sec:cond-dependency}). \citet{Flaxman16:BayesKME} proposes a Bayesian model for kernel mean embedding using a GP prior over the RKHS containing the mean embedding. While the posterior mean of the model resembles the \emph{frequentist} spectral KMSE \eqref{eq:spectral-kmse} when $g_{\lambda}(\kmat) = (\kmat + n\lambda \id)^{-1}$, a closed-form uncertainty estimate leads to a principled method for learning kernel parameters with the marginal pseudo-likelihood.

Several ways to improve the use of kernel mean embedding have been investigated from the regularization perspective. In an attempt to improve the overall performance of the hypothesis testing, \citet{Harchaoui07:Homogeneity} proposed the two-sample test based on kernel FDA with Tikhonov-type regularization of the covariance operator, like the spectral KMSE. It was subsequently applied to change-point detection \citep{Harchaoui09:KCP} and audio segmentation \citep{Harchaoui09:KASEC}. \citet{Harchaoui09:KASEC} also considers a spectral filtering approach. Furthermore, \citet{Danafar13:RMMD} used the RKHS norm of the mean embeddings to directly regularize the distance between them, resulting in estimates which are similar to \eqref{eq:shrinkage-est}. As we can see, there is a fundamental link between Stein estimation in statistics and Tikhonov regularization in inverse problems.

\begin{table}
  \centering
  \caption{The iterative update for $\bvec$ and associated filter function (see, \eg, \citealp{Heinz96:RegInverse,Vito061:Spectral,Muandet2014a} for further details). Here we define $\z := \kmat\mathbf{1}_n-\kmat\bvec^{t-1}$. We let $\eta$ be the step-size parameter, $\lambda$ be a non-negative shrinkage parameter, and $p_t(x)$ be a polynomial of degree $t-1$.}
  \resizebox{\textwidth}{!}{
  \begin{tabular}{lll}
    \toprule
    Algorithm & Iterative update & Filter function \\
    \midrule
    L2 Boosting & $\bvec^t\leftarrow\bvec^{t-1} + \eta\z$ & $g(\gamma) = \eta\sum_{i=1}^{t-1}(1-\eta\gamma)^i$ \\
    Acc. L2 Boosting & $\bvec^t\leftarrow\bvec^{t-1} + \omega_t(\bvec^{t-1}-\bvec^{t-2}) + \frac{\kappa_t}{n}\z$ & $g(\gamma) = p_t(\gamma)$ \\
    Iterated Tikhonov & $(\kmat + n\lambda\id)\bvec_i = \mathbf{1}_n + n\lambda\bvec_{i-1}$ & $g(\gamma) = \frac{(\gamma+\lambda)^t-\gamma^t}{\lambda(\gamma+\lambda)^t}$ \\
    Truncated SVD & None & $g(\gamma) = \gamma^{-1}\mathds{1}_{\gamma\geq\lambda}$ \\
    \bottomrule
  \end{tabular}}
  \label{tab:filter-func}
\end{table}

\subsection{Approximating the Kernel Mean}
\label{sec:kme-approximation}
 
In many applications of kernel methods such as computational biology \citep{Schoelkopf04:CompBio}, the computational cost may be a critical issue, especially in the era of ``big data''. Traditional kernel-based algorithms have become computationally prohibitive as the volume of data has exploded because most existing algorithms scale at least quadratically with sample size. Likewise, the use of kernel mean embedding has also suffered from this limitation due to two fundamental issues. First, any estimators of the kernel mean involve the (weighted) sum of the feature map of the sample. Second, for certain kernel functions such as the Gaussian RBF kernel, the kernel mean lives in an infinite dimensional space. We can categorize previous attempts in approximating the kernel mean into two basic approaches: 1) find a smaller subset of samples whose estimate approximates well the original estimate of the kernel mean, 2) find a finite approximation of the kernel mean directly.

The former has been studied extensively in the literature. For example, \citet{Cortes2014} considers the problem of approximating the kernel mean as a sparse linear combination of the sample. The proposed algorithm relies on a subset selection problem using a novel incoherence measure. The algorithm can be solved efficiently as an instance of the \emph{k-center problem} and has linear complexity in the sample size. Similarly, \citet{Grunewalder12:LGBPP} proposes a sparse approximation of the conditional mean embedding by relying on an interpretation of the conditional mean as a regressor. Note that the same idea can be adopted to find a sparse approximation of the standard kernel mean by imposing the sparsity-inducing norm on the coefficient $\bvec$, \eg, $\|\bvec\|_1$ \citep{Muandet2014}. An advantage of sparse representation is in applications where the kernel mean is evaluated repeatedly, \eg, the Kalman filter \citep{Kanagawa2013,McCalman2013}. One of the drawbacks is that it needs to solve a non-trivial optimization problem in order to find an optimal subsample.

An alternative approach to kernel mean approximation is to find a finite representation of the kernel mean directly. One of the most effective approaches depends on the \emph{random feature map} \citep{Rahimi2007}. That is, instead of relying on the implicit feature map provided by the kernel, the basic idea of random feature approximation is to explicitly map the data to a low-dimensional Euclidean inner product space using a randomized feature map $\mathbf{z}:\rr^d\rightarrow\rr^m$ such that
\begin{equation} 
  k(\x,\y) = \langle \phi(\x),\phi(\y) \rangle_{\hbspace} \approx \mathbf{z}(\x)^\top\mathbf{z}(\y) , 
\end{equation}
\noindent where $\mathbf{z}(\x) := \mathbf{W}^\top\x$ and $w_{ij}\sim p(\w)$. If elements of $\mathbf{W}$ are drawn from an appropriate distribution $p(\w)$, the Johnson-Lindenstrauss Lemma \citep{Dasgupta2003,Blum2005} ensures that this transformation will preserve similarity between data points. In \citet{Rahimi2007}, $p(\w)$ is chosen to be the Fourier transform of translation-invariant kernels $k(\x,\y) = \psi(\x-\y)$. Given a feature map $\mathbf{z}$, a finite approximation of the kernel mean can be obtained directly as
\begin{equation}
  \label{eq:kme-approximation}
  \tilde{\muv}_{\pp{P}} = \frac{1}{n}\sum_{i=1}^n \mathbf{z}(\x_i) \in \rr^m.
\end{equation}
Since $\mathbf{z}(\x_i)\in\rr^m$ for all $i$, so does $\tilde{\muv}_{\pp{P}}$. Hence, there is no need to store all the vectors $\mathbf{z}(\x_i)$. In addition to giving us a compact representation of kernel mean, these randomized feature maps accelerate the evaluation of the algorithms that use kernel mean embedding (see, \eg, \citealp{Kar2012,Le2013,Pham2013} and references therein for extensions). Note that the approximation \eqref{eq:kme-approximation} is so general that it can be obtained as soon as one knows how to compute $\mathbf{z}(\x)$. Other approaches such as low-rank approximation are also applicable. As we can see, the advantage of this approach is that given any finite approximation of $\phi(\x)$, it is easy to approximate the kernel mean. Moreover, the resulting approximation has been shown to enjoy good empirical performance. The downside of this approach is that the random features are limited to only a certain class of kernel functions.

\section{Maximum Mean Discrepancy} 
\label{sec:MMD}

The kernel mean embedding can be used to define a metric for probability distributions which is important for problems in statistics and machine learning. Later, we will see that the metric defined in terms of mean embeddings can be considered as a particular instance of an \emph{integral probability metric} (IPM) \citep{Mueller1997:IPM}. Given two probability measures $\pp{P}$ and $\pp{Q}$ on a measurable space $\inspace$, an IPM is defined as
\begin{equation}
  \label{eq:ipm}
  \gamma[\fspace,\pp{P},\pp{Q}] = \sup_{f\in\fspace}\left\{\int f(\x)\dd\pp{P}(\x) - \int f(\y)\dd\pp{Q}(\y)\right\} ,
\end{equation}
\noindent where $\fspace$ is a space of real-valued bounded measurable functions on $\inspace$. The function class $\fspace$ fully characterizes the IPM $ \gamma[\fspace,\pp{P},\pp{Q}]$. There is obviously a trade-off on the choice of $\fspace$. That is, on one hand, the function class must be rich enough so that $\gamma[\fspace,\pp{P},\pp{Q}]$ vanishes if and only if $\pp{P}=\pp{Q}$. On the other hand, the larger the function class $\fspace$, the more difficult it is to estimate $\gamma[\fspace,\pp{P},\pp{Q}]$. Thus, $\fspace$ should be restrictive enough for the empirical estimate to converge quickly (see, \eg, \citealp{Sriperumbudur2012}).

For example, if $\fspace$ is chosen to be a space of all bounded continuous functions on $\inspace$, the IPM is a metric over a space of probability distributions, as stated in the following theorem \citep{Mueller1997:IPM}.
\begin{shadowbox}
\begin{theorem} 
A mapping $\gamma : \pspace\times\pspace \rightarrow [0,\infty]$ given by
$$\gamma[C_b(\inx),\pp{P},\pp{Q}] = \sup_{f\in C_b(\inx)}\left\{ \int f(\x)\dd\pp{P}(\x) - \int f(\y)\dd\pp{Q}(\y)\right\}$$
is zero if and only if $\pp{P}=\pp{Q}$.
\end{theorem} 
\end{shadowbox}

\noindent Unfortunately, it is practically difficult to work with $C_b(\inx)$. A more restrictive function class is often used. For instance, let $\fspace_{\text{TV}} = \{f\,|\,\|f\|_{\infty}\leq 1\}$ where $\|f\|_\infty = \sup_{\x\in\inspace}|f(\x)|$. Then, $\gamma[\fspace_{\text{TV}},\pp{P},\pp{Q}] = \|\pp{P}-\pp{Q}\|_1$ is the \emph{total variation distance}. If $\fspace=\{\mathbf{1}_{(-\infty,t]}:t\in\mathbb{R}\}$, we get the \emph{Kolmogorov (or $L^\infty$) distance} between distributions, which is the max norm of the difference between their cumulative distributions. If $\|f\|_{L} := \sup\{|f(\x)-f(\y)|/\rho(\x,\y),\x\neq \y\in\inspace\}$ is the Lipschitz semi-norm of a real-valued function $f$ where $\rho$ is some metric on $\inspace$, setting $\fspace_{\text{W}} = \{f\,|\, \|f\|_L\leq 1 \}$ yields the \emph{earthmover distance}, which we denote by $\gamma[\fspace_{\text{W}},\pp{P},\pp{Q}]$. In mathematics, this metric is known as the  \emph{Wasserstein (or $L^1$) distance}. 

When the supremum in \eqref{eq:ipm} is taken over functions in the unit ball in an \gls{rkhs} $\hbspace$, \ie, $\fspace:=\{f\,|\,\|f\|_{\hbspace}\leq 1\}$, the resulting metric is known as the \emph{maximum mean discrepancy} (MMD), which we will denote by $\text{MMD}[\hbspace,\pp{P},\pp{Q}]$. As shown in \citet[Lemma 4]{Borgwardt06:MMD,Gretton12:KTT}, the MMD can be expressed as the distance in $\hbspace$ between mean embeddings. That is,
\begin{eqnarray}  
  \label{eq:mmd-rkhs} 
  \text{MMD}[\hbspace,\pp{P},\pp{Q}] &=& \sup_{\|f\|\leq 1}\left\{\int f(\x)\dd\pp{P}(\x) - \int f(\y)\dd\pp{Q}(\y)\right\} \nonumber \\
  &=& \sup_{\|f\|\leq 1}\left\{ \langle f, \muv_{\pp{P}}-\muv_{\pp{Q}}\rangle_\hbspace\right\} \nonumber \\
  &=& \|\muv_{\pp{P}} - \muv_{\pp{Q}}\|_{\hbspace} , 
\end{eqnarray} 
\noindent where we use the reproducing property of $\hbspace$ and the linearity of the inner product, respectively. Thus, we can express the MMD in terms of the associated kernel function $k$ as
\begin{equation}
  \label{eq:mmd-kernel}
  \text{MMD}^2[\hbspace,\pp{P},\pp{Q}] = \ep_{X,\tilde{X}}[k(X,\tilde{X})] - 2\ep_{X,Y}[k(X,Y)] + \ep_{Y,\tilde{Y}}[k(Y,\tilde{Y})] ,
\end{equation}
\noindent where $X,\tilde{X}\sim \pp{P}$ and $Y,\tilde{Y} \sim \pp{Q}$ are independent copies. Note that in deriving \eqref{eq:mmd-kernel}, we use
\[
	\| \muv_{\pp{P}} \|^2_\hbspace = \langle \ep[k(\cdot,X)],\ep[k(\cdot,\tilde{X})]\rangle_\hbspace = \ep[k(X,\tilde{X})].
\]
\noindent It was shown in \citet[Theorem 21]{Sriperumbudur10:Metrics} that $\text{MMD}[\hbspace,\pp{P},\pp{Q}] \leq \gamma[\fspace_{\text{W}},\pp{P},\pp{Q}]$ and $\text{MMD}[\hbspace,\pp{P},\pp{Q}] \leq \sqrt{C}\gamma[\fspace_{\text{TV}},\pp{P},\pp{Q}]$ for some constant $C$ satisfying $\sup_{\x\in\inspace} k(\x,\x) \leq C < \infty$. As a result, if $\pp{P}$ is close to $\pp{Q}$ in these distances, they are also close in the MMD metric.

If $\hbspace$ is characteristic, it follows from \eqref{eq:mmd-rkhs} that $\text{MMD}[\hbspace,\pp{P},\pp{Q}] = 0$ if and only if $\pp{P} = \pp{Q}$. Moreover, when $k$ is translation invariant, \ie, $k(\x,\x') = \psi(\x-\x'),\,\x,\x'\in\mathbb{R}^d$, we have $\text{MMD}(\hbspace,\pp{P},\pp{Q}) = \int_{\rr^d}|\cf{\pp{P}}(\bm{\omega}) - \cf{\pp{Q}}(\bm{\omega})|^2\,d\Lambda(\bm{\omega})$ where $\Lambda$ is the spectral measure appearing in Bochner's theorem and $\cf{\pp{P}},\cf{\pp{Q}}$ are characteristic functions of $\pp{P},\pp{Q}$, respectively, \citep[Corollary 4]{Sriperumbudur10:Metrics}. In other words, the MMD can be interpreted as the distance in $L^2(\Lambda)$ between $\cf{\pp{P}}$ and $\cf{\pp{Q}}$.
 
Before describing an empirical estimate of \eqref{eq:mmd-rkhs}, we give basic definitions of $U$-statistics and $V$-statistics which are the cores of the empirical estimate of MMD, see, \eg, \citet[Chapter 5]{Serfling80:Approx}.

\begin{shadowbox}
  \begin{definition}[$U$-statistics and $V$-statistics, \citealp{Serfling80:Approx}]
    Let $X_1,X_2,\ldots,X_m$ be independent observations on a distribution $\pp{P}$. Consider a functional $\theta = \theta(\pp{P})$ given by 
    \begin{eqnarray*}
      \theta(\pp{P}) &=& \ep_{\pp{P}}[h(X_1,\ldots,X_m)] \\
      &=& \int\cdots\int h(\x_1,\ldots,\x_m)\dd\pp{P}(\x_1)\cdots\dd\pp{P}(\x_m),
    \end{eqnarray*}
    for some real-valued measurable function $h = h(\x_1,\ldots,\x_m)$ called a \emph{kernel}. The corresponding \emph{U-statistic} of order $m$ for estimation of $\theta$ on the basis of a sample $\x_1,\x_2,\ldots,\x_n$ of size $n\geq m$ is obtained by averaging the kernel $h$ symmetrically over the observations:
    $$U_n = U(\x_1,\x_2,\ldots,\x_n) = \frac{1}{{n \choose m}}\sum_c h(\x_{i_1},\x_{i_2},\ldots,\x_{i_m}), $$
    where $\sum_c$ denotes summation over the ${n \choose m}$ combinations of $m$ distinct elements $\{i_1,i_2,\ldots,i_m\}$ from $\{1,2,\ldots,n\}$. 

    An \emph{associated von Mises statistic}, or $V$-statistic, corresponding to a $U$-statistic $U_n$ for estimation of $\theta(\pp{P})$ is given by
    $$V_n = \frac{1}{n^m}\sum_{i_1=1}^n\cdots\sum_{i_m=1}^n h(\x_{i_1},\ldots,\x_{i_m}) = \theta(\pp{P}_n), $$
    where $\pp{P}_n$ denotes the sample distribution function for a discrete uniform distribution on $\{\x_1,\x_2,\ldots,\x_n\}$.
  \end{definition}
\end{shadowbox}
   
\noindent Clearly, $U_n$ is an unbiased estimate of $\theta$, whereas $V_n$ is not. For instance, let $\theta(\pp{P}) = \int \x \dd\pp{P}(\x)$, \ie, a mean of $\pp{P}$. For the kernel $h(\x) = \x$, the corresponding $U$-statistic is $U_n = \frac{1}{n}\sum_{i=1}^n\x_i$, the sample mean. An example of a degree-2 $V$-statistic is the maximum likelihood estimator of variance $V_n = \frac{1}{n^2}\sum_{i,j=1}^n \frac{1}{2}(\x_i - \x_j)^2 = \frac{1}{n}\sum_{i=1}^n (\x_i - \bar{\x})^2$ using the kernel $h(\x,\y) = (\x-\y)^2/2$. 

Given i.i.d. samples $\X = \{\x_1,\ldots,\x_m\}$ and $\Y = \{\y_1,\ldots,\y_n\}$ from $\pp{P}$ and $\pp{Q}$, respectively, a biased empirical estimate of MMD can be obtained as 
\begin{equation}  
  \label{eq:empirical-mmd}  
  \widehat{\text{MMD}}^2_b[\hbspace,\X,\Y] := \sup_{\|f\|_{\hbspace}\leq 1}\left\{\frac{1}{m}\sum_{i=1}^mf(\x_i) - \frac{1}{n}\sum_{j=1}^nf(\y_j)\right\}.
\end{equation}
The empirical MMD can be expressed in terms of empirical mean embeddings as $\widehat{\text{MMD}}_b^2[\hbspace,\X,\Y] = \|\muh_{\X} - \muh_{\Y}\|_{\hbspace}^2$. Moreover, we can write an unbiased estimate of the MMD entirely in terms of $k$ as \citep[Corollary 2.3]{Borgwardt06:MMD}
\begin{eqnarray}
  \label{eq:empirical-mmd2}
  \widehat{\text{MMD}}^2_u[\hbspace,\X,\Y] &=& \frac{1}{m(m-1)}\sum_{i=1}^m\sum_{j\neq i}^mk(\x_i,\x_j) \nonumber \\
  && + \frac{1}{n(n-1)}\sum_{i=1}^n\sum_{j\neq i}^nk(\y_i,\y_j) \nonumber \\
  && - \frac{2}{mn}\sum_{i=1}^m\sum_{j=1}^nk(\x_i,\y_j) .
\end{eqnarray}
Note that \eqref{eq:empirical-mmd2} is an unbiased estimate which is a sum of two $U$-statistics and a sample average \citep[Chapter 5]{Serfling80:Approx}. That is, assuming that $m=n$,
\begin{equation}
  \label{eq:mmd-ustat}
  \widehat{\text{MMD}}^2_u[\hbspace,\X,\Y] = \frac{1}{m(m-1)}\sum_{i\neq j}^mh(\mathbf{v}_i,\mathbf{v}_j),
\end{equation} 
\noindent where $\mathbf{v}_i = (\x_i,\y_i)$ and $h(\mathbf{v}_i,\mathbf{v}_j) := k(\x_i,\x_j) + k(\y_i,\y_j) - k(\x_i,\y_j) - k(\x_j,\y_i)$. We assume throughout that $\ep[h^2] < \infty$. The biased counterpart $\widehat{\text{MMD}}^2_b[\hbspace,\X,\Y]$ can be obtained using $V$-statistics. The convergence of empirical MMD has been established in \citet[Theorem 7]{Gretton12:KTT}. Theorem \ref{thm:mmd-asymptotic} below describes an unbiased quadratic-time estimate of the MMD, and its asymptotic distribution under an alternative hypothesis that $\pp{P}\neq\pp{Q}$.

\begin{shadowbox}
\begin{theorem}[{\citealp[Lemma 6; Corollary 16]{Gretton12:KTT}}]\label{thm:mmd-asymptotic}
  Given i.i.d. samples $\X = \{\x_1,\ldots,\x_m\}$ and $\Y = \{\y_1,\ldots,\y_m\}$ from $\pp{P}$ and $\pp{Q}$, respectively. When $\pp{P}\neq\pp{Q}$, an unbiased empirical estimate $\widehat{\text{MMD}}^2_u[\hbspace,\X,\Y]$ given in \eqref{eq:mmd-ustat} converges in distribution to a Gaussian distribution
  \begin{equation*}
    \sqrt{m}(\widehat{\text{MMD}}^2_u[\hbspace,\X,\Y] - \text{MMD}^2[\hbspace,\pp{P},\pp{Q}]) \overset{p}{\to} \mathcal{N}(0,\sigma^2_{\mathit{XY}}), 
  \end{equation*}
  \noindent where
  \begin{equation*}
    \sigma^2_{\mathit{XY}} = 4(\ep_{\mathbf{v}_1}[(\ep_{\mathbf{v}_2}h(\mathbf{v}_1,\mathbf{v}_2))^2] - [\ep_{\mathbf{v}_1,\mathbf{v}_2}h(\mathbf{v}_1,\mathbf{v}_2)]^2)
  \end{equation*}
  \noindent uniformly at rate $1/\sqrt{m}$.
\end{theorem}
\end{shadowbox}

\citet{Sriperumbudur2012} studied the convergence rate of the empirical estimators of the IPM for several function classes $\fspace$ including the MMD. It was shown that the MMD enjoys rapid convergence, \ie, in the order of $1/\sqrt{n}$ and the rate is independent of the dimension $d$, whereas other metrics such as the Wasserstein distance suffers from a slow rate that depends on $d$. While this is generally true, $d$ may show up in a constant term, which can make the upper bound arbitrarily large for high-dimensional data. In other words, $|\widehat{\text{MMD}^2_u}[\hbspace,\X,\Y]-\text{MMD}^2[\hbspace,\pp{P},\pp{Q}]|\le c_d(m^{-1/2}+n^{-1/2}),$ where $c_d$ depends on $d$ but not on $m$ and $n$. As a result, the MMD may still suffer from the curse of dimensionality. In fact, when it is used as a test statistic in two-sample or independent tests, the effect of dimensionality is more pronounced. \citet{Sashank15:HighDim} shows that the power of the linear-time MMD tests---under a mean-shift alternative---depends on the dimension $d$ and may decrease as $d$ grows. \citet{Ramdas15:DPK} also demonstrates that the power of the kernel-based tests degrades with dimension. Furthermore, \citet{Ramdas15:AdaptiveKME} investigated the adaptivity of the MMD tests with the Gaussian kernel and gave numerous remarks on how dimension could affect the power of the tests. Understanding how the MMD, and the kernel mean embeddings in general, behave in high dimensions is a potential research direction.

A natural application of the MMD is \emph{two-sample testing}: a statistical hypothesis test for equality of two samples. In particular, we test the \emph{null hypothesis} $H_0: \|\muv_{\pp{P}}-\muv_{\pp{Q}}\|_{\hbspace}=0$ against the \emph{alternative hypothesis} $H_1: \|\muv_{\pp{P}}-\muv_{\pp{Q}}\|_{\hbspace}\neq 0$. However, even if the two samples are drawn from the same distribution, the MMD criterion may still be non-zero due to the finite sample. \citet{Gretton12:KTT} proposes two distribution-free tests based on large deviation bounds (using Rademacher complexity and bound on $U$-statistics of \citealp{Hoeffding48:UStat}) and the third one based on the asymptotic distribution of the test statistics. The tests based on large deviation bounds are generally more conservative than the latter as the critical value is distribution independent, \ie~it is independent of the knowledge of $\pp{P}$ and $\pp{Q}$. 
 
The MMD test has several advantages over existing methods proposed in the literature \citep{Anderson94:TwoSample,BiauG05:Asymp,Nguyen08:DivFunc}. First, the MMD test is distribution free.\footnote{Note that even if a test is consistent, it is not possible to distinguish distributions with high probability at a given, fixed sample size.} That is, the assumption on the parametric form of the underlying distribution is not needed. Furthermore, like most of the kernel-based tests, \eg, \citet{Harchaoui07:Homogeneity}, the test can be applied in structured domains like graphs and documents as soon as the positive definite kernel is well-defined. Lastly, an availability of the asymptotic distribution of the test statistic allows for better approximation of the empirical estimates of the test statistic.

A \emph{goodness-of-fit} testing is an equally important problem. That is, we test whether the given sample $\{\x_i\}$ is drawn from a particular distribution $q$ or from a family of distributions $\mathcal{Q}$. Traditional goodness-of-fit tests are the $\chi^2$-test and the Kolmogorov-Smirnov test. In principle, we can turn it into a two-sample test by drawing a sample $\{\y_i\}$ from $q$ and apply the MMD tests. However, it is often difficult in practice to draw samples from $q$, especially when $q$ corresponds to complex and high dimensional distributions. \citet{Liu16:KernelStein} and \citet{Chwialkowski16:KernelGoodness} independently and simultaneously develop a goodness-of-fit test based on Stein's identity applied to functions in RKHS \citep{Oates16:MonteCarlo}.

\subsection{Scaling up MMD} 

The MMD can be computed in quadratic time $O(n^2d)$, which might prohibit its applications in large-scale problems. In \citet{Gretton12:KTT}, the authors also propose the linear time statistics and test by using the subsampling of the term in \eqref{eq:empirical-mmd2}, \ie, drawing pairs from $\X$ and $\Y$ without replacement. This method reduces the time complexity of MMD from $O(n^2d)$ to $O(nd)$. However, the test has high variance due to loss of information. The $B$-test of \citet{Zaremba2013} trades off the computation and variance of the test by splitting two-sample sets into corresponding subsets and then computes the exact MMD in each block while ignoring between-block interactions with $O(n^{3/2}d)$ time complexity.\footnote{The $B$-test can be understood as a specific case of the tests using \emph{incomplete} $U$-statistic \citep{Blom76:incU}. While this kind of statistic can be obtained in numerous ways, it has not been explored much in the context of MMD.}

Recall that when $k(\x,\x') = \psi(\x-\x')$ for some positive definite function $\psi$ on $\rr^d$, we have
\begin{equation}
  \label{eq:mmd-fourier-2}
  \text{MMD}(\hbspace,\pp{P},\pp{Q}) = \int_{\rr^d}|\cf{\pp{P}}(\bm{\omega}) - \cf{\pp{Q}}(\bm{\omega})|^2\,d\Lambda(\bm{\omega}),
\end{equation}
\noindent where 
$\cf{\pp{P}},\cf{\pp{Q}}$ are characteristic functions of $\pp{P},\pp{Q}$, respectively. Based on \eqref{eq:mmd-fourier-2}, \citet{Ji15:FastMMD} proposes an efficient test called \emph{FastMMD} which employs the random Fourier feature to transform the MMD test with translation invariant kernel. In this case, the empirical MMD becomes the $L_2(\mathbb{R}^d,\Lambda)$ distance between the empirical characteristic functions. Later, \citet{Kacper15:Analytic} demonstrates that the original formulation fails to distinguish a large class of measures, and presents a ``smoothed'' version of the formulation using an analytic smoothing kernel. The proposed test statistic is a difference in expectations of analytic functions at particular ``random'' spatial or frequency locations. The resulting metric is shown to be a ``random metric'' satisfying all the conditions for a metric with qualification ``almost surely''. The time complexity also reduces to $O(nd)$. For kernels whose spectral distributions $\Lambda(\bm{\omega})$ are spherically invariant, \ie, $\Lambda(\bm{\omega})$ only depends on $\|\bm{\omega}\|_2$, the cost reduces further to $O(n\log d)$ by using the Fastfood technique \citep{Le2013}. Interestingly, the proposed linear-time test can outperform the quadratic-time MMD in terms of power of the test. Based on the work of \citet{Kacper15:Analytic}, \citet{Wittawat16:Interpretable} proposes test statistics as a difference of mean embeddings evaluated at locations chosen to maximize the test power lower bound. The benefits are two-fold. First, these features are chosen to maximize the distinguishability of the distributions, thus leading to more powerful tests. Second, the obtained set of features is interpretable. That is, they are features that best distinguish the probability distributions. 

As pointed out in previous work, we may pose the problem of distribution comparison as a binary classification (see, \eg, \citealp[Remark 20]{Gretton12:KTT} and \citealp{Sriperumbudur2009}). That is, any classifiers for which uniform convergence bounds can be obtained such as neural networks, support vector machines, and boosting, can be used for the purpose of distribution comparison. The benefit of this interpretation is that there is a clear definition of the loss function which can be used for the purpose of parameter selection. A slightly different interpretation is to look at this problem as a learning problem on probability distributions \citep{Muandet12:SMM,Muandet13:OCSMM,Zoltan15:DistReg}. For example, the goal of many hypothesis testing problems is to learn a function from an empirical distribution $\hat{\pp{P}}$ to $\{0,1\}$ which, for example, indicates whether or not to reject the null hypothesis. If the training examples $(\hat{\pp{P}}_1,y_1),\ldots,(\hat{\pp{P}}_n,y_n)$ are available, we can consider a hypothesis testing as a machine learning problem on distributions.

Lastly, a commonly used kernel for a MMD test on $\rr^d$ is the Gaussian RBF kernel $k(\x,\x')=\exp(-\|\x-\x'\|^2/2\sigma^2)$ whose bandwidth parameter is chosen via the \emph{median heuristic}: $\sigma^2 = \mathrm{median}\{\|\x_i-\x_j\|^2\,:\,i,j=1,\ldots,n\}$ \citep{Gretton05:KIND}. While this heuristic has been shown to work well in many applications, it may run into trouble when the sample size is small. In fact, it has been observed empirically that the median heuristic may not work well when estimating the kernel mean from a small sample and there is room for improvement, especially in the high-dimensional setting \citep{Danafar13:RMMD,Muandet2014,Muandet2014a,Sashank15:HighDim}. An alternative is to choose the kernel that maximizes the test statistic, which is found to outperform the median heuristic empirically \citep{Sriperumbudur2009}. \citet{Gretton12:SSSBPF} proposes a criterion to choose a kernel for two-sample testing using MMD. The kernel is chosen so as to maximize the test power for a fixed Type-I error. The proposed method corresponds to maximizing the Hodges and Lehmann asymptotic relative efficiency \citep{Hodges1956:Asymptotic}. Despite these efforts, how to choose a good kernel function on its own remains an open question.

The MMD has been applied extensively in many applications, namely, clustering \citep{Jegelka2009}, density estimation \citep{Song2007,Song08:TDE,Sriperumbudur11:Mixtures}, (conditional) independence tests \citep{Fukumizu2008,Doran2014,Chwialkowski2014}, causal discovery \citep{Sgouritsa2013,Chen2014,Schoelkopf15:KPP}, covariate shift \citep{Gretton09:CSKMM,Pan11:TCA} and domain adaptation \citep{Blanchard11:Generalize,Muandet13:DG}, selection bias correction \citep{Huang07:SSB}, herding \citep{Chen2010,Huszar2012}, Markov chain Monte Carlo \citep{Sejdinovic2014}, moment matching for training deep generative models \citep{Li15:GMMN,Dziugaite15:DeepMMD}, statistical model criticism \citep{Lloyd2015:Criticism,Kim16:Criticism}, approximate Bayesian computation \citep{Park16:K2ABC}, distribution regression \citep{Zoltan15:DistReg} and model selection in generative models \citep{Bounliphone16:GenModel}, for example.
 
\subsection{Application: Training Deep Generative Models via MMD}

In this section, we highlight the promising application of MMD in training deep generative models which was recently assayed by \citet{Li15:GMMN} and \citet{Dziugaite15:DeepMMD}. Deep generative models are neural networks that provide a distribution from which one can generate samples by feeding the models with random noise $Z$. Generative Adversarial Networks (GAN) proposed by \citet{Goodfellow14:GAN} is one of the most popular deep generative models used in several computer vision applications, see, \eg, \citet{Bouchacourt16:DiscoNets} and references therein. A GAN model consists of two neural networks---namely a \emph{generative network} and a \emph{discriminative network}---trained in an adversarial manner. The generative network, also referred as the generator $G_{\bm{\theta}}$, aims to replicate the training data from input noise, whereas the discriminative network, also referred as the discriminator $D_{\bm{\phi}}$, acts as an adversary that aims to distinguish the synthetic data generated by the generator from the training data. The learning is considered successful if the synthetic data is indistinguishable from the real data.

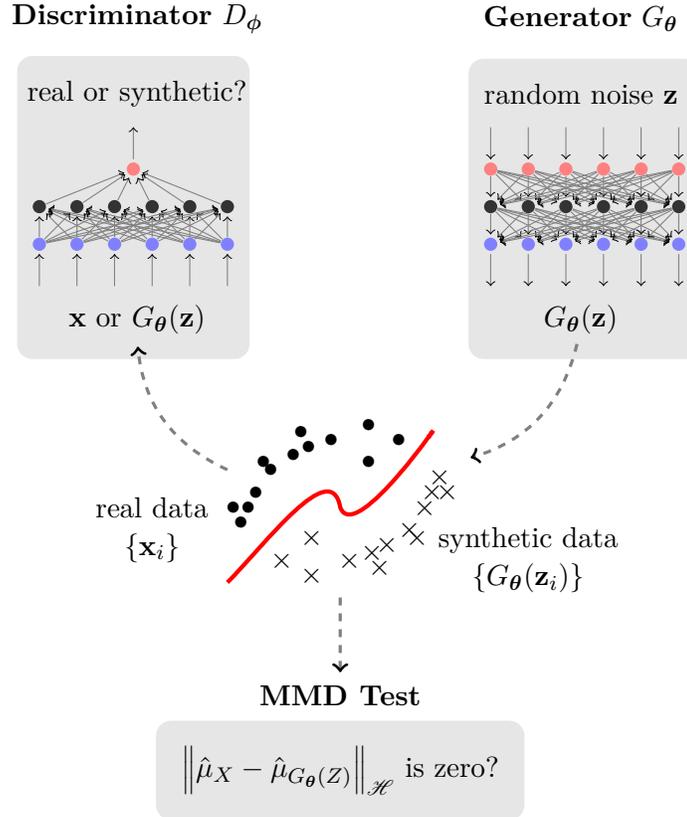
\begin{figure}[t!]
  \centering
  \def\layersep{1.5cm}
  \begin{tikzpicture}[shorten >=1pt,->,draw=black!50, node distance=\layersep]
    \tikzstyle{every pin edge}=[<-,shorten <=1pt]
    \tikzstyle{neuron}=[circle,fill=black!25,minimum size=5pt,inner sep=0pt]
    \tikzstyle{input neuron}=[neuron, fill=red!50];
    \tikzstyle{output neuron}=[neuron, fill=blue!50];
    \tikzstyle{hidden neuron}=[neuron, fill=black!80];
    \tikzstyle{annot} = [text width=4em, text centered]

    \begin{scope}
    \foreach \name / \x in {4,...,9}
        \node[input neuron,pin={[pin edge={<-}]above:}] (I-\name) at (\x*0.5,3) {};

        \node (txt) at (3.2,4) {random noise $\mathbf{z}$};

     \foreach \name / \x in {4,...,9}
        \path[yshift=0.5cm]
            node[hidden neuron] (H-\name) at (\x*0.5,2 cm) {};
           
    \foreach \name / \x in {4,...,9}
      \path[yshift=1cm]
          node[output neuron,pin={[pin edge={->}]below:}] (O-\name) at (\x*0.5,1 cm){};

           \node (txtg) at (3.2,1) {$G_{\bm{\theta}}(\mathbf{z})$};

    \foreach \source in {4,...,9}
        \foreach \dest in {4,...,9}
            \path (I-\source) edge (H-\dest);

    \foreach \source in {4,...,9}
       \foreach \dest in {4,...,9}
           \path (H-\source) edge (O-\dest);
    
           \node (labg) at (3.2,5) {\textbf{Generator} $G_{\bm{\theta}}$};

    \end{scope}

    \begin{scope}
    \foreach \name / \x in {4.5}
        \node[input neuron,pin={[pin edge={->}]above:}] (II) at (\x*0.5-5,3) {};

        \node (txtd2) at (-2.7,4) {real or synthetic?};

     \foreach \name / \x in {2,...,7}
        \path[yshift=0.5cm]
            node[hidden neuron] (HH-\name) at (\x*0.5-5,2 cm) {};
           
    \foreach \name / \x in {2,...,7}
      \path[yshift=1cm]
          node[output neuron,pin={[pin edge={<-}]below:}] (OO-\name) at (\x*0.5-5,1 cm){};

          \node (txtd) at (-2.7,1) {$\mathbf{x} \text{ or } G_{\bm{\theta}}(\mathbf{z})$};    

    \foreach \source in {4.5}
        \foreach \dest in {2,...,7}
            \path (HH-\dest) edge (II);

    \foreach \source in {2,...,7}
       \foreach \dest in {2,...,7}
           \path (OO-\dest) edge (HH-\source);

           \node (labd) at (-2.7,5) {\textbf{Discriminator} $D_{\bm{\phi}}$};

         \end{scope}

     \foreach \Point in
      {(-0.1,0.8),(-0.5,0.5),(-0.5,1),(-1,0.8),(-1.3,0.7),(-1.4,0.9),
      (-1.5,0.6),(-1.8,0.4),(-1.9,0.5),(-2,0.1),(-2.1,-0.1),
    (-2.3,-0.1),(-2.2,-0.3)}{
        \node at \Point[xshift=25pt,yshift=-40pt] {\large\textbullet};
      }

      \draw [ultra thick,red,-] (-1.5,-2.5) to[out=45,in=100] (0,-1.5) to[out=-140+70,in=55] (1.15,-0.6);

      \foreach \Point in
      {(0.1,-0.8) ,(0.5,-0.5),(0.5,-1),(1,-0.8),(1.3,-0.7),(1.4,-0.9),
      (1.5,-0.6),(1.8,-0.4),(1.9,-0.5),(2,-0.1),(2.1,0.1),
      (2.3,0.1),(2.2,0.3)}{
        \node at \Point[xshift=-25pt,yshift=-40pt] {\large$\times$};
      }

      \node (rdattxt) at (-2.5,-1.8)[align=center,text width=1.5cm] {real data $\{\x_i\}$};
      \node (sdattxt) at (2.5,-2.2)[align=center,text width=3.2cm] {synthetic data $\{G_{\bm{\theta}}(\z_i)\}$};

      \node (mmd) at (0,-5) {$\left\| \muh_X - \muh_{G_{\bm{\theta}}(Z)} \right\|_{\hbspace}$ is zero?};
      \node (mmdtxt) at (0,-4) {\textbf{MMD Test}};

      \path[->,very thick,dashed]
      (txtg) edge[bend left] node[below] {} (1.7,-0.85);
      \path[->,very thick,dashed]
      (-1.5,-1) edge[bend left] node[above] {} (txtd); 
      \path[->,very thick,dashed]
      (0,-2.7) edge[] node[below] {} (mmdtxt);

      \begin{pgfonlayer}{background}
        \filldraw [line width=4mm,join=round,black!10]
        (txt.north  -| I-4.west)  rectangle (txtg.south  -| I-9.east)
        (txtd2.north -| HH-2.west) rectangle (txtd.south -| HH-7.east)
        (mmd.north -| mmd.west) rectangle (mmd.south -| mmd.east);
      \end{pgfonlayer}
      
  \end{tikzpicture}
  \caption{A comparison between GANs and MMD nets. While GAN (top) uses a deep neural network as a discriminator aiming to discriminate real data from synthetic data generated by the generator, the MMD net (bottom) treats this as a two-sample testing problem---rather than a classification problem---and replaces the discriminator with the MMD test. In both approaches, the generator is trained to generate the synthetic data that are indistinguishable from the real data.}
  \label{fig:gan-mmd}
\end{figure}
  
Training the GAN amounts to solving the minimax optimization problem \citep{Goodfellow14:GAN} 
\begin{equation}
\label{eq:gan-obj}
\min_{\bm{\theta}}\max_{\bm{\phi}}V(G_{\bm{\theta}},D_{\bm{\phi}}) 
= \ep_X[\log D_{\bm{\phi}}(X)] + \ep_{Z}[\log(1 - D_{\bm{\phi}}(G_{\bm{\theta}}(Z)))],
\end{equation}
\noindent where $X$ is distributed according to the data distribution and $Z$ is distributed according to a prior over noise variable. As mentioned in \citet{Goodfellow14:GAN}, the algorithm can be interpreted as approximately minimizing the Jensen-Shannon divergence. On the other hand, the idea proposed in \citet{Li15:GMMN} and \citet{Dziugaite15:DeepMMD} is to replace the discriminators $\{D_{\bm{\phi}}\}$ by the MMD test. Specifically, rather than training a neural network to distinguish $X$ from $G_{\bm{\theta}}(Z)$, they view the problem as a two-sample testing using the MMD test statistic. As a result, the objective function $V(G_{\bm{\theta}},D_{\bm{\phi}}) $ in \eqref{eq:gan-obj} is substituted by
\begin{equation}
  \label{eq:mmd-nets}
  V_{\hbspace}(G_{\bm{\theta}},D_{\bm{\phi}}) 
= \sup_{\|f\|_\hbspace\leq 1} \Big|\ep_X[f(X)] - \ep_{Z}[f(G_{\bm{\theta}}(Z))]\Big|
\end{equation}
\noindent for some RKHS $\hbspace$. In this case, we can think of $\{D_{\bm{\phi}}\}$ as the unit ball of functions in RKHS $\hbspace$. The MMD test then plays the role of an adversary \citep{Dziugaite15:DeepMMD}. Figure \ref{fig:gan-mmd} gives an illustrative comparison.

The benefits of replacing the discriminative network with the MMD test are two-fold. Firstly, the MMD offers a closed-form solution for the discriminator in GAN. It is also more theoretically appealing because, if $\hbspace$ corresponds to the RKHS of a characteristic kernel, $V_{\hbspace}(G_{\bm{\theta}},D_{\bm{\phi}})$ vanishes if and only if $X$ and $G_{\bm{\theta}}(Z)$ come from the same distribution. Secondly, it avoids the hard minimax objective function used in traditional GAN. Hence, training the MMD nets is more computationally efficient.

Combining MMD with deep neural networks has recently become an active research area. In domain adaptation, for instance, the MMD has also been used to reduce the dataset bias and thus improve transferability of the features across different domains \citep{Long15:DeepAdapt,Long16:ResidTrans}. As briefly remarked by \citet{Dziugaite15:DeepMMD}, one of the promising research directions is to extend this idea to a broader class of (generative) models.\footnote{A similar remark was made by Sebastian Nowozin at the Dagstuhl seminar on New Directions for Learning with Kernels and Gaussian Processes that took place during November 27 -- December 2, 2016.} 

\section{Kernel Dependency Measures}
\label{sec:dependency}

A dependence measure is one of the most fundamental tools in statistical analysis. \citet{Renyi59:Dependence} outlines a list of seven desirable properties---known as \emph{R\'enyi's axioms}---which should be satisfied by any measure of dependence between two random variables. The classical criteria such as Spearman's rho and Kendall's tau can detect only linear dependences. Some other non-linear criteria such as mutual information based measures and density ratio methods, may require certain assumptions regarding the parametric form of the underlying distribution which is too restrictive in several applications.

When the dependence is non-linear, one of the most successful non-parametric measures is the \emph{Hilbert Schmidt Independence Criterion (HSIC)} \citep{Gretton05:MSD}. Let $\hbspace$ and $\hbspg$ be separable RKHSs on $\mathcal{X}$ and $\mathcal{Y}$ with reproducing kernels $k$ and $l$, respectively. The HSIC is defined as the square HS-norm of the associated cross-covariance operator:
\begin{eqnarray}
  \label{eq:hsic-criterion}
  \text{HSIC}(\hbspace,\hbspg,k,l) &=& \|\covxy\|^2_{\text{HS}} \nonumber \\
  &=& \ep_{\x\x'\y\y'}[k(\x,\x')l(\y,\y')] \nonumber \\
  && + \ep_{\x\x'}[k(\x,\x')]\ep_{\y\y'}[l(\y,\y')] \nonumber \\
  && -2\ep_{\x\y}[\ep_{\x'}[k(\x,\x')]\ep_{\y'}[l(\y,\y')]] .
\end{eqnarray}
If the product kernel $k(\cdot,\cdot)\times l(\cdot,\cdot)$ is characteristic on $\inx\times\iny$, it is not difficult to show that $\covxy =\mathbf{0}$ and $\text{HSIC}(X,Y,k,l)=0$ if and only if $X\ci Y$. To see this, by definition we have $\covxy = \muv_{\pp{P}_{\mathit{YX}}} - \muv_{\pp{P}_Y\otimes \pp{P}_X}$ (cf. \S\ref{sec:cross-covariance}). Hence, $\covxy = \mathbf{0}$ implies $\muv_{\pp{P}_{\mathit{YX}}} = \muv_{\pp{P}_Y\otimes \pp{P}_X}$. Since the product kernel is characteristic, it follows that $\pp{P}_{\mathit{YX}} = \pp{P}_Y\otimes \pp{P}_X$. As an analogy, when $X$ and $Y$ are jointly Gaussian in $\rr^d$, $C_{\mathit{XY}}=\mathbf{0}$ if and only if $X$ and $Y$ are independent where $C_{\mathit{XY}}$ denotes the cross-covariance matrix.

An empirical unbiased estimate of the HSIC statistic from an i.i.d. sample $(\x_1,\y_1),\ldots,(\x_n,\y_n)$ on $\inx\times\iny$ is given by
\begin{eqnarray}
  \label{eq:empirical-hsic} 
  \widehat{\text{HSIC}}(X,Y,k,l) &=& \|\ecovxy\|_{\text{HS}}^2 \nonumber \\
  &=& \frac{1}{n^2}\sum_{i,j=1}^nk(\x_i,\x_j)l(\y_i,\y_j) \nonumber \\
  && -\frac{2}{n^3}\sum_{i,j,p}^nk(\x_i,\x_j)l(\y_i,\y_p) \nonumber \\
  && + \frac{1}{n^4}\sum_{i,j=1}^nk(\x_i,\x_j)\sum_{p,q=1}^n l(\y_p,\y_q) ,
\end{eqnarray} 
\noindent or equivalently $\widehat{\text{HSIC}}(X,Y,k,l) = (1/n^2)\mathrm{tr}(\tilde{\kmat}\tilde{\lmat})$ where $\tilde{\kmat} = \hmat\kmat\hmat$, $\tilde{\lmat} = \hmat\lmat\hmat$, and $\hmat = \id - (1/n)\mathbf{1}\mathbf{1}^\top$. The empirical estimate \eqref{eq:empirical-hsic} is equivalent to the quadratic dependence measure of \citet{Achard03:QD}, although it does not imply that the random variables are independent when the test of \citet{Achard03:QD} is zero (see \citealp[Appendix B]{Gretton05:MSD} for the proof of equivalence and discussion). It can be computed in $O(n^2)$ time and converges to the population HSIC at a rate of $1/\sqrt{n}$ if the bias arising from the definition of HSIC is negligible (a slight improvement was later given in \citealp{Song07:FSDE} for an unbiased estimator of HSIC). Relying on the large deviation bound, \citet{Gretton05:MSD} defines an independence test with critical region $\{((\x_i,\y_i))^n_{i=1}:\widehat{\text{HSIC}}(X,Y,k,l)\ge C\sqrt{\log(1/\alpha)/n} \}$, where $\alpha$ is the significance level of the test and $C$ is a constant that depends only on $k$ and $l$.

Due to its generality, HSIC has proven successful in both statistics and machine learning. Many learning problems can be interpreted as a maximization or minimization of the dependence between $X$ and $Y$. For instance, \citet{Song07:FSDE} considers supervised feature selection as maximizing the dependence between the subsets of covariate $X'$, obtained from backward elimination algorithm \citep{Guyon03:IVF}, and the target $Y$, \eg, outputs of binary, multi-class, and regression settings. Likewise, when $Y$ denotes the cluster labels, \citet{Song2007} proposes a HSIC-based clustering algorithm called CLUHSIC that clusters the data by maximizing the HSIC between $X$ and $Y$. Interestingly, depending on the structure of the output kernel $l$, many classical clustering algorithms, \eg, k-means, weighted k-mean, and hierarchical clustering, can be considered as a special case of CLUHSIC. Moreover, the MaxMMD of \citet{Jegelka2009} is in fact equivalent to the dependence maximization framework. This concept has been extended to several applications including, for example, kernelized sorting \citep{Quadrianto09:KernSort}.

Alternatively, we may pose the independence test between $X$ and $Y$ as a two-sample testing problem. Recall that $X$ and $Y$ are said to be independent if and only if their joint distribution factorizes as $\pp{P}_{\mathit{XY}} = \pp{P}_X\otimes\pp{P}_Y$. Let $\mu_{\pp{P}_{\mathit{XY}}}$ and $\mu_{\pp{P}_X\otimes \pp{P}_Y}$ be the kernel mean embedding of $\pp{P}_{\mathit{XY}}$ and $\pp{P}_X\otimes\pp{P}_Y$, respectively. The MMD test statistic for testing independence can then be written as
\begin{equation}
  \label{eq:dependence-mmd}
  \widehat{\text{MMD}}(\hbspace\otimes\hbspg,\pp{P}_{\mathit{XY}},\pp{P}_X\otimes\pp{P}_Y) = \left\| \muh_{\pp{P}_{\mathit{XY}}}-\muh_{\pp{P}_X\otimes\pp{P}_Y}\right\|^2_{\hbspace\otimes\hbspg} ,
\end{equation}
\noindent where $\muh_{\pp{P}_{\mathit{XY}}}$ and $\muh_{\pp{P}_X\otimes\pp{P}_Y}$ denote the empirical estimates of $\pp{P}_{\mathit{XY}}$ and $\pp{P}_X\otimes\pp{P}_Y$, respectively. Since we only have access to sample $\{(\x_i,\y_i)\}_{i=1}^n$ from $\pp{P}_{\mathit{XY}}$, the corresponding sample from $\pp{P}_X\otimes\pp{P}_Y$ can be obtained approximately as $\{(\x_i,\y_{\pi(i)})\}_{i=1}^n$ where $\pi(\cdot)$ is a random permutation such that $\pi(i)\neq i$.\footnote{As noted in \citet{Janzing13:Causal} and \citet{Doran2014}, this random permutation is only an approximation: while it removes the dependence between $X$ and its corresponding $Y$, it introduces a dependence to one of the other $Y$ variables, which becomes negligible in the limit $n\rightarrow\infty$.} In light of \eqref{eq:dependence-mmd}, \citet{Doran2014} constructs a conditional independence test, relying on the learned permutation $\pi(\cdot)$ that additionally preserves the similarity on conditioning variable $Z$ (see \S\ref{sec:cond-dependency}).

Several extensions of HSIC have been proposed for settings in which there are three random variables $(X,Y,Z)$---\eg, conditional dependence measure, three-variable interaction, and relative dependency measure. For instance, \citet{Sejdinovic13:Lancaster} proposes a kernel nonparametric test to detect Lancaster three-variable interaction, \eg, V-structure. \citet{Bounliphone15:Relative} constructs a consistent test for relative dependency which---unlike the Lancaster test of \citet{Sejdinovic13:Lancaster}---measures dependency between a source variable and two candidate target variables. Taking into account the correlation between two HSIC statistics when deriving the corresponding \emph{joint} asymptotic distribution leads to a more powerful, consistent test than the test based on two independent HSIC statistics \citep[Theorem 4]{Bounliphone15:Relative}). Lastly, a nonparametric dependency test for an arbitrary number of random variables has also been considered in the literature and is required in many applications. However, the prominent issue is that, as the number of random variables grows, the convergence of the estimators may be arbitrarily slow. We will postpone the discussion of conditional dependence measures to \S\ref{sec:cond-dependency}.

\subsection{Extensions to Non-i.i.d. Random Variables}

The aforementioned dependence measures only work when the data are i.i.d. For non-i.i.d. data, assumptions on the form of dependency---often in terms of graphical models or \emph{mixing} conditions---are needed. For example, to cope with non-i.i.d. data, \citet{Zhang08:Dependence} proposes a \emph{Structured-HSIC} where $\pp{P}$ satisfies the conditional independence specified by an undirected graphical model, \eg, a Markovian dependence for sequence data and grid structured data. Thus, the kernel mean embedding decomposes along the maximal cliques of the graphical model. The Hammersley-Clifford theorem \citep{HammersleyClifford:1971} ensures the full support of $\pp{P}$ and hence the injectivity of the embedding. \citet{Zhang08:Dependence} applied the proposed measure to independent component analysis (ICA) and time series clustering and segmentation.

Kernel independence tests for time series---\eg, financial data and brain activity data---were recently proposed in a series of works including \citet{Besserve13:KCSD,Chwialkowski2014,ChwialkowskiSG14:WILD}. If one is interested in a serial dependency within a single time series, the test reduces to the i.i.d. case under null hypothesis. In order to test dependence between one time series and another, \citet{Besserve13:KCSD} characterizes dependences between time-series as the Hilbert-Schmidt norm of the kernel cross-spectral density (KCSD) operator which is the Fourier transform of the covariance operator at each time lag. In contrast, \citet{Chwialkowski2014} uses the standard HSIC test statistic whose null estimate is obtained by making a shift of one signal relative to the other, rather than the ordinary bootstrapping. The authors impose mixing conditions on random processes under investigation (see \citealp{Chwialkowski2014} for details).

\section{Learning on Distributional Data}
\label{sec:distributional-data}

In many machine learning applications, it may be more preferable to represent training data as probability measures $\pp{P}$ rather than just points $x$ in some input space $\inspace$. For instance, many classical problems such as multiple-instance learning \citep{Doran2013}, learning from noisy data, and group anomaly detection \citep{Poczos11:Divergence,Xiong11:HPM,Liang11:FGM,Muandet13:OCSMM}, may be viewed as empirical risk minimization when training data are probability distributions \citep{Muandet12:SMM}. More recently, modern applications of learning on distributions have also proven successful in statistical estimation and causal inference \citep{Poczos13:DistFree,Oliva14:FastD2R,Zoltan15:DistReg,Lopez-Paz15:Towards}. Furthermore, in the big data era, the benefits of representing data by distributions are two-fold: it reduces the effective sample size of the learning problems and also help conceal the identity of individual samples, which is an important milestone in privacy-preserving learning \citep{Dwork08:DPS}.

\subsection{Kernels on Probability Distributions}

Given distributions $\pp{P}_1,\pp{P}_2\ldots,\pp{P}_n$, the basic idea is to treat the embedding $\mu_{\pp{P}_1},\mu_{\pp{P}_2},\ldots,\mu_{\pp{P}_n}$ as their feature representation on which learning algorithms operate. For example, one of the most popular approaches is to define a positive definite kernel on probability distributions by
\begin{equation}
  \label{eq:dist-kernel}
  \kappa(\pp{P},\pp{Q}) := \langle \mu_{\pp{P}},\mu_{\pp{Q}}\rangle_{\hbspace}
  = \int\int_{\inspace} k(\x,\x')\dd\pp{P}(\x)\dd\pp{Q}(\x') \,.
\end{equation}
The empirical counterpart of \eqref{eq:dist-kernel} is obtained by replacing both integrals with finite sums over the i.i.d. samples $\x_1,\x_2,\ldots,\x_m\sim\pp{P}$ and $\y_1,\y_2,\ldots,\y_m\sim\pp{Q}$, \ie,
\begin{equation}
  \label{eq:empirical-dist-kernel}
  \kappa(\pp{P},\pp{Q}) \approx \langle \muh_{\pp{P}},\muh_{\pp{Q}}\rangle_{\hbspace} = \frac{1}{m^2}\sum_{i,j=1}^m k(\x_i,\y_j) \,.
\end{equation}
For some probability distributions and kernels $k$, the kernel \eqref{eq:dist-kernel} can be evaluated analytically, see, \eg, \citet[Table 1]{Song08:TDE} and \citet[Table 1]{Muandet12:SMM} which we reproduce here in Table \ref{tab:expected-kernel}. Since $\muh_{\pp{P}}$ can be infinite dimensional, another popular approach---especially in large-scale learning---is to find a finite approximation of $\muh_{\pp{P}}$ directly (see, \eg, \S\ref{sec:kme-approximation}). The advantages of this approach are that learning is often more efficient and any off-the-shelf learning algorithm can be employed.

\begin{table}[t!]
  \centering
  \caption{A closed-form solution of the kernel $\kappa(\pp{P},\pp{Q}) = \iint k(\x,\y)\dd\pp{P}(\x)\dd\pp{Q}(\y)$ for a certain class of probability distributions $\pp{P},\pp{Q}$ and kernel function $k$ (reproduced from \citealp{Muandet12:SMM}).}
  \label{tab:expected-kernel}
  \resizebox{\textwidth}{!}{
  \begin{tabular}{lll}
    \toprule
    Distribution & $k(\x,\y)$ & $\kappa(\pp{P}_i,\pp{P}_j) = \langle\muv_{\pp{P}_i},\muv_{\pp{P}_j}\rangle_{\hbspace}$\\
    \midrule
    $\pp{P}(\m;\Sigma)$& $\langle \x,\y \rangle$ & $\m_i^{\mathsf{T}}\m_j + \delta_{ij}\text{tr}\bm{\Sigma}_i$ \\
    $\mathcal{N}(\m,\Sigma)$ & $\exp(-\frac{\gamma}{2}\|\x-\y\|^2)$
    & $\exp(-\frac{1}{2}(\m_i-\m_j)^{\mathsf{T}}(\bm{\Sigma}_i+\bm{\Sigma}_j + \gamma^{-1}\id)^{-1}(\m_i-\m_j)) $ \\
    & & $/|\gamma\bm{\Sigma}_i + \gamma\bm{\Sigma}_j + \id|^{\frac{1}{2}}$\\
    $\mathcal{N}(\m,\Sigma)$ & $(\langle \x,\y\rangle + 1)^2$
    &  $(\langle \m_i,\m_j\rangle +1)^2 + \text{tr }\bm{\Sigma}_i\bm{\Sigma}_j + \m_i^{\mathsf{T}}\bm{\Sigma}_j\m_i + \m_j^{\mathsf{T}}\bm{\Sigma}_i\m_j $ \\
    $\mathcal{N}(\m,\Sigma)$ & $(\langle \x,\y\rangle + 1)^3$
    & $(\langle \m_i,\m_j\rangle +1)^3 + 6\m_i^{\mathsf{T}}\bm{\Sigma}_i\bm{\Sigma}_j\m_j $ \\
    & & $+ 3(\langle \m_i,\m_j\rangle +1)(\text{tr }\bm{\Sigma}_i\bm{\Sigma}_j + \m_i^{\mathsf{T}}\bm{\Sigma}_j\m_i + \m_j^{\mathsf{T}}\bm{\Sigma}_i\m_j)$ \\
    \bottomrule
  \end{tabular}}
\end{table}

The benefits of kernel mean representation for distributional data are three-fold. Firstly, the representation can be estimated consistently without any parametric assumption on the underlying distributions, unlike those based on generative probability models \citep{Jaakkola98:Generative,Jebara04:PPK}. Secondly, compared to density estimation approaches \citep{Poczos11:Divergence,Oliva14:FastD2R,Oliva13:D2D}, the kernel mean representation is less prone to the \emph{curse of dimensionality} and admits fast convergence in terms of sample size \citep{Shawe04:KMPA,Smola07Hilbert,Grunewalder12:LGBPP}. Lastly, being an element of the kernel feature space enables one to extend the whole arsenal of kernel methods to distributional data and to adopt existing tools for theoretical analysis, \eg, generalization bounds \citep{Zoltan15:DistReg,Lopez-Paz15:Towards,Muandet15:Thesis}.

\subsection{Properties of Distributional Kernels}

Note that the map $\mu: \pp{P}\mapsto\mu_{\pp{P}}$ is linear w.r.t. $\pp{P}$. Despite being non-linear in the input space $\inspace$, the function class induced by \eqref{eq:dist-kernel} is comprised of only linear functions over the probability space $\pspace(\mathcal{X})$. Hence, it cannot be a universal kernel in the sense of \citet{Steinwart:2002:IKC} for probability distributions. Let $\fspace := \{\pp{P}\mapsto\int_{\inspace} g\dd\pp{P}\,:\, \pp{P}\in\pspace(\inspace), g\in C_b(\inspace)\}$ where $C_b(\inspace)$ denotes the class of bounded continuous functions on $\inspace$. It follows that $C_b(\inspace)\subset\fspace\subset C_b(\pspace(\inspace))$ where $C_b(\pspace(\inspace))$ is the class of bounded continuous functions on $\pspace(\inspace)$. It is shown in \citet[Lemma 2]{Muandet12:SMM} that the function space induced by the kernel $\kappa$ is dense in $\fspace$ if $k$ is universal (see, \eg, Definition 4 in \citealp{Steinwart:2002:IKC}).

\begin{shadowbox}
\begin{lemma}[\citealp{Muandet12:SMM}]
  \label{lem:lin-universal}
  Assuming that $\inspace$ is compact, the RKHS $\hbspace$ induced by the kernel $k$ is dense in $\fspace$ if $k$ is universal in the sense of \citet{Steinwart:2002:IKC}.
\end{lemma}
\end{shadowbox}

A more challenging question to answer is whether there exists a non-linear kernel $\kappa$ on $\pspace(\inspace)$ which is dense in $C_b(\pspace(\inspace))$. For example, \citet{Christmann10:Kernels} considers a Gaussian-like kernel
\begin{equation}
  \label{eq:l2-rbf}
  \kappa(\pp{P},\pp{Q}) = \exp\left( -\frac{\|\mu_{\pp{P}} - \mu_{\pp{Q}}\|^2_{\hbspace}}{2\sigma^2}\right) .
\end{equation}

\begin{shadowbox}
\begin{theorem}[\citealp{Christmann10:Kernels}]
  If the map $\pp{P}\mapsto\mu_{\pp{P}}$ is injective for a compact space $\inspace$, the function space $\hbspace_\kappa$ is dense in $C_b(\pspace(\inspace))$.
\end{theorem}
\end{shadowbox}

\begin{figure}[t!]
  \centering
  \includegraphics[width=0.95\textwidth]{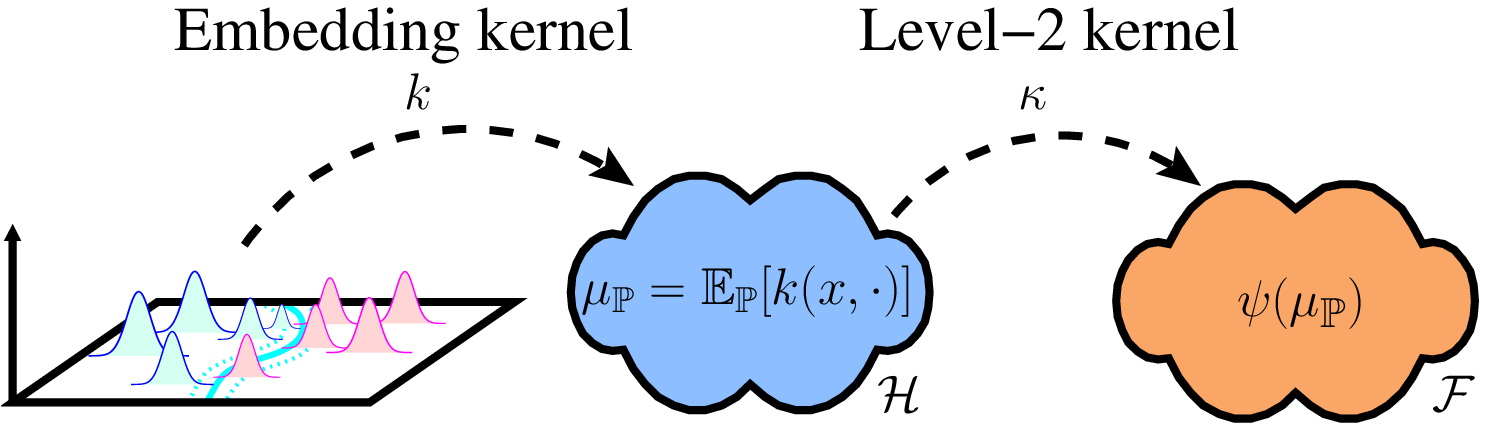}
  \caption{A visualization of a learning framework on distributional data. The embedding kernel $k$ defines feature representation for distributions, while the level-2 kernel $\kappa$ induces a class of non-linear functions over probability space based on such a representation.}
  \label{fig:dist-feature-map}
\end{figure}

Recently, the empirical kernel between distributions was employed in an unsupervised way for multi-task learning to generalize to a previously unseen task \citep{Blanchard11:Generalize}. Figure \ref{fig:dist-feature-map} summarizes the kernel-based framework for distributional data.

\subsection{Distributional Risk Minimization}

As also noted in \citet{Muandet12:SMM,Muandet15:Thesis} and \citet{Zoltan15:DistReg}, existing tools for theoretical analysis can be extended to a learning framework on distributional data. Let us assume that we have access to a training sample $(\pp{P}_1,y_1),\ldots,(\pp{P}_n,y_n) \in\pspace(\inspace)\times\outspace$ generated from some unknown distribution over $\pspace(\inspace)\times\outspace$, a strictly monotonically increasing function $\Omega:[0,\infty)\rightarrow\rr$, and a loss function $\ell:(\pspace(\inspace)\times\rr^2)\rightarrow\rr\cup\{+\infty\}$. \citet[Theorem 1]{Muandet12:SMM} shows that any function $\hat{f}\in\hbspace$ that minimizes a loss functional, called a \emph{distributional risk minimization} (DRM)
\begin{equation}
  \label{eq:drm}
  \ell(\pp{P}_1,y_1,\ep_{\pp{P}_1}[f],\ldots,\pp{P}_n,y_n,\ep_{\pp{P}_n}[f]) + \lambda\Omega(\|f\|_{\hbspace}),
\end{equation}
\noindent admits a representation of the form 
\begin{equation}
  \hat{f} = \sum_{i=1}^n\alpha_i\ep_{\x\sim\pp{P}_i}[k(\x,\cdot)] = \sum_{i=1}^n\alpha_i\muv_{\pp{P}_i}.
\end{equation}
Put differently, any solution $\hat{f}$ can be expressed in terms of the kernel mean embedding of $\pp{P}_1,\ldots,\pp{P}_n$. Note that if we restrict $\pspace(\inspace)$ to contain only Dirac measures $\delta_{\x}$, then the solution becomes $\hat{f}=\sum_{i=1}^n\alpha_ik(\x_i,\cdot)$, \ie, the classical representer theorem \citep{Scholkopf01:GRT}.

\begin{figure}[t!]
  \centering
  \includegraphics[width=\textwidth]{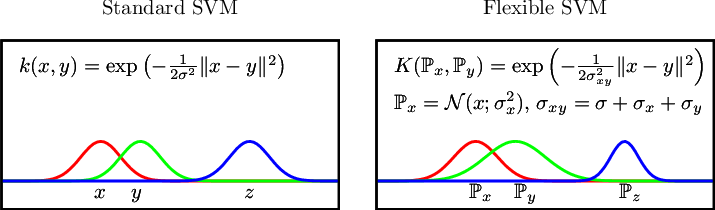}
  \caption{A picturial comparison between standard SVM algorithm and the flexible SVM algorithm. The flexible SVM allows us to put different kernel functions over training samples, as opposed to the standard SVM.}
  \label{fig:flexible-svm}
\end{figure}

On the one hand, it is instructive to compare the DRM in \eqref{eq:drm} to the vicinal risk minimization (VRM) of \citet{Chapelle00:Vicinal}. Specifically, they consider a slightly different regularized functional
\begin{equation}
  \label{eq:vrm}
  \ep_{\x_1\sim\pp{P}_1}\cdots\ep_{\x_n\sim\pp{P}_n}\left[\ell(\x_1,y_1,f(\x_1),\ldots,\x_n,y_n,f(\x_n)) + \lambda\Omega(\|f\|_{\hbspace}) \right].
\end{equation}
\noindent Intuitively, \eqref{eq:vrm} amounts to empirical risk minimization (ERM) on samples drawn from $\pp{P}_1,\ldots,\pp{P}_n$. Note that \eqref{eq:drm} and \eqref{eq:vrm} become equivalent only for a certain class of loss functional $\ell$. Arguably, \eqref{eq:vrm} is ultimately what we want to minimize when learning from distributional data. However, it is computationally expensive and is not suitable for some applications. On the other hand, one may consider the following regularized functional
\begin{equation}
  \label{eq:mrm}
  \ell(\mathbf{m}_1,y_1,f(\mathbf{m}_1),\ldots,\mathbf{m}_n,y_n,f(\mathbf{m}_n)) + \lambda\Omega(\|f\|_{\hbspace}) ,
\end{equation}
\noindent where $\mathbf{m}_i := \ep_{\x\sim\pp{P}_i}[\x]$. That is, \eqref{eq:mrm} corresponds to the ERM on the means of the distributions. Despite being more efficient, it throws away most information about high-level statistics. In some sense, the DRM can be viewed as something in between. Lastly, there is no specific assumption on the output space $\outspace$, making it applicable to binary, real-valued, structured, or even distributional outputs.

Based on the DRM, \citet{Muandet12:SMM} proposes the so-called \emph{support measure machine} (SMM) which is a variant of an SVM that operates on distributions rather than points, permitting modeling of input uncertainties. In the special case of Gaussian input distributions and SVMs with Gaussian kernel, the SMM leads to a multi-scale SVM---akin to an RBF network with variable bandwidths---which is still trained by solving a QP problem (see Figure \ref{fig:flexible-svm} for an illustration) \citep[Lemma 4]{Muandet12:SMM}. In \citet{Muandet12:SMM} and \citet{Poczos12:NonparamImg}, SVM on distributional data was applied to natural image categorization using bag-of-word (BoW) data representation, \ie, each image is viewed as a distribution over codewords. \citet{Yoshikawa14:LSMM} proposed \emph{latent SMM} which assumes that each codeword $t\in\mathcal{V}$ is represented by a $q$-dimensional latent vector $\x_t\in\rr^q$ which is learnt jointly with the SMM parameters (see also \citealp{Yoshikawa15:GPLVM} for a Gaussian process formulation). In unsupervised setting, one-class SMM (OCSMM) was studied in \citet{Muandet13:OCSMM} in connection with variable KDE and application in group anomaly detection. \citet{Guevara2014} considers another equivalent characterization of the one-class problem leading to the \emph{support measure data description} (SMDD).

The next theorem provides a generalization bound for a classification problem on distributional data using the kernel mean embedding.

\begin{shadowbox}
\begin{theorem}[\citealp{Lopez-Paz15:Towards}]
  \label{thm:dist-bound}
  Consider a class $\fspace_k$ of functionals mapping $\hbspace$ to $\rr$ with Lipschitz constants uniformly bounded by $L_{\fspace}$. Let $\rho: \rr\rightarrow\rr^+$ be a $L_{\rho}$-Lipschitz function such that $\rho(z) \geq \mathds{1}_{z>0}$. Let $\rho(-f(h)l) \leq B$ for every $f\in\fspace_k$, $h\in\hbspace$, and $l\in\mathcal{L}=\{-1,1\}$. Define a surrogate $\rho$-risk of any $f\in\fspace_k$ as
  \begin{equation*}
    R_\rho(f) = \ep_{(z,l)\sim P}[\rho(-f(z)l)].
  \end{equation*}
  \noindent Let $f^*\in\fspace_k$ be the minimizer of the expected $\rho$-risk $R_{\rho}(f)$ and $\tilde{f}_n$ be the minimizer of the empirical $\rho$-risk $\tilde{R}_{\rho}(f)$ evaluated over the sample $\{(\muh_{\pp{P}_i},l_i)\}_{i=1}^n$ where $\muh_{\pp{P}_i} := (1/n_i)\sum_{j=1}^{n_i}k(\x_j^{(i)},\cdot)$ and $\{\x_j^{(i)}\}_{j=1}^{n_i}\sim \pp{P}_i$. Then, with probability at least $1-\delta$,
  \begin{align*}
    R_{\rho}(\tilde{f}_n)-R_{\rho}(f^*) &\leq 4L_{\rho}R_n(\fspace_k) + 2B\sqrt{\frac{\log(2/\delta)}{2n}} \\
    & + \frac{4L_{\rho}L_{\fspace}}{n}\sum_{i=1}^n\left(\sqrt{\frac{\ep_{\pp{P}_i}[k(\x,\x)]}{n_i}}+\sqrt{\frac{\log(2n/\delta)}{2n_i}}\right),
  \end{align*}
  \noindent where $R_n(\fspace_k)$ denotes the Rademacher complexity of $\fspace_k$.
\end{theorem}
\end{shadowbox}

\noindent The bound reveals important ingredients for the convergence of the excess risk to zero, \ie, the consistency of the learning procedure on distributional data. Specifically, the upper bound converges to zero as both $n$ (the number of distributions) and $n_i$ (the size of the sample obtained from $\pp{P}_i$) tend to infinity, in such a way that $\sqrt{\log n}/n_i = o(1)$. Moreover, $n_i$ is only in the order of $\sqrt{\log n}$ for the second term of the rhs of the bound to be small. \citet{Zoltan15:DistReg} provides a detailed analysis of the ridge regression on distributional data.

Some recent works rely on a finite approximation of $\muh_{\pp{P}}$, \ie, the first step in Figure \ref{fig:dist-feature-map}. For example, \citet{Lopez-Paz15:Towards} proposes to solve a bivariate cause-effect inference between random variables $X$ and $Y$ as a classification problem on $\mu_{\pp{P}(X,Y)}$ approximated with random Fourier features \citep{Rahimi2007}. Inspired by the work of \citet{Eslami14:JIT}, \citet{Jitkrittum15:JIT} proposes Just-In-Time kernel regression for learning to pass expectation propagation (EP) messages. To perform the message passing efficiently, the authors use a two-stage random feature for $\kappa$, \ie, they first construct random Fourier features to approximate $\mu_{\pp{P}}$ on which the second set of random features can be generated to approximate \eqref{eq:l2-rbf}.

\subsubsection{Related Works on Learning from Distributions}

Several works have attempted to leverage information from distributions (or generative models such as hidden Markov models) over complex objects in discriminative models. \citet{Jaakkola98:Generative} provides a generic procedure for obtaining kernel functions from generative probability models. Given a generative probability model $\pp{P}(\x|\bm{\theta})$, where $\x$ is a data point and $\bm{\theta}$ is a vector of the model parameters, they define the \emph{Fisher kernel}
\begin{equation}
  K(\x_i,\x_j) = \mathbf{u}_i^\top \mathcal{I}^{-1}\mathbf{u}_j ,
\end{equation}
\noindent where $\mathbf{u}_i$ is the \emph{Fisher score vector} defined by $\mathbf{u}_i := \nabla_{\bm{\theta}}\log\pp{P}(\x_i|\bm{\theta})$ and $\mathcal{I}$ is the Fisher information matrix.\footnote{Intuitively, the gradient of log-likelihood $\mathbf{u}_i$ specifies how the parameter $\theta$ contributes to the process of generating the example $\x_i$ \citep{Jaakkola98:Generative}.} The benefit of the Fisher kernel is that the resulting discriminative models such as SVMs are well-informed about the underlying generative models such as GMMs and HMMs, which makes it useful for structured data such as biological data, logical sequences, and documents. Similarly, \citet{Jebara04:PPK} proposed the \emph{probability product kernel} (PPK)
\begin{equation}
  \label{eq:ppk-kernel}
  K_{\rho}(p,q)=\int_{\inspace}p(\x)^{\rho}q(\x)^{\rho}\dd \x,
\end{equation}
which is a generalized inner product between two input objects. That is, the probabilistic models $p$ and $q$ are learned for each example and used as a surrogate to construct the kernel between those examples. The kernel \eqref{eq:ppk-kernel} can be evaluated for all exponential families such as multinomials and Gaussians. Moreover, the PPK (with a certain value of $\rho$) can be computed analytically for some distributions such as mixture models, HMMs, and linear dynamical systems. For intractable models, \citet{Jebara04:PPK} suggested approximating \eqref{eq:ppk-kernel} by using structured mean-field approximations. As a result, the PPK can be applied for a broader class of generative models.

The PPK is in fact closely related to well-known kernels such as the Bhattacharyya kernel \citep{Bhattacharyya43Kernel} and the exponential symmetrized Kullback-Leibler (KL) divergence \citep{Moreno04KL}. In \citet{Hein05Hilbertian}, an extension of a two-parameter family of Hilbertian metrics of Tops\o e was used to define Hilbertian kernels on probability measures. In \citet{Cuturi05SKM}, the semi-group kernels were designed for objects with additive semi-group structure such as positive measures. Recently, \citet{Martins09NIT} introduced non-extensive information theoretic kernels on probability measures based on new Jensen-Shannon-type divergences. Although these kernels have proven successful in many applications, they are designed specifically for certain properties of distributions and application domains. Moreover, there has been no attempt in making a connection to the kernels on corresponding input spaces.

The kernel function $K(\pp{P},\pp{Q}) = \langle \muv_{\pp{P}},\muv_{\pp{Q}}\rangle_{\hbspace}$ considered earlier can in fact be understood as a special case of the Hilbertian metric \citep{Hein05Hilbertian} with the associated kernel
\begin{equation}
  \label{eq:hilbertian-metric}
  K(\pp{P},\pp{Q}) = \mathbb{E}_{\x\sim\pp{P},\tilde{\x}\sim\pp{Q}}[k(\x,\tilde{\x})],
\end{equation}
\noindent and a generative mean map kernel (GMMK) proposed by \citet{Mehta10Generative}. In the GMMK, the kernel between two objects $\x$ and $\y$ is defined via $\hat{p}_\x$ and $\hat{p}_\y$, which are estimated probabilistic models of $\x$ and $\y$, respectively. Moreover, the empirical version of \eqref{eq:hilbertian-metric} and \eqref{eq:dist-kernel} coincides with that of classical set kernel of \citet{Haussler99:Convolution} and \citet{Gaertner02:MIK} (see also \citealp{Kondor03:SetKernels}). This connection reveals an interesting fact that---although originally presented as a similarity measure between sets of vectors---this class of kernels has a natural interpretation as the similarity between the underlying probability distributions.

It has been shown that the PPK is a special case of GMMK when $\rho=1$ \citep{Mehta10Generative}. Consequently, GMMK, PPK with $\rho=1$, and linear kernel $\langle\muv_{\pp{P}},\muv_{\pp{Q}}\rangle_{\hbspace}$ are equivalent when the embedding kernel is $k(\x,\x')=\delta(\x-\x')$.

Distributions can be used to capture input uncertainty. The use of the kernel \eqref{eq:dist-kernel} in dealing with the uncertainty in input data has a connection to robust SVMs. For example, a generalized form of the SVM in \citet{Shivaswamy06SOCP} incorporates the probabilistic uncertainty into the maximization of the margin. This results in a second-order cone programming (SOCP) that generalizes the standard SVM. In SOCP, one needs to specify the parameter $\tau_i$ that reflects the probability of correctly classifying the $i$th training example. In the context of this section, we may represent data point $\x_i$ by a distribution $\mathcal{N}(\x_i,\sigma_i^2\id)$. Therefore, the parameter $\tau_i$ is closely related to the parameter $\sigma_i$, which specifies the variance of the distribution centered at the $i$th example. \citet{Anderson11Missing} showed the equivalence between SVMs using expected kernels and SOCP when $\tau_i=0$. When $\tau_i > 0$, the mean and covariance of missing kernel entries have to be estimated explicitly, making the SOCP more involved for nonlinear kernels. Although achieving comparable performance to the standard SVM with expected kernels, the SOCP requires a more computationally extensive SOCP solver, as opposed to simple quadratic programming (QP). From a Bayesian perspective, the problem of regression with uncertain inputs has been studied in the Gaussian process community, e.g., \citet{GirardRCM02:GP}.

A major drawback of previous studies is that they usually impose a strong \emph{parametric assumption} on the form of probability distribution. The kernel mean representation, on the other hand, allows one to learn directly from distributions without such an assumption. For example, \citet{Zoltan15:DistReg} has recently studied the \emph{nonparametric} distribution regression problem based on kernel mean embedding and the kernel ridge regression algorithm. They establish the consistency and convergence rate of the resulting algorithm whose challenge arises from the \emph{two-stage sampling}: a meta distribution generates an i.i.d. sample of distributions from which i.i.d observations have been generated (see also Theorem \ref{thm:dist-bound}). As a result, in practice we only observe samples from the distributions rather than the distributions themselves. The theoretical analysis uses the results of \citet{Caponnetto07:OptRate} who provides error bounds for the regularized least-squares algorithm in the standard setting.

In addition to the mean embedding approach, another line of research employs kernel density estimation (KDE) to perform regression on distributions with consistency guarantee (under the assumption that the true regressor is H\"older continuous, and the meta distribution have finite doubling dimension \citep{Kpotufe11:KNN}) \citep{Poczos13:DistFree,Oliva14:FastD2R}. In this case the covariates are nonparametric continuous distributions on $\rr^d$ and the outputs are real-valued. \citet{Oliva13:D2D} also considers the case when the output is also a distribution. The basic idea is to approximate the density function by KDE and then apply kernels on top of it. Unlike the mean embedding approach, the kernels used are classical smoothing kernels and not the reproducing kernel. Although the parametric assumption is not needed, drawbacks of the KDE-based approach are that the convergence rate is slow in high-dimensional space and it is not applicable to learning over structured data such as documents, graphs, and permutations. The use of kernel mean embedding allows us to deal with any kind of data as long as the positive definite kernel on such data is well-defined.

\section{Recovering Information from Mean Embeddings}
\label{sec:dist-preimage} 
  
Given a kernel mean embedding $\muv_{\pp{P}}$, can we recover essential properties of $\pp{P}$ from $\muv_{\pp{P}}$? We respond to this question by discussing two closely related problems, namely, the distributional pre-image problem\footnote{We call this a \emph{distributional pre-image} problem to distinguish it from the classical setting which does not involve probability distributions.} \citep{Kwok2004,Song08:TDE,Kanagawa2014} and kernel herding \citep{Chen2010}. We consider these two problems to be related because both of them involve finding objects in the input space which correspond to a specific kernel mean embedding in the feature space.

The classical pre-image problem in kernel methods involves finding patterns in input space that map to specific feature vectors in the feature space \citep[Chapter 18]{Scholkopf01:LKS}. Recovering a pre-image is considered necessary in some applications such as image denoising using kernel PCA \citep{Kwok2004,Kim2005} and visualizing the clustering solutions of a kernel-based clustering algorithm \citep{Dhillon04:KKS,Jegelka2009}. Moreover, it can be used as a reduced set method to compress a kernel expansion \citep[Chapter 18]{Scholkopf01:LKS}. \citet[Proposition 18.1]{Scholkopf01:LKS} shows that if the pre-image exists and the kernel is an invertible function of $\langle \x,\x'\rangle$, the pre-image will be easy to compute. Unfortunately, the exact pre-image typically does not exist, and the best one can do is to approximate it. There is a fair amount of work on this topic and the interested readers should consult \citet[Chapter 18]{Scholkopf01:LKS} for further details.

\subsection{Distributional Pre-Image Problem}

Likewise, in some applications of kernel mean embedding, it is important to recover the meaningful information of an underlying distribution from an estimate of its embedding. In a state-space model, for example, we typically obtain a kernel mean estimate of the predictive distribution from the algorithm \citep{Song10:KCOND,Nishiyama12:POMDPs,McCalman2013}. To obtain meaningful information, we need to extract the information of $\pp{P}$ from the estimate. Unfortunately, in these applications we only have access to the estimate $\muh_{X}$ which lives in a high-dimensional feature space.

The idea is similar to the \emph{approximate pre-image problem}. Let $\pp{P}_{\bm{\theta}}$ be an arbitrary distribution parametrized by $\bm{\theta}$ and $\muv_{\pp{P}_{\bm{\theta}}}$ be its mean embedding in $\hbspace$. One can find $\pp{P}_{\bm{\theta}}$ by the following minimization problem
\begin{equation}
  \label{eq:dist-pre-image}
  \bm{\theta}^* = \arg\min_{\bm{\theta}\in\Theta}\; \left\|  \muh_X - \muv_{\pp{P}_{\bm{\theta}}}\right\|^2_{\hbspace} ,
\end{equation}
\noindent subject to appropriate constraints on the parameter vector $\bm{\theta}$. Note that if $\pp{P}_{\bm{\theta}} = \delta_{\x}$ for some $\x\in\inspace$, the distributional pre-image problem \eqref{eq:dist-pre-image} reduces to the classical pre-image problem. The pre-image $\x$ can be viewed as a \emph{point estimate} of the underlying distribution, but may not correspond to the MAP estimate.

Another example is a mixture of Gaussians $\pp{P}_{\bm{\theta}} = \sum_{i=1}^m\pi_i\mathcal{N}(\mathbf{m}_i,\sigma^2_i\id)$ where the parameter $\bm{\theta}$ consists of $\{\pi_1,\ldots,\pi_m\}$, $\{\m_1,\ldots,\m_m\}$, and $\{\sigma_1,\ldots,\sigma_m\}$. It is required that $\sum_{i=1}^m\pi_i = 1$ and $\sigma_i \geq 0$. Let assume that $\muh_{X} = \sum_{i=1}^n\beta_i\phi(\x_i)$ for some $\bvec\in\rr^n$. In this case, the optimization problem \eqref{eq:dist-pre-image} reduces to
\begin{equation}
  \label{eq:dist-quadratic}
  \bm{\theta}^* = \arg\min_{\bm{\theta}\in\Theta}\; \bvec^\top\kmat\bvec - 2\bvec^\top\mathbf{Q}\bm{\pi} + \bm{\pi}^\top\mathbf{R}\bm{\pi} \, ,
\end{equation}
\noindent where
\begin{eqnarray*}
  \kmat_{ij} &=& k(\x_i,\x_j) \\
  \mathbf{Q}_{ij} &=& \int k(\x_i,\x')\dd\mathcal{N}(\x';\m_j,\sigma_j^2\id) \\
  \mathbf{R}_{ij} &=& \int\int k(\x,\x')\dd\mathcal{N}(\x;\m_i,\sigma_i^2\id)\dd\mathcal{N}(\x';\m_j,\sigma_j^2\id) \,.
\end{eqnarray*}
\noindent Note that \eqref{eq:dist-quadratic} is quadratic in $\bm{\pi}$ and is also convex in $\bm{\pi}$ as $\kmat$, $\mathbf{Q}$, and $\mathbf{R}$ are positive definite. The integrals $\mathbf{Q}_{ij}$ and $\mathbf{R}_{ij}$ can be evaluated in closed form for some kernels (see \citealp[Table 1]{Song08:TDE} and \citealp[Table 1]{Muandet12:SMM}). Unfortunately, the problem is often non-convex in both $\m_i$ and $\sigma_i$, $i=1,\ldots,m$. A derivative-free optimization is often used to find these parameters. In practice, $\pi_i$ and $\{\m_i,\sigma_i\}$ are solved alternately until convergence (see, \eg, \citealp{Song08:TDE,Chen2010}).

The reduced set problem is slightly more general than the pre-image problem because we do not just look for single pre-images, but for expansions of several input vectors. Interestingly, we may view the reduced set problem as a specific case of the distributional pre-image problem. To understand this, assume we are given a function $g\in\hbspace$ as a linear combination of the images of input points $\x_i\in\inspace$, \ie, $g = \sum_{i=1}^n\alpha_i\phi(\x_i)$. The function $g$ is exactly the kernel mean embedding of the finite signed measure $\nu=\sum_{i=1}^n\alpha_i\delta_{\x_i}$ whose supports are the points $\x_1,\ldots,\x_n$. That is, $g = \int \phi(\y)\dd\nu(\y)$. Given the reduced set vector $\z_1,\ldots,\z_m$ where $m\ll n$, the reduced set problem amounts to finding another finite signed measure $\muv=\sum_{j=1}^m\beta_j\phi(\z_j)$ whose supports are $\z_1,\ldots,\z_m$ that approximates well the original measure $\nu$. From the distributional pre-image problem, the reduced set methods can be viewed as an approximation of a finite signed measure by another signed measure whose supports are smaller.

Although it is possible to find a distributional pre-image, it is not clear what kind of information of $\pp{P}$ this pre-image represents. \citet{Kanagawa2014} considers the recovery of the information of a distribution from an estimate of the kernel mean when the Gaussian RBF kernel on Euclidean space is used. Let $\muh_{\pp{P}} = \sum_{i=1}^n w_i k(\cdot,X_i)$ be a consistent estimator of the kernel mean $\muv_{\pp{P}}$, \ie, $\lim_{n \to \infty} \| \muh_{\pp{P}} - \muv_{\pp{P}} \|_{\hbspace} = 0$. Recall that if $f\in\hbspace$, the weighted average
$$\sum_{i=1}^n w_i f(X_i) = \langle f, \sum_{i=1}^n w_i k(\cdot,X_i)\rangle_{\hbspace} = \langle f,\muh_{\pp{P}}\rangle_{\hbspace}$$
\noindent converges to the expectation $\ep_{X \sim \pp{P}}[f(X)] = \langle f, \muv_{\pp{P}}\rangle_{\hbspace}$. Note that the expectation $\ep_{X \sim \pp{P}}[f(X)]$ contains information about the distribution $\pp{P}$. This implies that by choosing an appropriate function $f$, the weighted average $\sum_{i=1}^n w_i f(X_i)$ can be used as a way of recovering the information about the distribution $\pp{P}$.

Based on this idea, \citet{Kanagawa2014} proposes methods for recovering certain statistics of $\pp{P}$, namely its moments and measures on intervals, from $\muh_{\pp{P}}$, and also a nonparametric estimator of the density of $\pp{P}$ using $\muh_{\pp{P}}$. They justify these estimators by arguing that the weighted average of function $f$ in some \emph{Besov space} converges to the expectation of $f$, \ie, $\sum_i w_i f(X_i) \to \ep_{X \sim \pp{P}}[f(X)]$ \citep[Theorem 1]{Kanagawa2014}.\footnote{The Besov space is a complete quasinormed space,  which also coincides with the more classical Sobolev spaces for certain parameter settings. For details of Besov spaces, see \citet[Chapter 7]{Adams03:Sobolev}.} This result is a generalization of the known result for functions in an RKHS. Unfortunately, their main result, \ie, Theorem 1 in \citet{Kanagawa2014}, has a mistake, so the entire results in the paper do not provide any guarantee (M. Kanagawa, personal communication, April 6, 2017).\footnote{The mistake was made in the derivation of the upper bound in Eq. (16) on page 463 in \citet{Kanagawa2014}. In the derivation, the equality is wrong: the reproducing property is not applicable here, because $\muv_{\pp{P}}$ and $\muh_{\pp{P}}$ are defined in terms of the original Gaussian kernel, whereas the RKHS is that of a Gaussian kernel with smaller bandwidth. See also a note on Kanagawa's web page \url{https://sites.google.com/site/motonobukanagawa/publications}.}

Subsequently, \citet{Kanagawa16:ConvMiss} has provided theoretical guarantees for general RKHSs, \ie, Theorem 1 in \citet{Kanagawa16:ConvMiss}. Their main assumption is that the integrand $f$ belongs to a certain power of the RKHS $\hbspace$ defined via Mercer's theorem, which is an interpolation space between the RKHS $\hbspace$ and $L_2(\inspace,\pp{P})$. This effectively expresses a \emph{misspecified setting}, \ie, the situation where $f\notin\hbspace$. Nevertheless, they prove that, under certain assumptions, the weighted average $\sum_{i=1}^n w_i f(X_i)$ still converges to the expectation $\ep_{X \sim \pp{P}}[f(X)]$. For concrete examples of the power of RKHS, we refer to \citet{Kanagawa16:ConvMiss}.

\subsection{Kernel Herding}

Instead of finding a distributional pre-image of the mean embedding, another common application is obtaining a sample from the distribution or sampling. To avoid an intractable learning of a joint probabilistic model and a direct sampling from that model, \citet{Welling2009} proposes a novel approach called the \emph{Herding} algorithm. Herding directly generates \emph{pseudo-samples} in a fully deterministic fashion and in a way that asymptotically matches the empirical moments of the data. Afterwards, \citet{Chen2010} proposes a kernel herding algorithm that extends the herding algorithm \citep{Welling2009,Welling2009a,Welling2010} to continuous spaces by using the kernel trick. Herding can be understood concisely as a weakly chaotic non-linear dynamical system $\w_{t+1} = F(\w_t)$. They re-interpret herding as an infinite memory process in the state space $\x$ by marginalizing out the parameter $\w$, resulting in a mapping $\x_{t+1} = G(\x_1,\ldots,\x_t; \w_0)$. Under some technical assumptions, herding can be seen to greedily minimize the squared error
\begin{equation}
  \label{eq:herding-criterion}
  \mathcal{E}^2_T := \left\|\muv_{\pp{P}} - \frac{1}{T}\sum_{t=1}^T\phi(\x_t)\right\|^2_{\hbspace}
  = \left\|\muv_{\pp{P}} - \muh_T\right\|^2_{\hbspace} ,
\end{equation}
\noindent where $\muh_T$ denotes the empirical mean embedding obtained from herding. Following the result of \citet{Welling2009}, kernel herding is shown to decrease the error of expectations of functions in the RKHS at a rate of $O(1/T)$ as opposed to the random samples whose rate is $O(1/\sqrt{T})$. The fast rate is guaranteed even when herding is carried out with some error. This condition is reminiscent of the Boosting algorithm and perceptron cycling theorem \citep[Corollary 2]{Chen2010}. The fast convergence is due to the \emph{negative autocorrelation}, \ie, herding tends to find samples in an unexplored high-density region. This kind of behaviour can also be observed in Quasi Monte Carlo integration and Bayesian quadrature methods \citep{RasmussenG02:MonteCarlo}.

\citet{Huszar2012} also investigates the kernel herding problem and suggests a connection between herding and Bayesian quadrature. Bayesian quadrature (BQ) \citep{RasmussenG02:MonteCarlo} estimates the integral $Z = \int f(\x)p(\x)\dd \x$ by putting a prior distribution on $f$ and then inferring a posterior distribution over $f$ conditioned on the observed evaluations. An estimate of $Z$ can be obtained by a posterior expectation, for example. The sampling strategy of BQ is to select the sample so as to minimize the posterior variance. \citet{Huszar2012} shows that the posterior variance in BQ is equivalent to the criterion \eqref{eq:herding-criterion} minimized when selecting samples in kernel herding. An advantage of Bayesian interpretation of herding is that the kernel parameters can be chosen by maximizing the marginal likelihood. As an aside, it is instructive to note that the herding sample exhibits similar behaviour to those defined under Determinantal Point Process (DPP) prior, \ie, a DPP assigns a high probability to sets of items that are \emph{repulsive} \citep{Kulesza12:DPPML}. Like kernel herding, a DPP prior is characterized via a marginal kernel $\kmat$. For a subset $A\subset\mathcal{X}$, it assigns a probability proportional to $\mathrm{det}(\kmat_A)$ where $\kmat_A$ denotes a restriction of $\kmat$ to the elements of $A$. Important applications of DPP are, for example, information retrieval and text summarization.

Herding can be problematic in a high dimensional setting when optimizing over the new sample. \citet{Bach2012} also pointed out that the fast convergence rate is not guaranteed in an infinite dimensional Hilbert space. To alleviate this issue, \citet{Bach2012} shows that the herding procedure of \citet{Welling2009} takes the form of a convex optimization algorithm in which convergence results can be invoked. \citet{LacosteJulien15:FrankWolfe} takes this interpretation and proposes the Frank-Wolfe optimization algorithm for particle filtering. Lastly, it is instructive to mention that kernel herding in RKHSs has an implicit connection to Quasi-Monte Carlo (QMC) theory, \ie, one can show by the reproducing property and the Cauchy-Schwartz inequality that the integration error
\begin{equation*}
  \mathcal{E}_{p,S}(f) := \left| \int_{\rr^d}f(\x)p(\x)\dd\x - \frac{1}{s}\sum_{\mathbf{w}\in S}f(\mathbf{w})\right|
  \leq \|f\|_{\hbspace}D(S) , 
\end{equation*}
\noindent where $D(\cdot)$ is a discrepancy measure:
\begin{equation}
  D(S) := \left\|\mu_p - \frac{1}{s}\sum_{\mathbf{w}\in S}k(\mathbf{w},\cdot)\right\|^2_{\hbspace} .
\end{equation}
This connection has been noted by several authors including \citet{Bach2012} and \citet{Yang14:QMC}.



\chapter{Hilbert Space Embedding of Conditional Distributions}
\label{sec:conditional-embedding}
 
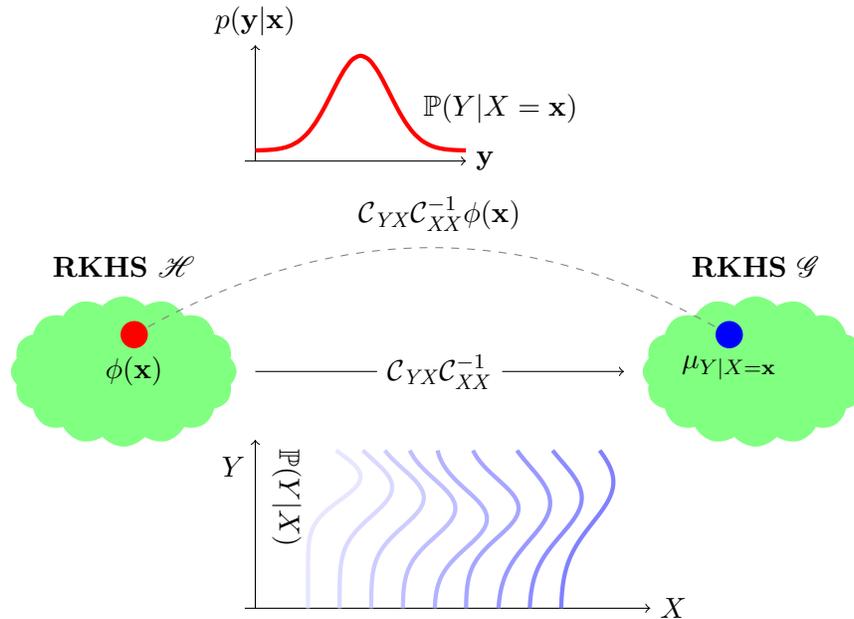
\begin{figure}[t!]
  \centering
  \begin{tikzpicture}[scale=0.7]
      \draw[->] (-0.2,0.5) -- (7.5,0.5) node[right] {$X$};
      \draw[->] (0,0.5) -- (0,3.7) node[left,yshift=-10] {$Y$};
      \draw[rotate=-90,color=blue!10,smooth,domain=-3.5:-0.5,ultra thick] plot (\x,{1+exp(-(3+\x)*(3+\x)*0.5/0.2)});
      \draw[rotate=-90,color=blue!15,smooth,domain=-3.5:-0.5,ultra thick] plot (\x,{1.6+exp(-(2.8+\x)*(2.8+\x)*0.5/0.3)});
      \draw[rotate=-90,color=blue!20,smooth,domain=-3.5:-0.5,ultra thick] plot (\x,{2.2+exp(-(2.6+\x)*(2.6+\x)*0.5/0.3)});
      \draw[rotate=-90,color=blue!25,smooth,domain=-3.5:-0.5,ultra thick] plot (\x,{2.8+exp(-(2.4+\x)*(2.4+\x)*0.5/0.3)});
      \draw[rotate=-90,color=blue!30,smooth,domain=-3.5:-0.5,ultra thick] plot (\x,{3.4+exp(-(2.2+\x)*(2.2+\x)*0.5/0.3)});
      \draw[rotate=-90,color=blue!35,smooth,domain=-3.5:-0.5,ultra thick] plot (\x,{4+exp(-(2.4+\x)*(2.4+\x)*0.5/0.3)});
      \draw[rotate=-90,color=blue!40,smooth,domain=-3.5:-0.5,ultra thick] plot (\x,{4.6+exp(-(2.5+\x)*(2.5+\x)*0.5/0.5)});
      \draw[rotate=-90,color=blue!45,smooth,domain=-3.5:-0.5,ultra thick] plot (\x,{5.2+exp(-(2.6+\x)*(2.6+\x)*0.5/0.5)});
      \draw[rotate=-90,color=blue!50,smooth,domain=-3.5:-0.5,ultra thick] plot (\x,{5.8+exp(-(2.9+\x)*(2.9+\x)*0.5/0.6)});
      \node [rotate=-90] at (0.6,2.6) {$\pp{P}(Y|X)$};


      \coordinate (P1) at (-2.3,5.7);
      \coordinate (P2) at (9,5.7);



      \node [cloud, fill=green!50, cloud puffs=16, cloud puff arc= 100,
      minimum width=3cm, minimum height=2cm, aspect=1] at (-2.5,5) {};
      \node [] at (-2.5,7) {\textbf{RKHS} $\hbspace$};

      \node [cloud, fill=green!50, cloud puffs=16, cloud puff arc= 100,
      minimum width=3cm, minimum height=2cm, aspect=1] at (9.5,5) {};
      \node [] at (9.5,7) {\textbf{RKHS} $\hbspg$};


      \node at (3.5,8) {$\covyx\covx^{-1}\phi(\x)$};
      \draw[color=gray,dashed,->] (P1) to[out=30,in=150] (P2);

      \node [fill=blue,circle,minimum width=0.05cm] at (P2) {};
      \node [below,yshift=-0.15cm] at (P2) {$\muv_{Y|X=\x}$};
      \node [fill=red,circle,minimum width=0.05cm] at (P1) {};
      \node [below,yshift=-0.15cm] at (P1) {$\phi(\x)$};

      \coordinate (n1) at (0,5);
      \coordinate (n2) at (7,5);
      \path (n1) -- node[sloped] (text) {$\covyx\covx^{-1}$} (n2);
      \draw[->] (n1)--(text)--(n2) node[right] {};
 

      \draw[->] (-0.2,9) -- (4,9) node[right] {$\y$};
      \draw[->] (0,9) -- (0,11.2) node[above] {$p(\y|\x)$};
      \draw[color=red,smooth,domain=0:4,ultra thick] plot (\x,{9.2+1.8*exp(-(\x-2)*(\x-2)*0.5/0.3)});
      \node [right] at (3,10) {$\pp{P}(Y|X=\x)$};

    \end{tikzpicture}
  \caption{From marginal distribution to conditional distribution: unlike the embeddings discussed in the previous section, the embedding of conditional distribution $\pp{P}(Y|X)$ is not a single element in the \gls{rkhs}. Instead, it may be viewed as a family of Hilbert space embeddings of the conditional distributions $\pp{P}(Y|X=\x)$ indexed by the conditioning variable $X$. In other words, the conditional mean embedding can be viewed as an operator mapping from $\hbspace$ to $\hbspg$. We will see in \S\ref{sec:regression-view} that there is a natural interpretation in a vector-valued regression framework.}
  \label{fig:conditional-embedding}
\end{figure}

In the previous section, we discussed the embedding of marginal distributions in a RKHS and gave comprehensive reviews of various applications. In this section we will extend the concept of kernel mean embedding to a \emph{conditional distribution} $\pp{P}(Y|X)$ and $\pp{P}(Y|X=\x)$ for some $\x\in\inspace$ \citep{Song10:KCOND,Song2013}. Unlike the marginal distribution $\pp{P}(X)$, the conditional distribution $\pp{P}(Y|X)$ captures the functional relationship between two random variables, namely, $X$ and $Y$. Hence, the conditional mean embedding extends the capability of kernel mean embedding to model more complex dependence in various applications such as dynamical systems \citep{Song10:KCOND,Boots2013}, Markov decision processes and reinforcement learning \citep{Grunewalder12:MDPs,Nishiyama12:POMDPs}, latent variable models \citep{Song10:NTGM,Song11:KBP,Song11:LTGM}, the kernel Bayes' rule \citep{Fukumizu11:KBR}, and causal discovery \citep{Janzing2011,Sgouritsa2013,Chen2014}. Figure \ref{fig:conditional-embedding} gives a schematic illustration of conditional mean embedding.
   
\section{From Marginal to Conditional Distributions}  
\label{sec:marginal2conditional} 

To better understand the distinction between the kernel mean embedding of marginal and conditional distributions, and the problems that we may encounter, we briefly summarize the concept of marginal, joint, and conditional distributions. Further details are available in most statistics textbooks, see, \eg, \citet{Wasserman2010}. Readers already familiar with these concepts may wish to proceed directly to the definition of conditional mean embedding.

Given two random variables $X$ and $Y$, probabilities defined on them may be marginal, joint, or conditional. Marginal probabilities $\pp{P}(X)$ and $\pp{P}(Y)$ are the (unconditional) probabilities of an event occurring. For example, if $X$ denotes the level of cloudyness of the outside sky, $\pp{P}(X)$ describes how likely it is for the outside sky to be cloudy. Joint probability $\pp{P}(X,Y)$ is the probability of event $X=x$ and $Y=y$ occurring. If $Y$ indicates whether or not it is raining, the joint distribution $\pp{P}(X,Y)$ gives the probability that it is both raining and cloudy outside. As we can see, joint distributions allow us to reason about the relationship between multiple events, which in this case are cloudyness and rain. Following the above definitions, one may subsequently ask given that it is cloudy outside, \ie, $X=\texttt{cloudy}$, what is the probability that it is also raining? The conditional distribution $\pp{P}(Y|X)$ governs such a question. Formally, the conditional probability $\pp{P}(Y=y|X=x)$ is the probability of event $Y=y$ occurring, given that event $X=x$ has occurred. In other words, conditional probabilities allow us to reason about causality.\footnote{To be more precise, the fundamental question in causal inference/discovery from observational data is to identify conditions under which $\pp{P}(Y\,|\,\text{do}(X=\x))$ is equal to $\pp{P}(Y\,|\,X=\x)$ where $\text{do}(X=\x)$ denotes the operation of setting the value of $X$ to be equal to $\x$ \citep{Pearl2000}. Under such conditions, one is allowed to make a causal claim from the conditional distribution $\pp{P}(Y\,|\,X=\x)$.}

The basic relationships between marginal, joint, and conditional distributions can be illustrated via the following equations:
\begin{equation}
  \label{eq:bayes-rule}
  \pp{P}(Y|X) = \frac{\pp{P}(X,Y)}{\pp{P}(X)} = \frac{\pp{P}(X|Y)\pp{P}(Y)}{\pp{P}(X)} .
\end{equation}
As we can see in the first equation of \eqref{eq:bayes-rule}, the conditional probability of $Y$ given $X$ is equal to the joint probability of $X$ and $Y$ divided by the marginal probability of $X$. Marginal, joint, and conditional distributions equipped with the formulation \eqref{eq:bayes-rule} provide a powerful language for statistical inference in statistics and machine learning.

\subsection{Conditional Mean Embeddings}

Suppose $\inspace$ and $\iny$ be measurable spaces and let $X$ and $Y$ be random variables taking values in $\inspace$ and $\iny$ respectively. Assume
$k : \inspace\times\inspace\rightarrow\rr$ and $l : \iny\times\iny \rightarrow \rr$ to be positive definite kernels with corresponding RKHS's $\hbspace$ and $\hbspg$. Let $\mathcal{U}_{Y|X}:\hbspace\rightarrow\hbspg$ and $\mathcal{U}_{Y|\x}\in\hbspg$ be conditional mean embeddings of the conditional distribution $\pp{P}(Y|X)$ and $\pp{P}(Y|X=\x)$, respectively, such that they satisfy
\begin{eqnarray}
  \mathcal{U}_{Y|\x} &=& \ep_{Y|\x}[\varphi(Y)|X=\x] = \mathcal{U}_{Y|X}k(\x,\cdot) \label{eq:cond1} \\
  \ep_{Y|\x}[g(Y)|X=\x] &=& \langle g,\mathcal{U}_{Y|\x} \rangle_{\hbspg} \label{eq:cond2}, \quad \forall g\in\hbspg .
\end{eqnarray}
Note that $\mathcal{U}_{Y|X}$ is an operator from $\hbspace$ to $\hbspg$, whereas $\mathcal{U}_{Y|\x}$ is an element in $\hbspg$. As an interpretation, condition \eqref{eq:cond1} says that the conditional mean embedding of $\pp{P}(Y|X=\x)$ should correspond to the conditional expectation of the feature map of $Y$ given that $X=\x$ (as in the marginal embedding). Moreover, the embedding operator $\mathcal{U}_{Y|X}$ represents the \emph{conditioning operation} that when applied to $\phi(\x)\in\hbspace$ outputs the conditional mean embedding $\mathcal{U}_{Y|\x}$ (see also Figure \ref{fig:conditional-embedding}). Condition \eqref{eq:cond2} ensures the reproducing property of $\mathcal{U}_{Y|\x}$, \ie, it should be a representer of conditional expectation in $\hbspg$ \wrt $\pp{P}(Y|X=\x)$ (as in the marginal embedding).

The following definition provides explicit forms of $\mathcal{U}_{Y|X}$ and $\mathcal{U}_{Y|\x}$.

\begin{shadowbox}
\begin{definition}[\citealp{Song10:KCOND,Song2013}]
  \label{def:cond-mean}
  Let $\covx : \hbspace\rightarrow\hbspace$ and $\covyx:\hbspace\rightarrow\hbspg$ be the covariance operator of $X$ and cross-covariance operator between $X$ to $Y$, given in \eqref{eq:xcov-optr}, respectively. Then, the conditional mean embedding $\mathcal{U}_{Y|X}$ and $\mathcal{U}_{Y|\x}$ are defined as
  \begin{eqnarray}
    \mathcal{U}_{Y|X} & := & \covyx\covx^{-1} \label{eq:cond-emb-1}\\
    \mathcal{U}_{Y|\x} & := & \covyx\covx^{-1}k(\x,\cdot) . \label{eq:cond-emb-2}
  \end{eqnarray}
\end{definition}
\end{shadowbox} 

Under the assumption that $\ep_{Y|X}[g(Y)|X]\in\hbspace$, \citet{Song10:KCOND} shows that the conditional mean embedding given in Definition \ref{def:cond-mean} satisfies both \eqref{eq:cond1} and \eqref{eq:cond2}. This result follows from \citet[Theorem 2]{Fukumizu04:DRS} which is also given in Theorem \ref{thm:fukumizu:thm} in this survey. Recall from Theorem \ref{thm:fukumizu:thm} that for any $g\in\hbspg$
\begin{equation}
  \label{eq:fukumizu:thm}
  \covx\ep_{\mathit{Y|X}}[g(Y)|X=\cdot] = \covxy g . 
\end{equation} 
\noindent For some $\x\in\inspace$, we have by virtue of reproducing property that $\ep_{Y|\x}[g(Y)|X=\x] = \langle \ep_{Y|X}[g(Y)|X], k(\x,\cdot)\rangle_{\hbspace}$. Using relation \eqref{eq:fukumizu:thm} and taking the conjugate transpose of $\covx^{-1}\covxy$ yields $\ep_{Y|\x}[g(Y)|X=\x] = \langle g,\covyx\covx^{-1}k(\x,\cdot)\rangle_{\hbspg} = \langle g,\mathcal{U}_{Y|\x}\rangle_{\hbspg}$.

One should keep in mind that, unlike the marginal mean embedding, the operator $\covyx\covx^{-1}$ may not exist in the continuous domain because the assumption that $\ep_{Y|X}[g(Y)|X=\cdot]$ is an element of $\hbspace$ for all $g\in\hbspg$ may not hold in general \citep{Fukumizu04:DRS,Song10:KCOND}.\footnote{For example, when $\hbspace$ and $\hbspg$ are both RKHSs associated with Gaussian kernels, and $X$ and $Y$ are independent, $\ep_{Y|X}[g(Y)|X=\x]$ is a constant function of $\x$, which is not contained in the Gaussian RKHS \citep[Corollary 4.44]{Steinwart08:SVM}. If $X$ and $Y$ are discrete random variables and the kernels are characteristic, then $\ep_{Y|X}[g(Y)|X=\cdot]\in\hbspace$.} This technical issue can be circumvented by resorting to a regularized version of \eqref{eq:cond-emb-2}, \ie, $\covyx(\covx + \lambda\id)^{-1}k(\x,\cdot)$ where $\lambda > 0$ denotes a regularization parameter. \citet[Theorem 8]{Fukumizu13:KBR} showed that---under some mild conditions---its empirical estimator is a consistent estimator of $\ep_{Y|X}[g(Y)|X=\x]$.

\begin{shadowbox}
\begin{theorem}[{\citealp[Eq. 6]{Song10:KCOND}, \citealp[Theorem 2]{Fukumizu13:KBR}}]
  \label{thm:cond-existence}
  Let $\mu_{\Pi}$ and $\mu_{Q_y}$ be the kernel mean embeddings of $\Pi$ in $\hbspace$ and $Q_y$ in $\hbspg$, respectively. Let $\mathcal{R}(\covx)$ denote the range space of a covariance operator $\covx$. If $\covx$ is injective, $\mu_{\Pi}\in\mathcal{R}(\covx)$, and $\ep[g(Y)|X=\cdot]\in\hbspace$ for any $g\in\hbspg$, then
  \begin{equation*}
    \mu_{Q_y} = \covyx\covx^{-1}\mu_{\Pi} ,
  \end{equation*}
  \noindent where $\covx^{-1}\mu_{\Pi}$ denotes the function mapped to $\mu_{\Pi}$ by $\covx$.
\end{theorem}
\end{shadowbox}

\subsection{Empirical Estimation of Conditional Mean Embeddings}

Since the joint distribution $\pp{P}(X,Y)$ is unknown in practice, we cannot compute $\covx$ and $\covyx$ directly. Instead, we must rely on an i.i.d. sample $(\x_1,\y_1),\ldots,(\x_n,\y_n)$ from $\pp{P}(X,Y)$. Let us define sampling operators $S_\x:\hbspace\rightarrow\mathbb{R}^n$ and $S_\y:\hbspg\rightarrow\mathbb{R}^n$ defined as $S_{\x}f=(f(\x_1),\ldots,f(\x_n))^\top$ and $S_{\y}g=(g(\y_1),\ldots,g(\y_n))^\top$ respectively. It can be shown (see \citealp{Smale-07} for details) that the adjoint of these operators are given by $S^*_\x:\mathbb{R}^n\rightarrow\hbspace$, $\bm{\alpha}\mapsto \sum^n_{i=1}\alpha_i k(\x_i,\cdot)$ and $S^*_\y:\mathbb{R}^n\rightarrow\hbspg$, $\bm{\beta}\mapsto \sum^n_{i=1}\beta_i l(\y_i,\cdot)$, where $\bm{\alpha}:=(\alpha_1,\ldots,\alpha_n)^\top$ and $\bm{\beta}:=(\beta_1,\ldots,\beta_n)^\top$.
Then, the empirical estimator of the conditional mean embedding is given by
\begin{eqnarray}
  \label{eq:cme-empirical}
  \ecovyx (\ecovx+\lambda\mathcal{I})^{-1} k(\x,\cdot) &=& \frac{1}{n}S^*_{\y}S_{\x}\left(\frac{1}{n}S^*_{\x}S_{\x} + \lambda\mathcal{I}\right)^{-1}k(\x,\cdot) \nonumber \\
  &=& S^*_{\y}\left(S_{\x}S^*_{\x} + n\lambda\id_n\right)^{-1}S_{\x}k(\x,\cdot) \nonumber \\
  &=& S^*_{\y}\left(\kmat + n\lambda\id_n\right)^{-1}\mathbf{k}_\x ,
\end{eqnarray}
\noindent where $\mathcal{I}$ denotes the identity operator in $\hbspace$, $\id_n$ is an $n\times n$ identity matrix, $\mathbf{k}_\x = S_{\x} k(\x,\cdot)=(k(\x,\x_1),\ldots,k(\x,\x_n))^\top$ and $S_{\x}S^*_{\x}:\mathbb{R}^n\rightarrow\mathbb{R}^n$ is the Gram matrix associated with $(\x_1,\ldots,\x_n)$ denoted as $\kmat$. The most important step of the derivation uses the identity 
$S_{\x}\left(S^*_{\x}S_{\x} + n\lambda\mathcal{I}\right)^{-1} = \left(S_{\x}S^*_{\x} + n\lambda\id_n\right)^{-1}S_{\x}$. 
Since $\ecovx$ is a compact operator, it has an arbitrary small positive eigenvalue when $\hbspace$ is infinite dimensional.\footnote{If $\inspace$ and $\iny$ are Banach spaces and $\mathbf{T} : \inspace\rightarrow \iny$ is a bounded linear operator, $\mathbf{T}$ is compact if and only if for every bounded sequences $\{x_n\}\subset \inspace$, $\{\mathbf{T}x_n\}$ has a subsequence convergent in $\iny$ \citep[Chapter 6]{ReedSimon81:FA}.} We thus need a regularizer $\lambda\mathcal{I}$ for the inverse of $\ecovx$ to be well-posed. Another possibility is to employ spectral filtering algorithms, \ie, $\muh = S^*_{\y} g_\lambda(\kmat)\mathbf{k}_\x$ where $g_\lambda$ is a filter function, as also suggested by \citet{Muandet2014a}. That is, we can construct a wide class of conditional mean estimators via different regularization strategies.

Theorem \ref{thm:conditional-estimator} gives a formal characterization of the empirical estimator of the conditional mean embedding.

\begin{shadowbox}
\begin{theorem}[\citealp{Song10:KCOND}]
  \label{thm:conditional-estimator}
  The conditional mean embedding $\muv_{Y|\x}$ can be estimated using
  \begin{equation}
    \label{eq:conditional-estimator}
    \muh_{Y|\x} = S^*_{\y}(\kmat + n\lambda\id_n)^{-1}\mathbf{k}_\x .
  \end{equation}
\end{theorem}
\end{shadowbox}
 
Alternately, we may write \eqref{eq:conditional-estimator} as $\muh_{Y|\x} = \sum_{i=1}^n\beta_i l(\y_i,\cdot)$ where $\bm{\beta} := (\kmat + n\lambda\id_n)^{-1}\mathbf{k}_\x \in\rr^n$. That is, it is in the same form as the embedding of the marginal distribution discussed previously, except that the values of the coefficients $\bm{\beta}$ now depend on the value of the conditioning variable $X$ instead of being uniform \citep{Song10:KCOND}. It is important to note that in this case the coefficient $\bm{\beta}$ need not be positive and it does not have to sum to one. In some applications of conditional mean embedding such as state-space models and reinforcement learning, however, one needs to interpret the $\bm{\beta}$ as probabilities, which is almost always not the case for conditional embeddings. In \citet[Theorem 6]{Song10:KCOND}, the rate of convergence is $O_p((n\lambda)^{-1/2}+\lambda^{1/2})$, suggesting that the conditional mean embeddings are harder to estimate than the marginal embeddings, which converge at a rate $O_p(n^{-1/2})$. In \citet{Fukumizu2015:Brief}, under the condition that the eigenvalues $(\gamma_m)_{m=1}^\infty$ of $\covx$ decays as $\gamma_m\leq \beta m^{-b}$ for some $\beta>0$, it is shown that the convergence rate is of $O_p(n^{-b/(4b+1)})$ with the appropriate choice of regularization coefficient.

\section{Regression Interpretation}
\label{sec:regression-view}

As illustrated in Figure \ref{fig:conditional-embedding}, the conditional mean embedding has a natural interpretation as a solution to the vector-valued regression problem. This observation has been made in \citet{Zhang2011} and later thoroughly in \citet{Grunewalder12:LGBPP}, which we review below.

Recall that the conditional mean embedding is defined via $\ep[g(Y)|X=\x] = \langle g,\muh_{Y|\x}\rangle_{\hbspg}$. That is, for every $\x\in\inx$, $\muh_{Y|\x}$ is a function on $\iny$ and thereby defines a mapping from $\inx$ to $\hbspg$. Furthermore, the empirical estimator in \eqref{eq:conditional-estimator} can be expressed as $\muh_{Y|\x} = S^*_{\y}(\kmat + n\lambda\id_n)^{-1}\mathbf{k}_\x$, which already suggests that the conditional mean embedding is the solution to an underlying regression problem. Given a sample $(\x_1,\z_1),\ldots,(\x_n,\z_n) \in \inx\times\hbspg$, a vector-valued regression problem can be formulated as
\begin{equation}
  \label{eq:vv-loss}
  \widehat{\mathcal{E}}_{\lambda}(f) = \sum_{i=1}^n\left\| \z_i - f(\x_i)\right\|^2_{\hbspg} + \lambda\|f\|^2_{\hbspace_{\Gamma}} , 
\end{equation} 
\noindent where $\hbspg$ is a Hilbert space and $\hbspace_{\Gamma}$ denotes a \gls{rkhs} of vector-valued functions from $\inx$ to $\hbspg$ (see \citealp{Micchelli05:Learning} for more details). \citet{Grunewalder12:LGBPP} shows that $\muh_{Y|X}$ can be obtained as a minimizer of the optimization of the form \eqref{eq:vv-loss}.\footnote{In fact, a regression view of conditional mean embedding has already been noted very briefly in \citet[Section 6]{Song10:KCOND} with connections to the solutions of Gaussian process regression \citep{Rasmussen2005} and kernel dependency estimation \citep{Cortes05:KDE}. Nevertheless, \citet{Grunewalder12:LGBPP} gives a more rigorous account of this perspective.}

Following the analysis of \citet{Grunewalder12:LGBPP}, a natural optimization problem for the conditional mean embedding is to find a function $\muv:\inx\rightarrow\hbspg$ that minimizes the following objective:
\begin{equation}
  \label{eq:natural-loss}
  \mathcal{E}[\muv] = \sup_{\|g\|_{\hbspg}\leq 1} \mathbb{E}_X\left[ (\mathbb{E}_Y[g(Y)|X] - \langle g,\muv(X)\rangle_{\hbspg})^2 \right] .
\end{equation}
Unfortunately, we cannot estimate $\mathcal{E}[\muv]$ because we do not observe $\mathbb{E}_Y[g(Y)|X]$. \citet{Grunewalder12:LGBPP} shows that $\mathcal{E}[\muv]$ can be bounded from above by a \emph{surrogate loss function} given by
\begin{equation}
  \label{eq:surrogate-loss}
  \mathcal{E}_s[\muv] = \mathbb{E}_{(X,Y)}\left[\|l(Y,\cdot) - \muv(X)\|^2_{\hbspg}\right] ,
\end{equation} 
\noindent which can then be replaced by its empirical counterpart
\begin{equation}
  \label{eq:emp-surr-loss}
  \widehat{\mathcal{E}}_s[\muv] = \sum_{i=1}^n\|l(\y_i,\cdot) - \muv(\x_i)\|^2_{\hbspg} + \lambda\|\muv\|^2_{\hbspace_{\Gamma}}.
\end{equation}
The regularization term is added to provide a well-posed problem and prevent overfitting.

If $\Gamma$ is a kernel function associated with $\hbspace_{\Gamma}$, it follows from \citet[Theorem 4]{Micchelli05:Learning} that the solution to the above optimization problem can be written as $\muh(\x) = \sum_{i=1}^n\Gamma_{\x_i}(\x)c_i = \sum_{i=1}^n\Gamma(\x_i,\x)c_i$ for some coefficients $\{c_i\}_{i\leq n},c_i\in\hbspg$. Note that $\Gamma$ is an \emph{operator-valued kernel} \citep{Alvarez12:KVF}. \citet{Grunewalder12:LGBPP} considers $\Gamma(\x,\x') = k(\x,\x')\mathrm{Id}$ where $\mathrm{Id}:\hbspg\rightarrow\hbspg$ is the identity map on $\hbspg$. Under this particular choice of kernel, $c_i = \sum_{j\leq n}W_{ij}l(\y_i,\cdot)$ where $\mathbf{W} = (\kmat + \lambda\id)^{-1}$ and $\muh = \sum_{i=1}^n\Gamma_{\x_i}(\kmat+\lambda\id)^{-1}l(\y_j,\cdot)$ which is exactly the embedding in \eqref{eq:conditional-estimator}. It remains an interesting question as to whether one can also employ a more general kernel $\Gamma(\x,\x')$ that is useful in practice.
 
The advantages of vector-valued regression interpretation of conditional mean embedding are two-fold. First, since we have a well-defined loss function, we can use a cross-validation procedure for parameter or model selection, \eg, $\lambda$. Second, it improves the performance analysis of conditional mean embedding as one has access to a rich theory of vector-valued regression \citep{Micchelli05:Learning,Carmeli06:RKHS,Caponnetto07:OptRate,Caponnetto08:Universal}. In particular, by applying  the convergence results of \citet{Caponnetto07:OptRate}, \citet{Grunewalder12:LGBPP} derives minimax convergence rates of the estimator $\muh$ which are $O(\log(n)/n)$. Since the analysis is done under the assumption that $\hbspg$ is finite dimensional, the conditional mean embedding is simply the ridge regression of feature vectors, and thus the better rate---compared to the rate of $O(n^{-1/4})$ of \citet{Song10:KCOND}---is quite natural.

Based on the new interpretation, \citet{Grunewalder12:LGBPP} derives a sparse formulation of the conditional mean embedding $\muh = \sum_{i=1}^n\Gamma_{\x_i}\mathbf{M}l(\y_j,\cdot)$ using a sparse matrix $\mathbf{M}$ obtained from the optimization problem 
$$\min_{\mathbf{M}\in\rr^{n\times n}} \left\|\sum_{i,j=1}^n (\mathbf{M}_{ij} - \mathbf{W}_{ij})k(\x_i,\cdot)l(\y_j,\cdot) \right\|_{k\otimes l} + \gamma\|\mathbf{M}\|_{1,1} , $$ 
\noindent where $\gamma$ is a non-negative regularization parameter and $\|\mathbf{M}\|_{1,1} := \sum_{i,j=1}^n|\mathbf{M}_{ij}|$. This is a Lasso problem with $n^2$ variables. Moreover, one can construct different estimators of conditional mean embedding by introducing a new regularizer in \eqref{eq:emp-surr-loss} (see, \eg, \citealp[Table 1]{Muandet2014a}). It may be of interest to investigate theoretical properties of these new estimators. Lastly, it is instructive to point out that the regression interpretation of the conditional mean embedding can be considered as an instance of a \emph{smooth operator} framework proposed later in \citet{Grunewalder13:SO}.

\section{Basic Operations: Sum, Product, and Bayes' Rules}
\label{sec:basic-operations}

In this section we review basic operations in probabilistic inference and show how they can be carried out in terms of kernel mean embeddings. Sum and product rules are elementary rules of probability. Unlike the traditional recipe, the idea is to perform these operations directly on the marginal and conditional embeddings to obtain a new element in the \gls{rkhs} which corresponds to the embedding of the resulting distribution. One of the advantages of this idea is that the product and sum rules can be performed without making any parametric assumptions on the respective distributions.

Formally, sum and product rules describing the relations between $\pp{P}(X)$, $\pp{P}(Y|X)$, and $\pp{P}(X,Y)$ are given as follows:
\begin{eqnarray}
  \text{Sum rule:} \qquad \pp{P}(X) &=& \sum_Y\pp{P}(X,Y) \,; \label{eq:sum-rule} \\
  \text{Product rule:} \qquad \pp{P}(X,Y) &=& \pp{P}(Y|X)\pp{P}(X)\,. \label{eq:product-rule}
\end{eqnarray}
Combining \eqref{eq:sum-rule} and \eqref{eq:product-rule} yields the renowned Bayes' rule: $\pp{P}(Y|X) = \pp{P}(X|Y)\pp{P}(Y)/\pp{P}(X)$. In the continuous case, the sum in \eqref{eq:sum-rule} turns into an integral. Sum and product rules are fundamental in machine learning and statistics, so much so that nearly all of the probabilistic inference and learning, no matter how complicated they are, amount to repeated application of these two equations. 

Next, we will show how these operations can be achieved as an algebraic manipulation of the (conditional) mean embedding in the \gls{rkhs}. These results are due to \citet{Song10:KCOND,Song2013}, \citet{Fukumizu13:KBR}, and \citet{Schoelkopf15:KPP}.

\subsection{Kernel Sum Rule}

  Using the law of total expectation, which states that for any integrable random variable $X$ and any random variable $Y$, $\ep[X] = \ep_Y[\ep_{X|Y}[X|Y]]$, we have $\muv_X = \ep_{X}[\phi(X)] = \ep_Y[\ep_{X|Y}[\phi(X)|Y]]$. Plugging in the conditional mean embedding yields
  \begin{equation}
    \label{eq:sum-rule-kernel}
    \muv_X = \ep_Y[\mathcal{U}_{X|Y}\varphi(Y)]
    = \mathcal{U}_{X|Y}\ep_Y[\varphi(Y)]
    = \mathcal{U}_{X|Y}\muv_Y .
  \end{equation}
  Theorem \ref{thm:cond-existence} provides sufficient conditions for \eqref{eq:sum-rule-kernel} to be well-defined. Alternatively, we can also use a tensor product $\phi(\x)\otimes\phi(\x)$ to define the kernel sum rule as
\begin{equation}
  \label{eq:sum-rule-tensor}
  \mathcal{C}_{\mathit{XX}} = \mathcal{C}_{(\mathit{XX})|Y}\mu_Y ,
\end{equation}
\noindent where we used a conditional embedding operator $\ep_{X|\y}[\phi(X)\otimes\phi(X)] = \mathcal{C}_{(\mathit{XX})|Y}\phi(\y)$ \citep{Song2013}.

Suppose we have an empirical estimate of $\muv_Y$ and $\mathcal{U}_{X|Y}$, \ie, $\muh_Y = \sum_{i=1}^{m}\alpha_i\varphi(\tilde{\y}_i)$ with a sample $\tilde{\y}_1,\ldots,\tilde{\y}_{m}$ and some coefficients $\bm{\alpha}\in\rr^m$, and $\widehat{\mathcal{U}}_{X|Y} = \ecovxy\ecovy^{-1}$ obtained from an i.i.d. sample $(\x_1,\y_1),\ldots,(\x_n,\y_n)$ drawn from $\pp{P}(X,Y)$.  Note that the samples $(\y_i)$ and $(\tilde{\y}_j)$ used to estimate the covariance operators and the kernel mean $\muh_Y$, respectively, can be different.  Applying \eqref{eq:cme-empirical} to these estimates, the kernel sum rule can be expressed in terms of kernel matrices as
  \begin{equation}
    \label{eq:sum-rule-kernel-emp}
    \muh_X = \widehat{\mathcal{U}}_{X|Y}\muh_Y = \ecovxy\ecovy^{-1}\muh_Y
    = S^*_{\x}(\lmat + n\lambda\id)^{-1}\tilde{\lmat}\bm{\alpha} ,
  \end{equation}
  where $\avec = (\alpha_1,\ldots,\alpha_m)^\top$, $\lmat_{ij} = l(\y_i,\y_j)$, and $\tilde{\lmat}_{ij} = l(\y_i,\tilde{\y}_j)$. Note that we can write $\muh_X = \sum_{j=1}^n\beta_j\phi(\x_j)$ with $\bvec = (\lmat + n\lambda\id)^{-1}\tilde{\lmat}\bm{\alpha}$, which enables us to use it in subsequent operations.

\subsection{Kernel Product Rule}
  
Consider a tensor product of the joint feature map $\phi(X)\otimes\varphi(Y)$. We can then factorize $\muv_{\mathit{XY}} = \ep_{XY}[\phi(X)\otimes\varphi(Y)]$ according to the law of total expectation as
\begin{eqnarray*}
  \ep_Y[\ep_{X|Y}[\phi(X)|Y]\otimes\varphi(Y)] &=& \mathcal{U}_{X|Y}\ep_Y[\varphi(Y)\otimes\varphi(Y)] \\
  \ep_X[\ep_{Y|X}[\varphi(Y)|X]\otimes\phi(X)] &=& \mathcal{U}_{Y|X}\ep_X[\phi(X)\otimes\phi(X)] .
\end{eqnarray*}
Let $\muv_X^{\otimes} := \ep_X[\phi(X)\otimes\phi(X)]$ and $\muv_Y^{\otimes} := \ep_Y[\varphi(Y)\otimes\varphi(Y)]$, \ie, the kernel embeddings using $\phi(\x)\otimes\phi(\x)$ and $\varphi(\y)\otimes\varphi(\y)$ features, respectively. Then, we can write the product rule in terms of kernel mean embeddings as 
\begin{equation}
  \label{eq:product-rule-kernel}
  \muv_{\mathit{XY}} = \mathcal{U}_{X|Y}\muv_Y^{\otimes} = \mathcal{U}_{Y|X}\muv_X^{\otimes} .
\end{equation}
We can see that---unlike the sum rule in \eqref{eq:sum-rule-kernel}---the input to the product rule in \eqref{eq:product-rule-kernel} is a tensor product $\ep_X[\phi(X)\otimes\phi(X)]$. If we identify the tensor product space $\hbspace\otimes\hbspg$ with the Hilbert-Schmidt operators ${\rm HS}(\hbspace,\hbspg)$ as in Section \ref{sec:cross-covariance}, we can rewrite \eqref{eq:product-rule-kernel} as
\begin{equation}
\label{eq:product-rule-op}
\covxy = \mathcal{U}_{X|Y} \covy \quad \text{and} \quad \covyx = \mathcal{U}_{Y|X} \covx.
\end{equation}

Assume that---for some $\avec\in\rr^m$---the embedding $\muh_X^{\otimes}$ and $\muh_Y^{\otimes}$ are given respectively by
\begin{eqnarray*}
  \sum_{i=1}^m\alpha_i\phi(\tilde{\x}_i)\otimes\phi(\tilde{\x}_i) &=& S^*_{\tilde{\x}}\Lambda S_{\tilde{\x}},\\
  \sum_{i=1}^m\alpha_i\varphi(\tilde{\y}_i)\otimes\varphi(\tilde{\y}_i) &=& S^*_{\tilde{\y}}\Lambda S_{\tilde{\y}},
\end{eqnarray*}
\noindent where $\Lambda := \mathrm{diag}(\avec)$ and $S_{\tilde{\x}}$ and $S_{\tilde{\y}}$ are constructed based on $(\tilde{\x}_1,\ldots,\tilde{\x}_m)$ and $(\tilde{\y}_1,\ldots,\tilde{\y}_m)$.
Consequently, we obtain
\begin{equation}
  \label{eq:emp-product-rule}
  \muh_{\mathit{XY}} = \widehat{\mathcal{U}}_{X|Y}\muh_Y^{\otimes} = \ecovxy\ecovy^{-1}\muh_Y^{\otimes}
  = S^*_{\tilde{\x}}(\lmat + n\lambda\id)^{-1}\tilde{\lmat}\Lambda S_{\tilde{\y}}.
\end{equation}
\noindent The operation in terms of $\muh_X^{\otimes}$ has a similar expression, \ie, $\muh_{\mathit{XY}} = \widehat{\mathcal{U}}_{Y|X}\muh_X^{\otimes} = \ecovyx\ecovx^{-1}\muh_X^{\otimes} = S^*_{\y}(\kmat + n\lambda\id)^{-1}\tilde{\kmat}\Lambda S_{\tilde{\x}}$. 
Corresponding to \eqref{eq:product-rule-op}, the formula \eqref{eq:emp-product-rule} can be also expressed as
\begin{equation}
  \label{eq:emp-product-op}
  \ecovxy = \widehat{\mathcal{U}}_{X|Y}\ecovy =  S^*_{\x}(\lmat + n\lambda\id)^{-1}\tilde{\lmat}\Lambda S_{\tilde{\y}}.
\end{equation}
Note that we can also write $\muh_{\mathit{XY}}$ or $\ecovxy$ in terms of the coefficient matrix, \ie, $S^*_{\x}\mathbf{B}S_{\tilde{\y}}$ 
where $\mathbf{B} := (\lmat + n\lambda\id)^{-1}\tilde{\lmat}\Lambda$.


Although both \eqref{eq:sum-rule-kernel} and \eqref{eq:product-rule-kernel} do not require any parametric assumptions on the underlying distributions, these operations can be both statistically difficult and computationally costly in some applications (see below).

\subsection{Kernel Bayes' Rule}

The kernelization of Bayes' rule, called the \emph{kernel Bayes' rule} (KBR)---proposed in \citet{Fukumizu13:KBR}---realizes Bayesian inference in completely nonparametric settings without any parametric models.  The inference is thus not to obtain the posterior of parameters, which is often the purpose of Bayesian inference, but rather a more general posterior of a variable given some observation of another variable.  The likelihood and prior, which are the probabilistic ingredients of Bayes' rule, are expressed in terms of covariance operators and kernel means, respectively. 
Namely, the KBR provides a mathematical method for obtaining an embedding $\muv^{\Pi}_{\mathit{Y|X}}$ of posterior $\pp{P}(Y|X)$ from the embeddings of the prior $\Pi(Y)$ and the likelihood $\pp{P}(X|Y)$. The kernel sum and product rules form the backbone of the kernel Bayes' rule. See \citet{Fukumizu13:KBR} and \citet{Song2013} for technical details.

Essentially, the embedding of the posterior $\pp{P}(Y|X=\x)$ can be obtained as $\muv^{\Pi}_{Y|\x} = \mathcal{C}^{\Pi}_{\mathit{Y|X}}\phi(\x)$ where $\mathcal{C}^{\Pi}_{Y|X}$ denotes the conditional mean embedding which depends on the prior distribution $\Pi(Y)$. Like the classical conditional mean embedding, we may view $\mathcal{C}^{\Pi}_{Y|X}$ as an embedding of the whole posterior $\pp{P}(Y|X)$. The \emph{kernel Bayes' rule} is given by
\begin{equation}
  \label{eq:kernel-bayes-rule}
  \muv^{\Pi}_{Y|\x} = \mathcal{C}^{\Pi}_{\mathit{Y|X}}\phi(\x) = \mathcal{C}^{\Pi}_{\mathit{YX}}(\mathcal{C}^{\Pi}_{\mathit{XX}})^{-1}\phi(\x) ,
\end{equation}
\noindent where the operators $\mathcal{C}^{\Pi}_{\mathit{YX}}$ and $\mathcal{C}^{\Pi}_{\mathit{XX}}$ are given by 
\begin{equation*}
  \mathcal{C}^{\Pi}_{\mathit{YX}} = (\mathcal{C}_{\mathit{X|Y}}\mathcal{C}^{\Pi}_{\mathit{YY}})^\top, \quad \mathcal{C}^{\Pi}_{\mathit{XX}} = \mathcal{C}_{(\mathit{XX})|Y}\mu_Y^{\Pi},
\end{equation*}
\noindent where the former uses the kernel product rule \eqref{eq:product-rule-op}, and the latter uses the kernel sum rule \eqref{eq:sum-rule-tensor}. The embeddings $\mu_Y^{\Pi}$ and $\mathcal{C}^{\Pi}_{\mathit{YY}}$ correspond to the embeddings of $\Pi(Y)$ using features $\varphi(\y)$ and $\varphi(\y)\otimes\varphi(\y)$, respectively.
 
Let $\muh_Y^{\Pi} = \sum_{i=1}^m\alpha_i\varphi(\tilde{\y}_i)$ and $\widehat{C}_{\mathit{YY}}^{\Pi} = \sum_{i=1}^m\alpha_i\varphi(\tilde{\y}_i)\otimes\varphi(\tilde{\y}_i)$ be the empirical estimates obtained from the weighted sample $\{\tilde{\y}_1,\tilde{\y}_2,\ldots,\tilde{\y}_m\}$. Then, the empirical counterpart of \eqref{eq:kernel-bayes-rule} can be written as
\begin{eqnarray*}
  \muh_{Y|\x} &=& \widehat{\mathcal{C}}_{\mathit{YX}}^{\Pi}((\widehat{\mathcal{C}}_{\mathit{XX}}^{\Pi})^2 + \tilde{\lambda}\mathcal{I})^{-1}\widehat{\mathcal{C}}_{\mathit{XX}}^{\Pi}\phi(\x) \\
  &=& S^*_{\tilde{\y}}\Omega^\top((\mathbf{D}\kmat) + \tilde{\lambda}\id)^{-1}\kmat\mathbf{D}\mathbf{k}_{\x} ,
\end{eqnarray*}
\noindent where $\Omega := (\lmat + \lambda\id)^{-1}\tilde{\lmat}\Lambda$ and $\mathbf{D} := \mathrm{diag}((\lmat + \lambda\id)^{-1}\tilde{\lmat}\avec)$. In the above equation, the regularization is different from the standard one used in the conditional kernel mean or kernel sum rule. In essence, we use $(\mathbf{B}^2+\lambda \id)^{-1}\mathbf{B}\z$ to  regularize $\mathbf{B}^{-1}\z$, instead of $(\mathbf{B}+\lambda \id)^{-1}\z$ used in the previous discussions, since in the above case the estimator $\widehat{\mathcal{C}}_{\mathit{XX}}^{\Pi}$ may not be positive definite and thus $\widehat{\mathcal{C}}_{\mathit{XX}}^{\Pi}+\lambda \mathcal{I}$ may not be invertible. Note also that we can write $\muh_{Y|\x} = \sum_{i=1}^m\beta_i\varphi(\tilde{\y}_i)$ where $\bvec := \Omega^\top((\mathbf{D}\kmat) + \tilde{\lambda}\id)^{-1}\kmat\mathbf{D}\mathbf{k}_\x$.

For any function $g\in\hbspg$, we can evaluate the expectation of $g$ w.r.t. the posterior $\pp{P}(Y|\x)$ by means of $\muh_{Y|\x}$ as $\langle g,\muh_{Y|\x}\rangle_{\hbspg}$. If we assume specifically that $g = \sum_{i=1}^n\alpha_i\varphi(\y_i)$ for some $\avec\in\rr^n$, then $\langle g,\muh_{Y|\x}\rangle_{\hbspg} = \bvec^\top\tilde{\lmat}\avec$. Moreover, the parametric form of posterior $\pp{P}(Y|\x)$ can be reconstructed from $\muh_{Y|\x}$ using one of the distributional pre-image techniques discussed previously in \S\ref{sec:dist-preimage}.

\citet{Fukumizu13:KBR} demonstrates consistency of the estimator of the posterior mean $\muh_{Y|\x}$ (and the posterior expectation $\langle g,\muh_{Y|\x} \rangle$), \ie, $\|\muh_{Y|\x} - \muv_{Y|\x}\|_{\hbspg}\rightarrow 0$ in probability as $n\rightarrow\infty$ \citep[Theorem 5]{Fukumizu13:KBR}, and shows that if $\|\muh_Y^{\Pi} - \muv_Y^{\Pi}\|_{\hbspg}=O_p(n^{-\alpha})$ as $n\rightarrow\infty$ for some $0\leq\alpha\leq 1/2$, then---under certain specific assumptions---we have for any $\x\in\inspace$,
\begin{equation*}
  \langle g,\muh_{Y|\x}\rangle_{\hbspg} - \ep[f(Y)|X=\x] = O_p\left(n^{-\frac{8}{27}\alpha}\right)
\end{equation*} 
\noindent as $n\rightarrow\infty$ \citep[Theorem 6]{Fukumizu13:KBR}. In $L^2(\pp{P})$, the rate improves slightly to $O_p(n^{-\frac{1}{3}\alpha})$ \citep[Theorem 7]{Fukumizu13:KBR}. \citet{Fukumizu13:KBR} claims that---while these seem to be slow rates---the convergence can in practice be much faster.

A common application of KBR is in a situation where computing the likelihood of observed data is intractable, while generating samples from the corresponding density is relatively easy. \citet{Kanagawa2016NC}, for example, proposes a filtering method with KBR in the situation where the observation model must be estimated nonlinearly and nonparametrically with data.

As an aside, \citet{Song2013} also provides the formulations of sum, product, and Bayes' rules as a \emph{multi-linear algebraic operation} using a tensor product feature $\phi(\x)\otimes\phi(\x)$---instead of the standard feature $\phi(\x)$.

\subsection{Functional Operations on Mean Embeddings}

  Functional operations---\eg, the multiplication and exponentiation---on random variables are quite common in certain domains such as probabilistic programming.\footnote{In probabilistic programming (PP), instead of representing probabilistic models by graphical models or Bayesian networks, one uses \emph{programs} to represent the models which are more expressive and flexible \citep{Gordon14:PP}.} Given two independent random variables $X$ and $Y$ with values in $\inx$ and $\iny$, and a measurable function $f:\inx\times\iny\rightarrow\inz$, we are interested in estimating the distribution of $Z = f(X,Y)$. Suppose we have the consistent estimates of $\mu[X]$ and $\mu[Y]$, \ie, for some coefficients $\bm{\alpha}\in\rr^m$ and $\bm{\beta}\in\rr^n$,
  \begin{equation*}
    \muh[X] = \sum_{i=1}^m\alpha_i\phi_x(\x_i), \qquad \muh[Y] = \sum_{j=1}^n\beta_j\phi_y(\y_i), 
  \end{equation*}
  \noindent where $\x_1,\ldots,\x_m$ and $\y_1,\ldots,\y_n$ are mutually independent i.i.d. samples from $\pp{P}(X)$ and $\pp{P}(Y)$, respectively. Note that the embedding of conditional random variables can also be written in this form (see the discussion following Theorem \ref{thm:conditional-estimator}). \citet{Schoelkopf15:KPP} proposed to estimate $\mu[f(X,Y)]$ by the following estimator
  \begin{equation}
    \label{eq:kpp-estimate}
    \muh[f(X,Y)] := \frac{1}{\sum_{i=1}^m\alpha_i\sum_{j=1}^n\beta_j}\sum_{i=1}^m\sum_{j=1}^n\alpha_i\beta_j\phi_z(f(\x_i,\y_j)), 
  \end{equation}
  \noindent and showed that \eqref{eq:kpp-estimate} is a consistent estimate of $\muv[f(X,Y)]$ and the convergence happens at a rate $O_p(\sqrt{\sum_i\alpha_i^2} + \sqrt{\sum_j\beta_j^2})$, \ie, see \citet[Theorem 3]{Schoelkopf15:KPP}. Note that the feature maps $\phi_x$, $\phi_y$, and $\phi_z$ may correspond to different kernels. The estimators and theoretical analysis can be extended to functional operations on a larger set of random variables. \citet{Carl-johann16:ConstKME} provides consistency results and finite sample guarantees for the mean embedding of $f(Z)$, where $f$ is a continuous function and $Z$ may consist of multiple random variables. However, if the variables are dependent, a consistent estimator of the joint embedding $\mu_Z$ is also required. Furthermore, their results cover estimators of $\muv[f(Z)]$ that are expressed in terms of ``reduced set'' expansions \citep[Chapter 18]{Scholkopf01:LKS}.

As we can see from \eqref{eq:kpp-estimate}, the gain here is that the embedding of $f(X,Y)$ can be obtained directly from the embedding of $X$ and $Y$ without resorting to density estimation of $f(X,Y)$. Many well-established frameworks for arithmetic operations on independent random variables involve \emph{integral transform} methods such as Fourier and Mellin transforms \citep{Springer79:Algebra} which only allow for some simple cases, \eg, the sum, difference, product, or quotient of independent random variables. Moreover, the functional $f$ is applicable on any domain---no matter how complicated---as long as useful positive definite kernels are introduced on such domains.
 
\begin{remark}
  We provide crucial remarks on the conditional mean embedding.
  \begin{enumerate}[rightmargin=1cm]
    \setlength\itemindent{5pt}
  \item[i)] To guarantee the convergence of the empirical conditional mean embedding to the population one, some old studies use a strong condition such as $k(\x,\cdot)\in \mathcal{R}(\covx)$ or $\ep[g(Y)|X=\cdot]\in\hbspace$ so that $\covyx\covx^{-1}k(\x,\cdot)$ can be well-defined.  Such a condition, however, is often impractically strong: as noted in \citet{Song10:KCOND} and \citet{Fukumizu13:KBR}, Gaussian kernel do not satisfy such an assumption for a wide class of distributions.  If $X$ and $Y$ are independent, for example, $\ep[g(Y)|X=\x]$ is a constant function of $\x$ and then is not included in the RKHS given by a Gaussian kernel \citep[Corollary 4.44]{Steinwart08:SVM}.

  \item[ii)] To alleviate such a strong assumption, a regularized version $\covyx(\covx +\lambda_n\mathcal{I})^{-1}k(\x,\cdot)$ is used instead. For a separable Hilbert space $\hbspace$, it has been shown that $\ecovyx(\ecovx +\lambda_n\mathcal{I})^{-1}k(\x,\cdot)$ converges to $\covyx\covx^{-1}k(\x,\cdot)$ if $\lambda_n$ decays to zero sufficiently slowly as $n\rightarrow\infty$. In practice, a cross-validation procedure is commonly used to find the best $\lambda_n$ for the problems at hand.
  \end{enumerate}
\end{remark}

Below we review some applications that employ the above operations on kernel mean embedding.

\section{Graphical Models and Probabilistic Inference}
\label{sec:graphical-models}

Conditional mean embedding has enjoyed successful applications in graphical models and probabilistic inference  \citep{Song10:KCOND,Song10:HMM,Song11:KBP,Song11:LTGM,Song10:NTGM}. Probabilistic graphical models are ubiquitous in many fields including natural language processing, computational biology, computer vision, and social science. Most of the traditional algorithms for inference often specify explicitly the parametric distributions underlying the observations and then apply basic operations such as sum, product, and Bayes rules on these distributions to obtain the posterior distribution over desired quantities, \eg, parameters of the model. On the other hand, the philosophy behind embedding-based algorithms is to represent distributions by their mean embedding counterparts, and then to apply the operations given in Section \ref{sec:basic-operations} on these embeddings instead. This method leads to several advantages over the classical approach. First, an inference can be performed in a non-parametric fashion; one does not need a parametric assumption about the underlying distribution as well as prior-posterior conjugacy. Second, most algorithms do not require density estimation which is difficult in high-dimensional spaces \citep[Section 6.5]{Wasserman2006}. Lastly, many models are only restricted to deal with discrete latent variables, \eg, a hidden Markov model (HMM) \citep{Baum1966:HMM}. The embedding approach allows for (possibly structured) non-Gaussian continuous variables, which makes these models applicable for a wider class of applications. Nevertheless, there are some disadvantages as well. First, relying on the kernel function, the resulting algorithms are usually sensitive to the choice of kernel and its parameters which need to be chosen carefully. Second, the algorithms only have access to the embedding of the posterior distribution rather than the distribution itself. Hence, to recover certain information such as the shape of the distribution, one needs to resort to a pre-image problem to obtain an estimate of the full posterior distribution (cf. Section \ref{sec:dist-preimage} and \citealp{Song08:TDE,Kanagawa2014,McCalman2013}). Lastly, the algorithms can become computationally costly. Approximation techniques such as low-rank approximation are often used to reduce the computation time and memory storage. See also \citet{Song2013} for a unified view of nonparametric inference in graphical models with conditional mean embedding.

The conditional mean embedding was first introduced in \citet{Song10:KCOND} with applications in dynamical systems. In dynamical systems, one is interested in a joint distribution $\pp{P}(\mathbf{s}_1,\ldots,\mathbf{s}_T,\mathbf{o}_1,\ldots,\mathbf{o}_T)$ where $\mathbf{s}_t$ is the hidden state at time step $t$ and $\mathbf{o}_t$ is the corresponding observation. A common assumption is that a dynamical system follows a \emph{partially observable} Markov model under which the joint distribution factorizes as
\begin{equation*}
  \pp{P}(\mathbf{o}_1,\mathbf{s}_1)\prod_t\pp{P}(\mathbf{o}_t|\mathbf{s}_t)\pp{P}(\mathbf{s}_t|\mathbf{s}_{t-1}) .
\end{equation*}
\noindent Thus, the system is characterized by two important models, namely, a \emph{transition model} $\pp{P}(\mathbf{s}_t|\mathbf{s}_{t-1})$ which describes the evolution of the system and the \emph{observation model} $\pp{P}(\mathbf{o}_t|\mathbf{s}_t)$ which captures the uncertainty of a noisy measurement process. \citet{Song10:KCOND} focuses on \emph{filtering} which aims to query the posterior distribution of the state conditioned on all past observations, \ie, $\pp{P}(\mathbf{s}_{t+1}|\mathbf{h}_{t+1})$ where $\mathbf{h}_t=(\mathbf{o}_1,\ldots,\mathbf{o}_t)$. The distribution $\pp{P}(\mathbf{s}_{t+1}|\mathbf{h}_{t+1})$ can be obtained in two steps. First, we \emph{update} the distribution by
\begin{equation*}
  \pp{P}(\mathbf{s}_{t+1}|\mathbf{h}_t) = \ep_{\mathbf{s}_t|\mathbf{h}_t}[\pp{P}(\mathbf{s}_{t+1}|\mathbf{s}_t)|\mathbf{h}_t].
\end{equation*}
\noindent Then, we \emph{condition} the distribution on a new observation $\mathbf{o}_{t+1}$ using Bayes' rule to obtain
\begin{equation*}
  \pp{P}(\mathbf{s}_{t+1}|\mathbf{h}_t\mathbf{o}_{t+1}) \propto \pp{P}(\mathbf{o}_{t+1}|\mathbf{s}_{t+1})\pp{P}(\mathbf{s}_{t+1}|\mathbf{h}_t).
\end{equation*}
\noindent \citet{Song10:KCOND} propose the exact updates for prediction \citep[Theorem 7]{Song10:KCOND} and conditioning \citep[Theorem 8]{Song10:KCOND} which can be formulated entirely in terms of kernel mean embeddings. Despite the exact updates, one still needs to estimate the conditional cross-covariance operator in each conditioning step, which is both statistically difficult and computationally costly. This problem is alleviated by using approximate inference under some simplifying assumptions. See \citet[Theorem 9]{Song10:KCOND} for technical details. Empirically, although requiring labeled sequence of observations to perform filtering, for strongly nonlinear dynamics it has been shown to outperform standard Kalman filter which requires the exact knowledge of the dynamics. \citet{McCalman2013} also considers the filtering algorithm based on kernel mean embedding, \ie, the kernel Bayes' rule \citep{Fukumizu11:KBR}, to address the multi-modal nature of the posterior distribution in robotics.

One of the advantages of mean embedding approach in graphical models is that it allows us to deal with (possibly structured) non-Gaussian continuous variables. For example, \citet{Song10:HMM} extends the \emph{spectral algorithm} of \citet{HsuKZ09:Spectral} for learning traditional hidden Markov models (HMMs), which are restricted to discrete latent state and discrete observations, to structured and non-Gaussian continuous distributions (see also \citealp{Jaeger00:Discrete} for a formulation of discrete HMMs in terms of \emph{observation operator} $\mathcal{O}_{ij} = \pp{P}(\mathbf{h}_{t+1}=i|\mathbf{h}_t=j)\pp{P}(X_t=\mathbf{x}_t|\mathbf{h}_t=j)$.
In \citet{HsuKZ09:Spectral}, HMM is learned by performing a singular value decomposition (SVD) on a matrix of joint probabilities of past and future observations. \citet{Song10:HMM} relies on the embeddings of the distributions over observations and latent states, and then constructs an operator that represents the joint probabilities in the feature space. The advantage of the spectral algorithm for learning HMMs is that there is no need to perform a local search when finding the distribution of observation sequences, which usually leads to more computationally efficient algorithms. Unlike \citet{Song10:KCOND}, the algorithm only requires access to an unlabeled sequence of observations.

A nonparametric representation of tree-structured graphical models was introduced in \citet{Song10:NTGM}. Inference in this kind of graphical model relies mostly on message passing algorithms. In case of a discrete variable, or Gaussian distribution, the message passing can be carried out efficiently using the sum-product algorithm. \citet{Minka2001:EP} proposes the expectation-propagation (EP) algorithm which requires an estimation of only certain moments of the messages. \citet{Sudderth10:CACM} considers messages as mixture of Gaussians. The drawback of this method is that the number of mixture components grows exponentially as the message is propagated. \citet{Ihler2009:PBP} considers a particle belief propagation (BP) where the messages are expressed as a function of a distribution of particles at each node. Unlike these algorithms, the embedding-based algorithm proposed in \citet{Song10:NTGM} expresses the message $\mathbf{m}_{\mathit{ts}}(s)$ between pairs of nodes as \gls{rkhs} functions on which sum and product steps can be performed using the linear operation in the \gls{rkhs} to obtain a new message. In addition, \citet{Song10:NTGM} also proves the consistency of the conditional mean embedding estimator, \ie, $\|\hat{\mathcal{U}}_{Y|X}-\mathcal{U}_{Y|X}\|_{\text{HS}}$ converges in probability under some reasonable assumptions \citep[Theorem 1]{Song10:NTGM}. The algorithm was applied in cross-lingual document retrieval and camera orientation recovery from images. The idea has been used later for latent tree graphical models \citep{Song11:LTGM}, which are often used for expressing hierarchical dependencies among many variables in computer vision and natural language processing, and for belief propagation algorithm \citep{Pearl1988:PRI,Song11:KBP} for pairwise Markov random fields. 

Assuming that the latent structure underlying the data-generating process has a low-rank structure, \eg, latent tree, \citet{Song13:LowRank} constructs an improved estimator of the kernel mean embedding for multivariate distribution using the truncated SVD (TSVD) algorithm.
 
\section{Markov Chain Monte Carlo Methods}  

Attempts have been made in improving Markov Chain Monte Carlo (MCMC) methods using the kernel mean embedding. MCMC is a classical technique for estimating the expectations in approximate Bayesian inference \citep{Robert05:MonteCarlo, Neal93:MCMC}. Given a probability distribution $\pi$ on a set $\inspace$, the problem is to generate random elements of $\inspace$ with distribution $\pi$. For example, $\pi$ might be a posterior distribution $\pp{P}(\bm{\theta}\,|\,\{\y_i\}_{i=1}^n)$ over some parameter vector $\bm{\theta}$ given the observations $\{\y_i\}_{i=1}^n$. Since $\pi$ is often intractable, we cannot sample from $\pi$ directly. Instead, MCMC methods generate random samples by constructing a \emph{Markov chain} that has as its equilibrium distribution the target distribution $\pi$. Designing general purpose transition probabilities that can have any distribution of interest as their stationary distribution is the key to MCMC methods. Metropolis-Hastings, Gibbs, and Hamiltonian Monte Carlo (HMC) are well-known examples of algorithms for generating a Markov chain. The interested reader is referred to \citet{Robert05:MonteCarlo} for further details on MCMC methods.


Adaptive MCMC is a class of MCMC methods that try to learn an optimal proposal distribution from previously accepted samples \citep{Haario01:AdapMCMC,Andrieu08:TutorialAMCMC}. \citet{Sejdinovic2014} introduces a kernel adaptive Metropolis-Hasting (KAMH) algorithm for sampling from a target distribution with strongly non-linear support. The idea is similar to that proposed in \citet{Haario01:AdapMCMC}, but when the target distributions are strongly non-linear, the samplers of \citet{Haario01:AdapMCMC} may suffer from low acceptance probability and slow mixing. The idea of \citet{Sejdinovic2014} is to embed the trajectory of the Markov chain into the RKHS and then to construct the proposal using the information encoded in the empirical covariance operator. Intuitively, the empirical covariance operator contains the information about the non-linear support of the distribution. Hence, kernel PCA \citep{Scholkopf98:KPCA} direction may be used to construct the proposal distribution. The proposed algorithm first obtains the RKHS sample 
\begin{equation*}
  f = k(\y,\cdot) + \sum_{i=1}^n\beta_i[k(\z_i,\cdot) - \mu_{\z}]
\end{equation*}
\noindent from the Gaussian measure in the RKHS and then finds a point $\x^*\in\inspace$ whose canonical feature map $k(\x^*,\cdot)$ is close to $f$ in an RKHS norm. Consequently, the resulting sampler is adaptive to the local structure of the target at the current chain state, is oriented towards a nearby region of high density, and does not suffer from a wrongly scaled proposal distribution. \citet{Strathmann15:HMC} introduces a kernel Hamiltonian Monte Carlo (KHMC) to address the case where one cannot run a traditional Hamiltonian Monte Carlo (HMC) since the target is \emph{doubly intractable}, \ie, gradients of the energy functions are not available. They use an infinite dimensional exponential family model in an RKHS as a surrogate model and learn it via score matching \citep{Hyvaerinen05:ScoreMatching,Sriperumbudur13:DenExp}. Recently, \citet{Schuster16:KSMC} looks into the ways of using RKHS representation for sequential Monte Carlo (SMC), which is more convenient than the MCMC methods as it does not require \emph{vanishing adaptation} to ensure convergence to the correct invariant distribution.
 
\section{Markov Decision Processes and Reinforcement Learning}
\label{sec:mdp-reinforcement}

\begin{figure}[t!]
  \centering
  \begin{tikzpicture}[node distance=2.5cm,auto,>=latex']
    
    \node (A0) at (-2,2) {};      
    \node (S0) at (-2,0) {};
    \node (SS) at (8,0) {};
    \begin{scope}[every node/.style={circle,ultra thick,draw}]     

      \node (A1) at (0,2) {$\mathbf{a}_1$};      
      \node (S1) at (0,0) {$\mathbf{s}_1$};
      \node (R1) at (0,-2) {$\mathbf{r}_1$};

      \node (A2) at (2,2) {$\mathbf{a}_2$};      
      \node (S2) at (2,0) {$\mathbf{s}_2$};
      \node (R2) at (2,-2) {$\mathbf{r}_2$};

      \node (AT) at (6,2) {$\mathbf{a}_T$};      
      \node (ST) at (6,0) {$\mathbf{s}_T$};
      \node (RT) at (6,-2) {$\mathbf{r}_T$};
    \end{scope}
    
    \node (DD) at (4,0) {{\Large $\cdots$}};
    \node (DD2) at (4,2) {{\Large $\cdots$}};
    \node (DD3) at (4,-2) {{\Large $\cdots$}};
    \path [->,very thick] (S0) edge (S1);
    \path [->,very thick] (S1) edge (S2);
    \path [->,very thick] (S2) edge (DD);
    \path [->,very thick] (DD) edge (ST);
    \path [->,very thick] (ST) edge (SS);

    \path [->,very thick] (A0) edge (S1);

    \path [->,very thick] (A1) edge (S2);
    \path [->,very thick] (S1) edge (R1);
    \path [->,very thick] (A1) edge[out=-50,in=50] (R1);

    \path [->,very thick] (A2) edge (DD);
    \path [->,very thick] (S2) edge (R2);
    \path [->,very thick] (A2) edge[out=-50,in=50] (R2);

    \path[->,very thick] (DD2) edge (ST);

    \path [->,very thick] (AT) edge (SS);
    \path [->,very thick] (ST) edge (RT);
    \path [->,very thick] (AT) edge[out=-50,in=50] (RT);
\end{tikzpicture}
  \caption{A graphical illustration of a Markov decision process. The state is dependent on the previous state and action, and the reward depends on the current state and action.}
  \label{fig:dynamical-systems}
\end{figure}
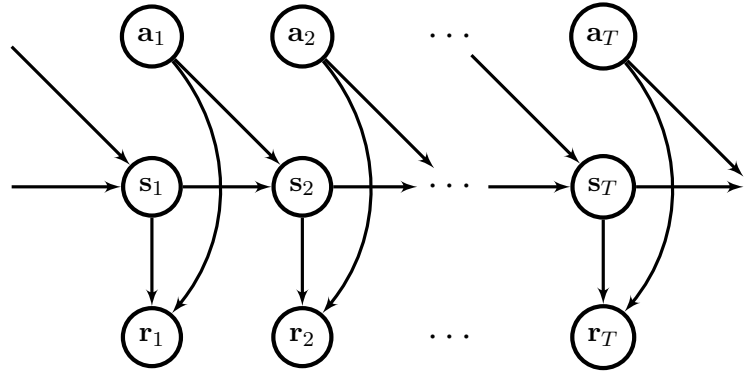

A Markov decision process (MDP) is a discrete time stochastic control which is widely studied and applied in robotics, automated control, economics, and manufacturing \citep{Bellman57:MDP}. See Figure \ref{fig:dynamical-systems} for a graphical illustration of a MDP. Formally, an MDP is a 5-tuple $(\mathcal{S},\mathcal{A},P(\cdot,\cdot),R(\cdot,\cdot),\gamma)$ where $\mathcal{S}$ is a finite set of states, $\mathcal{A}$ is a finite set of actions, $P_{\mathbf{a}}(\mathbf{s},\mathbf{s}')$ is a \emph{state-transition probability} $P(\mathbf{s}_{t+1}=\mathbf{s}' \,|\, \mathbf{s}_t=\mathbf{s},\mathbf{a}_t=\mathbf{a})$, $R_{\mathbf{a}}(\mathbf{s},\mathbf{s}')$ is the immediate reward received after the transition from state $\mathbf{s}$ to state $\mathbf{s}'$, and $\gamma\in[0,1]$ is the discount factor. At each time step, the process is in some state $\mathbf{s}$, and we may choose any action $\mathbf{a}$ that is available in state $\mathbf{s}$. The process moves into the new state $\mathbf{s}'$ according to the state-transition function $P_{\mathbf{a}}(\mathbf{s},\mathbf{s}')$ and gives a corresponding reward $R_{\mathbf{a}}(\mathbf{s},\mathbf{s}')$. The discount factor $\gamma$ represents the difference in importance between future rewards and present rewards. The goal of MDP is to obtain an optimal policy that maximizes a value function $V(\cdot)$ defined on beliefs (distributions over states), and determined by the reward function $R$, see, \eg, \citet{Sutton98:IRL} for further details.

As in a MDP, any problems that involve \emph{control optimal theory} can be solved by analyzing the appropriate Bellman equation. According to Bellman's \emph{principle of optimality}, an optimal policy has the property that whatever the intial state and initial decision are, the remaining decisions must constitute an optimal policy with regard to the state resulting from the first decision \citep{Bellman03:DP}. In MDP, this principle is reflected in the \emph{Bellman optimality equation}:
\begin{equation}
  \label{eq:bellman-operator}
  (\mathcal{B}V)(\x) := \max_{\mathbf{a}\in \mathcal{A}}\,\left\{ r(\x,\mathbf{a}) + \gamma\ep_{X\sim P(\cdot |\mathbf{x},\mathbf{a})}[V(X)]\right\} .
\end{equation}
\noindent The operator $\mathcal{B}$ is known as the \emph{Bellman operator}. If the image of $\mathcal{B}$ is always a measurable function, $V_{t+1} = \mathcal{B}V_t$ converges in sup-norm to the optimal solution.

The Bellman equation is in general difficult to solve. Traditional approaches for solving the Bellman equation such as dynamic programming \citep{Bellman03:DP}, parametric methods \citep{Engel03:GPBellman}, Monte Carlo methods \citep{Sutton98:IRL}, and piecewise linear and convex (PWLC)-based methods \citep{Smallwood73:POMDP} have several drawbacks. Dynamic programming has well-understood mathematical properties, but requires a complete and accurate model of the environment. Monte Carlo simulation is conceptually simple and requires no model, but can result in high-variance estimates. To overcome some of these difficulties, \citet{Grunewalder12:MDPs} proposes to learn transition dynamics, \ie, value iteration and optimal policy learning, in MDPs by representing the stochastic transitions as conditional mean embeddings. That is, they are interested in the embedding of the expectation operator corresponding to the state transition probability kernel $\pp{P}$ such that $\langle \mu_{(\mathbf{x},\mathbf{a})},f\rangle_{\hbspace} = \ep[f(X_{t+1})\,|\, X_t=\mathbf{x},A_t=\mathbf{a}]$. Specifically, the (empirical) Bellman operators $\mathcal{B}$ are first estimated using the kernel mean embedding:
\begin{equation}
  \label{eq:bellman-rkhs}
  (\mathcal{B}V)(\avec) := \max_{\mathbf{a}\in A}\left\{ \avec^\top\mathbf{R}_{\mathbf{a}} + \gamma\bvec^\top\mathbf{V}(\cdot)\right\} ,
\end{equation}
\noindent where $\mathbf{R}_{\mathbf{a}} \in\rr^n$ is the reward vector on a sample for action $\mathbf{a}$ and $\mathbf{V}(\cdot)$ denotes the vector of value functions defined in terms of RKHS quantities (see \citealp{Grunewalder12:MDPs,Nishiyama12:POMDPs} for precise definitions). Then, these estimated operators are used in the standard approach for solving MDPs. While the ordinary Bellman operator \eqref{eq:bellman-operator} is \emph{contractive} and \emph{isotonic} which guarantee that the value iteration converges monotonically \citep{Porta06:PVI}, the kernel Bellman operator in \eqref{eq:bellman-rkhs} may not have these two properties due to the unnormalized coefficients of the conditional mean embedding (see also Theorem \ref{thm:conditional-estimator} and the following discussion). To alleviate this problem, \citet{Grunewalder12:MDPs} instead considers the normalized version of the conditional mean estimate and shows that a certain contraction mapping for $\mathcal{B}$ can be achieved.

\citet{Nishiyama12:POMDPs} extends the kernel-based algorithm of \citet{Grunewalder12:MDPs} to a partially observable MDP (POMDP). In POMDP, we do not assume that the state $\mathbf{s}$ is known when action $\mathbf{a}$ is to be taken. This approach also considers the distributions over states, observations, and actions, and employs the kernel Bayes' rule \citep{Fukumizu13:KBR} to update the embeddings. The major contribution of \citet{Nishiyama12:POMDPs} is to formulate POMDPs in feature space (using the kernel Bellman equation) and to propose a kernelized value iteration algorithm. However, the algorithm requires the labeled training data for the true latent states.

\citet{Boots2013} proposes to model \emph{predictive state representations} (PSRs) using conditional mean embeddings. The PSR represents a state as a set of predictions of future observable events which differ fundamentally from the latent variable model as it depends only on the observable quantities. For that reason, learning PSRs should be easier. The key idea of this work is to represent the state as nonparametric conditional mean embeddings in an RKHS. The benefit of mean embeddings here is that it allows for learning algorithms which work well for \emph{continuous} actions and observations, unlike traditional algorithms which often run into trouble due to lack of data. Moreover, previous approaches often use kernel density estimation, which suffer from slow convergence and difficult numerical integration. The algorithm also updates the states using the kernel Bayes' rule \citep{Fukumizu13:KBR}. Lastly, it is claimed that although all prediction functions in the proposed algorithm are linear PSRs---\ie, there is a linear relationship between conditional probabilities of tests---it can still represent non-linear dynamics \citep{Boots2013}.

Finally, it should be emphasized that the traditional approach for MDP and POMDP involves a conditional density estimation which is difficult for high dimensional problems. In addition, computation of high dimensional integrals to obtain expectation can be very costly. On the other hand, the kernel mean embedding turns the expectation operator into an RKHS inner product, which has linear complexity in the number of training points. Moreover, the kernel mean estimates do not scale poorly with the dimension of the underlying space.

\section{Conditional Dependency Measures}
\label{sec:cond-dependency} 

The task of determining the conditional dependency is ubiquitous in Bayesian network learning, gene expression, and causal discovery. Like an unconditional case, a joint distribution $\pp{P}(X,Y,Z)$ over variables $X$, $Y$, and $Z$ satisfies a conditional independence relationship $X\ci Y | Z$ if $\pp{P}(X,Y,Z)$ factorizes as $\pp{P}(X|Z)\pp{P}(Y|Z)\pp{P}(Z)$. Several other equivalent characterizations of conditional independence are given in \citet{Dawid79:Cond}, for example. Testing for conditional independence is generally a challenging problem due to the ``curse of dimensionality'' in terms of the dimensionality of the conditioning variable $Z$ \citep{Bergsma04:Cond}.

There exist numerous tests for conditional dependency such as \emph{partial correlation} tests \citep{Baba04:Partial}, conditional densities based tests \citep{Su08:Hellinger}, and permutation-based tests \citep{Tsamardinos10:Permute}. These classical tests require either the parametric assumption of the underlying distribution, \eg, Gaussianity, or a linear relationship between random variables, or the discretization of the conditioning variable $Z$ or all. Thus, they still suffer from the curse of dimensionality which makes their application domains quite limited. In what follows, we will focus on the kernel-based conditional dependency tests that have been proposed for non-linear and non-Gaussian data to remedy some of the aforementioned issues.

Similar to the idea of HSIC described in \S\ref{sec:dependency}, \citet{Fukumizu2008} proposes a nonparametric conditional dependence measure based on the \emph{normalized conditional cross-covariance operator}
\begin{equation}
  \label{eq:normalized-CCCO}
  \mathcal{V}_{\mathit{YX}|Z} := \mathcal{V}_{\mathit{YX}} - \mathcal{V}_{\mathit{YZ}}\mathcal{V}_{\mathit{ZX}} , 
\end{equation}
\noindent where $\mathcal{V}_{\mathit{YX}}$ is a unique bounded operator such that $\mathcal{C}_{YX} = \mathcal{C}_{YY}^{1/2}\mathcal{V}_{YX}\mathcal{C}_{XX}^{1/2}$ \citep{Baker1973}.\footnote{See also Theorem \ref{thm:baker-thm} for a brief discussion on the operator $\mathcal{V}_{\mathit{YX}}$.} The operators $\mathcal{V}_{YZ}$ and $\mathcal{V}_{ZX}$ can be defined in a similar way. Substituting $\mathcal{V}_{YX} = \mathcal{C}_{YY}^{-1/2}\mathcal{V}_{XY}\mathcal{C}_{XX}^{-1/2}$ (as well as $\mathcal{V}_{YZ}$ and $\mathcal{V}_{ZX}$) in \eqref{eq:normalized-CCCO} yields
\begin{equation}
  \mathcal{V}_{YX|Z} = \mathcal{C}_{YY}^{-1/2}(\underbrace{\mathcal{C}_{YX} - \mathcal{C}_{YZ}\mathcal{C}_{ZZ}^{-1}\mathcal{C}_{ZX}}_{\mathcal{C}_{\mathit{YX}|Z}})\mathcal{C}^{-1/2}_{XX} .
\end{equation}
\noindent Intuitively, the operator $\mathcal{C}_{\mathit{YX}|Z}$ can be viewed as a non-linear operator reminiscent of the \emph{conditional covariane matrix} $\mathcal{C}_{\mathit{YX}|Z}= \mathcal{C}_{\mathit{YX}} - \mathcal{C}_{\mathit{YZ}}\mathcal{C}_{\mathit{ZZ}}^{-1}\mathcal{C}_{\mathit{ZX}}$ of Gaussian random variables. Given extended variables $\ddot{X} = (X,Z)$ and $\ddot{Y} = (Y,Z)$, they show that $\mathcal{C}_{\mathit{\ddot{Y}\ddot{X}}|Z}=\mathbf{0}$ if and only if $X$ and $Y$ are conditionally independent given $Z$, and propose the squared Hilbert-Schmidt norm $\|\mathcal{V}_{\mathit{\ddot{Y}\ddot{X}|Z}}\|^2_{\text{HS}}$ as a measure of conditional dependency and $\|\mathcal{V}_{\mathit{YX}}\|^2_{\text{HS}}$ as an unconditional counterpart. Note that $\mathcal{V}_{\mathit{\ddot{Y}\ddot{X}|Z}} = \mathbf{0}$ and $\mathcal{V}_{\mathit{YX}} = \mathbf{0}$ imply $\mathcal{C}_{\mathit{YX}}=\mathbf{0}$ and $\mathcal{C}_{\mathit{\ddot{Y}\ddot{X}}|Z}=\mathbf{0}$, respectively. Although mathematically rigorous, it is not known how to analytically compute the null distribution of the test statistics. In \citet{Fukumizu2008}, they resort to a bootstrapping approach which is computationally expensive.  See \citet[Section 2.2 and 2.3]{Fukumizu2008} for a rigorous treatment of this measure.

\citet{Zhang2011} resorts to the characterization based on the concept of \emph{partial association} of \citet{Daudin80:Partial} which says that $X\ci Y | Z$ if and only if $\ep[fg]=0$ for all suitable $f$ chosen to be a function from $Z$ to $X$ and for all suitable $g$ chosen to be a function from $Z$ to $Y$ (see, \eg, \citealp[Lemma 2]{Zhang2011}  for details). This characterization is reminiscent of the partial correlation based characterization of conditional independence for Gaussian variables. Most importantly, \citet{Zhang2011} further derives the asymptotic distribution of the test under the null hypothesis, and provides ways to estimate such a distribution. In a similar vein, \citet{Flaxman15:GPI} adopts Gaussian process regression in conditional independence test on non-i.i.d. observations. First, the residual variables $\varepsilon_X,\varepsilon_Y,\varepsilon_Z$ are obtained by regressing from the latent variable $T$, \eg, time or spatial location, to variables $X$, $Y$, and $Z$ to remove their dependency on $T$. In the second step, the residuals $\varepsilon_{\mathit{XZ}}$ and $\varepsilon_{\mathit{YZ}}$ are obtained by regressing from $Z$ to $X$ and $Y$, respectively. Finally, the conditional independence $X\ci Y | Z$ can be cast as an unconditional test between $\varepsilon_{\mathit{XZ}}$ and $\varepsilon_{\mathit{YZ}}$.

Recall that we can view the unconditional independence testing as a two-sample testing between $\pp{P}(X,Y)$ and $\pp{P}(X)\pp{P}(Y)$, see, \eg, \eqref{eq:dependence-mmd}. Likewise for the conditional independence, $X\ci Y | Z$ holds if and only if $\pp{P}(X,Y,Z)=\pp{P}(X|Z)\pp{P}(Y|Z)\pp{P}(Z)$. Hence, we can in principle turn any unconditional independence tests into conditional ones. \citet{Doran2013} extends this view to an unconditional independence test using kernel mean embeddings. That is, given the i.i.d. sample $\{(\x_i,\y_i,\z_i)\}_{i=1}^n$ from $\pp{P}(X,Y,Z)$, the sample from the corresponding $\pp{P}(X|Z)\pp{P}(Y|Z)\pp{P}(Z)$ under the null hypothesis can be obtained approximately as $\{(\x_i,\y_{\pi(i)},\z_i\}_{i=1}^n$ where $\pi(\cdot)$ denotes a random permutation. Unlike in the unconditional case, \citet{Doran2013} proposes to learn $\pi(\cdot)$ from the data in such a way that if $\pi(i) =j$, then $\z_i\approx \z_j$. To get the intuition, when $Z$ is a discrete random variable, \ie, $z\in\{1,2,3,\ldots,m\}$, we have that $X\ci Y | Z$ if and only if $X\ci Y | Z=i$ for all $i$. The latter can be achieved by performing an unconditional independence test in each bin using only $\{\x_k,\y_k\}_{k=1}^{n_i}$ whose $z_k$ are the same.

\section{Causal Discovery}
\label{sec:causal}

Causal inference from observational data concerns how the distribution of effect variable would have changed if we were to intervene on one of the cause variables while keeping all others fixed \citep{Pearl2000}. There has recently been a great deal of research and interest on causality in the machine learning community \citep{Scholkopf12:Causal,Kaggle13,Mooij14,Codalab14}. The kernel mean embedding has proven to be a successful  mathematical tool in this research area.

Given a set of random variables $X=(X_1,X_2,\ldots,X_d)$, the goal of causal discovery is to uncover the underlying causal \emph{directed acyclic graph} (DAG) denoted by $G(V,E)$. Each vertex $V_i\in V$ corresponds to the random variable $X_i$ and an edge $E_{ij}$ from $V_i$ to $V_j$ indicates a direct causal relationship from $X_i$ to $X_j$ (denoted by $X_i\rightarrow X_j$). The causal relationships in the DAG are usually parametrized by a \emph{structural equation model} (SEM)
\begin{equation*}
  X_i\leftarrow f_i(\text{Pa}(X_i),E_i) ,
\end{equation*}
\noindent where $f_i$ is an arbitrary function and $\text{Pa}(X_i)$ is a parental set of $V_i$. It is often assumed that $E_i$ are mutually independent, \ie, there is no hidden confounder.

Under Markov and faithfulness assumptions \citep{Pearl2000}, the PC algorithm \citep{Spirtes00} exploits conditional dependencies to recover the Markov equivalence class of $G$. Owing to this approach, several modern kernel-based conditional independence tests have been developed with this particular application in mind \citep{Fukumizu2008,Zhang2011,Doran2014}. Nevertheless, this approach is not suitable for the bivariate case in which we observe only two random variables $X$ and $Y$. In this case, \citet{Schoelkopf15:KPP} recently proposed a \emph{kernel probabilistic programming} (KPP) which provides an expression of a functional of random variables, \eg, $f(X,Y) = X\times Y$, by means of kernel mean embedding. In particular, they apply it to causal inference by embedding the SEM associated with the additive noise model (ANM) and constructing a test based on such embeddings \citep[Theorem 4]{Schoelkopf15:KPP}. For a comprehensive review of research along this direction, see, \eg, \citet{Mooij14}.

Inspired by the competition organized by \citet{Kaggle13,Codalab14}, \citet{Lopez-Paz15:Towards} proposes a ``data-driven'' method for bivariate causal inference using kernel mean embedding. In contrast to previous studies, this work assumes access to a large collection of cause-effect samples $\mathcal{S} = \{(S_i,l_i)\}_{i=1}^n$ where $S_i = \{(\x_{ij},\y_{ij})\}_{j=1}^{n_i}$ is drawn from $\pp{P}(X_i,Y_i)$ and $l_i$ is a label indicating the causal direction between $X_i$ and $Y_i$. \citet{Lopez-Paz15:Towards} avoids the handcrafted features of $S_i$ by resorting to the kernel mean representation $\hat{\mu}[\hat{\pp{P}}(X_i,Y_i)]$ where $\hat{\pp{P}}(X_i,Y_i)$ is the empirical distribution of $S_i$. Then, inferring causal direction proceeds as a classification problem on distributions \citep{Muandet12:SMM}.

A major challenge of most causal inference algorithms is the presence of hidden confounders. \citet{Sgouritsa2013} proposes the method to detect finite confounders based on the kernel mean embedding of the joint distribution $\pp{P}(X_1,X_2,\ldots,X_d)$. Specifically, they show that if the the joint distribution decomposes as $\pp{P}(X_1,X_2,\ldots,X_d) = \sum_{i=1}^m \pp{P}(Z_i)\prod_{j=1}^d\pp{P}(X_j|Z_i)$, then the tensor rank of the joint embedding
\begin{equation*}
  \mathcal{U}_{X_{[d]}} = \sum_{i=1}^m\pp{P}(Z_i)\otimes_{j=1}^d\mathbb{E}_{X_j}[\phi_j(X_j)|Z_i]
\end{equation*}
\noindent is exactly $m$ \citep[Theorem 1]{Sgouritsa2013}. This allows them to use clustering to identify confounders under various scenarios.

Recently, \citet{Scholkopf12:Causal} postulates that under a specific causal assumption there is an asymmetry in the decomposition
\begin{equation*}
  \pp{P}(X,Y) = \pp{P}(Y|X)\pp{P}(X) = \pp{P}(X|Y)\pp{P}(Y) .
\end{equation*}
\noindent In other words, if $X\rightarrow Y$, it is believed that $\pp{P}(Y|X)$ and $\pp{P}(X)$ are ``independent'', whereas in an anti-causal direction $\pp{P}(X|Y)$ and $\pp{P}(Y)$ may contain information about each other. Owing to this postulate, \citet{Chen2014} defines the uncorrelatedness criterion between $\pp{P}(X)$ and $\pp{P}(Y|X)$ and formulates it in terms of the complexity metric using the Hilbert space embedding of $\pp{P}(X)$ and $\pp{P}(Y|X)$.


\chapter{Relationships between KME and Other Methods}
\label{sec:connections}

In this section, we discuss the relationships between kernel mean embedding and other approaches ranging from kernel density estimation (KDE) to probabilistic models for Bayesian inference. 

\hiddensection{Beyond Density Estimation and Characteristic Function}

Kernel density estimation has long been a popular method of choice for approximating the density function of the underlying distribution \citep{Silverman86:KDE}. Let $\x_1,\ldots,\x_n\in\rr^d$ be a random sample from a distribution $\pp{P}$ with a density $f_p$. The kernel density estimator of this density is
\begin{equation} 
  \hat{f}_p(\x) = \frac{1}{n}\sum_{i=1}^n\kappa_{\sigma}(\x_i-\x) ,
\end{equation}
\noindent where $\kappa_{\sigma}$ satisfies $\int_{\inspace} \kappa_{\sigma}(\x)\dd\x=1$ and $\kappa_{\sigma}(\x)\geq 0$ with bandwidth $\sigma$. Well-known examples of kernels satisfying all of the above properties are the Gaussian kernel, the multivariate Student kernel, and the Laplacian kernel. \citet{Anderson94:TwoSample} constructs a two-sample test statistic using the $L_2$ distance between the kernel density estimators, \ie, $\|\hat{f}_p - \hat{f}_q\|_{L_2(\mathbb{R}^d)}$ where $\hat{f}_p(\x)$ and $\hat{f}_q(\x)$ are the kernel density estimators of densities $p$ and $q$, respectively. More generally, we may define $D_r(p,q) = \|f_p-f_q\|_{L_r(\mathbb{R}^d)}$ as a distance between $p$ and $q$, which subsumes seveal well-known distance measures such as L\'{e}vy distance ($r=1$) and the Renyi entropy based measure ($r=2$). 

As shown in \citet{Gretton12:KTT}, the $L_2$ distance between kernel density estimates $\hat{f}_p$ and $\hat{f}_q$ is a special case of the biased MMD in \eqref{eq:empirical-mmd} between $\muh_{p}$ and $\muh_{q}$. That is, let $\hat{f}_p$ and $\hat{f}_q$ be given by $\hat{f}_p(\x)=\frac{1}{n}\sum_{i=1}^n\kappa_{\sigma}(\x_i-\x)$ and $\hat{f}_q(\x)=\frac{1}{m}\sum_{i=1}^m\kappa_{\sigma}(\y_i-\x)$, respectively. Then, the $L_2$ distance between $\hat{f}_p$ and $\hat{f}_q$ can be written as
\begin{align*}
  \|\hat{f}_p - \hat{f}_q\|^2_{L_2(\mathbb{R}^d)} &= \int_{\mathbb{R}^d}\left(\frac{1}{n}\sum_{i=1}^n\kappa_{\sigma}(\x_i-\z) - \frac{1}{m}\sum_{i=1}^m\kappa_{\sigma}(\y_i-\z)\right)\dd \z \\
  &=\frac{1}{n^2}\sum_{i,j=1}^n k(\x_i,\x_j) + \frac{1}{m^2}\sum_{i,j=1}^m k(\y_i,\y_j)  \\
  & \qquad- \frac{2}{nm}\sum_{i=1}^n\sum_{j=1}^m k(\x_i,\y_j) \\
  &= \left\| \frac{1}{n}\sum_{i=1}^nk(\x_i,\cdot) -\frac{1}{m}\sum_{i=1}^mk(\y_i,\cdot)\right\|^2_{\hbspace_k} \\
  &= \|\muh_p - \muh_q\|^2_{\hbspace_k} , 
\end{align*}
\noindent where $k(\x,\y) = \int \kappa_{\sigma}(\x-\z)\kappa_{\sigma}(\y-\z)\dd\z$. By definition, $k(\x,\y)$ is positive definite, and thereby is a reproducing kernel. As mentioned earlier in the survey, an RKHS-based approach provides advantages over the $L_2$ statistic in a number of important respects. Most notably, it bypasses the density estimation problem which is often considered more difficult especially in high dimensions. A similar approach used in estimating divergences using the kernel density estimate (\eg, $L_2$ divergence, R\' enyi, Kullback-Leibler, or many other divergences) avoids the need to estimate the density by adopting a direct divergence approximation using, \eg, density ratio estimation and regression \citep{Nguyen08:DivFunc,Poczos11:Divergence}.

Another popular approach for representing a probability distribution is through the \emph{empirical characteristic function (ECF)} \citep{Feuerverger1977:ECF,Fernandez08:Charac}. There is a one-to-one correspondence between characteristic functions and distributions and \citet{Feuerverger1977:ECF} has shown that the ECF converges uniformly over compact sets almost surely to the population characteristic function. As discussed in \S\ref{sec:universal-characteristic}, we may view both the kernel mean embedding and the characteristic function as \emph{integral transforms} of the distribution. \citet{Kankainen98:ECF}, for instance, proposes a consistent and asymptotically distribution-free test for independence based on the empirical characteristic function; see also \citet{Kankainen95:ECFIND}. Unfortunately, ECF can be difficult to obtain in many cases, especially for conditional distributions.

\hiddensection{Classical Kernel Machines}

The kernel mean has appeared in the literature since the beginning of the field of kernel methods itself \citep{Scholkopf98:KPCA,Scholkopf01:LKS}. Classical algorithms for classification and anomaly detection employ a mean function in the \gls{rkhs} as their building block. Given a data set $(\x_1,y_1),(\x_2,y_2),\ldots,(\x_n,y_n)$ where $y_i\in\{-1,+1\}$, \citet[Chapter 4]{Shawe04:KMPA}, for example, considers a simple classifier that classifies a data point $\x_*$ by measuring the \gls{rkhs} distance between $\phi(\x_*)$ and the class-conditional means in the feature space
\begin{eqnarray*}
  \muh_{p} &:=& \frac{1}{|\{i\,|\,y_i=+1\}|}\sum_{y_i=+1}\phi(\x_i), \\
  \muh_{n} &:=& \frac{1}{|\{i\,|\,y_i=-1\}|}\sum_{y_i=-1}\phi(\x_i) . 
\end{eqnarray*}
\noindent This algorithm is commonly known as a \emph{Parzen window classifier} \citep{Duda73a}. Likewise, an anomaly detection algorithm can be obtained by constructing a high-confidence region around the kernel mean $\muh := \frac{1}{n}\sum_{i=1}^n\phi(\x_i)$ and considering points outside of this region as outliers. Although the original studies did not provide a link to the embedding of distributions, one can naturally interpret $\muh_{p}$ and $\muh_{n}$ as kernel mean embeddings of conditional distributions $\pp{P}(X|Y=+1)$ and $\pp{P}(X|Y=-1)$, respectively. Furthermore, \citet{Sriperumbudur2009} links the distance between kernel means (its MMD) to empirical risk minimization and the large-margin principle in classification. Lastly, the centering operation commonly used in many kernel algorithms involves an estimation of the mean function in \gls{rkhs} \citep{Scholkopf01:LKS}. As a result, reinterpreting these algorithms from the distribution embedding perspective may shed light on novel methods using these learning algorithms.

\hiddensection{Distance-based Two-Sample Test}

The \emph{energy distance} and \emph{distance covariance} are among important classes of statistics used in two-sample and independence testing that have had a major impact in the statistics community. \citet{Sejdinovic2012,Sejdinovic2013} shows that these statistics are in fact equivalent to the distance between embeddings of distributions with specific choice of kernels. The \emph{energy distance} between probability distributions $\pp{P}$ and $\pp{Q}$ as proposed in \citet{Szekely04:EngDist,Szekely05:EngDist} is given by
\begin{equation}
  \label{eq:energy-distance}
  D_E(\pp{P},\pp{Q}) = 2\ep_{\mathit{XY}}\|X-Y\| - \ep_{XX'}\|X-X'\| - \ep_{YY'}\|Y-Y'\|,
\end{equation} 
\noindent where $X,X'\sim\pp{P}$ and $Y,Y'\sim\pp{Q}$. The distance covariance was later introduced in \citet{Szekely07:DistCov,Szekely09:DistCov} for an independence test as a weighted $L_2$-distance between characteristic functions of the joint and product distributions. The \emph{distance kernel} is a positive definite kernel obtained from $k(\z,\z') = \rho(\z,\z_0) + \rho(\z',\z_0) - \rho(\z,\z')$ where $\rho$ is a semi-metric of negative type.\footnote{A function $\rho$ is said to be semi-metric if the ``distance'' function need not satisfy the triangle inequality. It is of negative type if it is also negative definite (see Definition 2 and 3 in \citealp{Sejdinovic2012}).}  
\citet{Sejdinovic2012} shows that the distance covariance, originally defined with Euclidean distance, can be  generalized with a semi-metric of negative type.  This generalized distance covariance is simply a special case of the HSIC with the corresponding distance kernel.  In this respect, the kernel-based methods provide a more flexible choice for dependence measures. 


\hiddensection{Fourier Optics}
    
\citet{Harmeling2013} establishes a link between Fourier optics and kernel mean embedding from the computer vision viewpoint. A simple imaging system can be described by the so-called \emph{incoherent imaging equation}
\begin{equation}
  \label{eq:imaging-eq}
  q(\mathbf{u}) = \int f(\mathbf{u}-\bm{\xi})p(\bm{\xi})\dd\bm{\xi} ,
\end{equation}
\noindent where both $q(\mathbf{u})$ and $p(\bm{\xi})$ describe image intensities. The function $f$ represents the impulse response function, \ie, the point spread function (PSF), of the imaging system. In this case, the image $p(\bm{\xi})$ induces, up to normalization, a probability measure which represents the light distribution of the object being imaged. The kernel $f(\mathbf{u}-\bm{\xi})$ in \eqref{eq:imaging-eq}, which is shift-invariant, can be interpreted physically as the point response of an optical system. Based on this interpretation, \citet{Harmeling2013} asserts that the Fraunhofer diffraction is in fact a special case of kernel mean embedding and that in theory an object $p(\bm{\xi})$ with bounded support can be recovered completely from its diffraction-limited image, using an argument from the injectivity of the mean embedding \citep{Fukumizu04:DRS,Sriperumbudur08injectivehilbert}. In other words, Fraunhofer diffraction does not destroy any information. A simple approach to compute the inversion in practice is also given in \citet{Harmeling2013}.  

\hiddensection{Probabilistic Perspective}

The kernel mean embedding can also be understood probabilistically. Consider the following example.\footnote{This example was obtained independently via personal communication with Zoubin Gharamani.} Suppose $f\sim\text{GP}(0,k)$ where $k$ is the covariance function of a zero mean Gaussian process, i.e., $\mathbb{E}_f[f(\x)]=0$ and $\mathbb{E}_{f}[f(\x)f(\y)]=k(\x,\y)$. Assume the data generating process $\x\sim\pp{P}$. Then it follows that 
\begin{eqnarray*}
  \ep[f(\x)f(\cdot)] &=& \mathbb{E}_{\x\sim\pp{P}}\mathbb{E}_{f\sim \text{GP}|\x}[f(\x)f(\cdot)] \\ 
  &=& \mathbb{E}_{\x\sim\pp{P}}k(\x,\cdot)=\int k(\x,\cdot)p(\x)\dd\x = \muv_{\pp{P}} .
\end{eqnarray*}
In other words, the kernel mean can be viewed as an expected covariance of the functions induced by the GP prior whose covariance function is $k$. Note that unlike what we have seen so far, the function $f$ is drawn from a GP prior which is almost surely outside  $\hbspace_k$. It turns out that this interpretation coincides with the one given in \citet{Shawe-Taylor07:Density}. Specifically, let $\hbspace$ be a set of functions and $\Pi$ be a probability distribution over $\hbspace$. \citet{Shawe-Taylor07:Density} defines the distance between two distributions $\pp{P}$ and $\pp{Q}$ as
\begin{equation*}
  D(\pp{P},\pp{Q}) := \ep_{f\sim\Pi(f)}\left|\ep_X[f(X)] - \ep_Y[f(Y)]\right|,
\end{equation*}
\noindent where $X\sim\pp{P},Y\sim\pp{Q}$. That is, we compute the average distance between $\pp{P}$ and $\pp{Q}$ \wrt the distribution over \emph{test function class} $\hbspace$ (see also \citealp[Lemma 27, Section 7.5]{Gretton12:KTT} for the connection to MMD). Nevertheless, a fully Bayesian interpretation of the kernel mean embedding remains an open question.
 
%


\chapter{Future Directions} 
\label{sec:future}

We have seen that kernel mean embedding of distributions has made a considerable impact in machine learning and statistics. Nevertheless, some questions remain open and there are numerous potential research directions. In this section, we give suggestions on some of the directions we believe to be important.

\hiddensection{Kernel Choice Problem}
 
How to choose the ``right'' kernel function remains an ultimate problem in kernel methods and there is no exception to the kernel mean embedding. For some applications, one may argue that characteristic kernels should be sufficient as they have been shown theoretically to preserve all necessary information about the distributions. However, this property may not hold empirically because we only have access to finite samples. In which case, prior knowledge about the problem becomes more relevant in choosing the ``right'' kernel. It is also not trivial how to choose parameter values of such a kernel. These issues have been addressed in \citet{Sriperumbudur2009,Gretton12:SSSBPF}, for example. Finally, understanding how to interpret the associated representation of distributions is essential for several applications in statistics. 
 
\hiddensection{Stochastic Processes and Bayesian Interpretation}

A probability distribution is often defined over finite dimensional objects. Highly structured data can be taken into account using a kernel, but it is not clear how to employ kernel mean embedding for \emph{stochastic processes}---\ie, distributions over infinite dimensional objects---in a similar manner to the distribution. See \citet{Chwialkowski2014} and \citet{ChwialkowskiSG14:WILD} for some preliminary results. This line of research has been quite fruitful in Bayesian nonparametrics. Hence, extending the kernel mean embedding to a Bayesian nonparametric framework provides potential applications of probabilistic inference. It is also interesting to obtain a Bayesian interpretation of the kernel mean embedding. Having an elegant interpretation could potentially lead to several extensions of the previous studies along the line of Bayesian inference, \eg, Gaussian processes.

\hiddensection{Scalability}
In the era of ``big data'', it is imperative that modern learning algorithms are able to deal with increasingly complex and large-scale data. Recently, there has been a growing interest in developing large-scale kernel learning, which is probably inspired by the lack of theoretical insight of deep neural networks despite their success in various application domains. The advances along this direction will benefit the development of algorithms using kernel mean embedding. In the context of MMD, many recent studies have addressed this issue \citep{Zaremba2013,Cortes2014,Ji15:FastMMD, Kacper15:Analytic}.

\hiddensection{High-dimensional Inference} 

It has been observed that the improvement of a shrinkage estimator tends to increase as the ambient dimensionality increases. Kernel-based methods are known to be less prone to the \emph{curse of dimensionality} compared to classical approaches such as kernel density estimation, but little is known about the underlying theory. Better understanding of the role of ambient dimension in kernel methods may shed light in novel applications of kernel mean embedding in high-dimensional regime. There are some recent studies on high-dimensional analysis of MMD and related tests: see \citet{Sashank15:HighDim,Ramdas15:AdaptiveKME,Ramdas15:DPK} for example.

\hiddensection{Causality}
  

  Causal inference involves the investigation of how the distribution of outcome changes as we change some other variable. Several frameworks for causal inference using kernel mean embedding have been proposed. There have been some recent studies in this direction \citep{Zhang2011,Sgouritsa2013,Chen2014,Lopez-Paz15:Towards}, but it remains a challenging problem. For example, \citet{Lopez-Paz15:Towards} considers bivariate causal inference as a classification task on the joint distributions of cause and effect variables. In the potential outcome framework, the causal effect is defined as the difference between the distributions of outcomes under \emph{control} and \emph{treatment} populations. Due to the \emph{fundamental problem of causal inference}, either one of them would never be observed in practice. In this case, the question is whether we can use the kernel mean embedding to represent the counterfactual distribution.


\hiddensection{Privacy-Preserving Embedding}

Privacy is an essential requirement in medicine and finance---especially in the age of big data---because we now have access to an unprecedented amount of data of individuals that could compromise their privacy. Unlike most parametric models, most applications of kernel mean embedding such as two sample testing and dependency measures often require direct access to individual data points which renders these algorithms vulnerable to adversarial attack. As a result, it is imperative to ask to what extent we can still achieve the same level of performance under an additional requirement that the privacy must be preserved. Another challenge in the context of kernel mean embedding is to preserve the privacy at the distributional level.

\hiddensection{Statistics, Machine Learning, and Invariant Distribution Embeddings}

Fundamental problems in statistics such as parameter estimation, hypothesis testing, and testing for model families, can be viewed as an estimation of the functional, \ie, the estimator, which maps the empirical distribution to the value of the estimate. The classical approach to constructing such an estimator is to choose estimators which satisfy certain criteria, \eg, maximum likelihood estimators. On the other hand, there have been recent studies that look at these problems from the machine learning perspective. That is, it is assumed that we have access to the training set $\{(\widehat{\pp{P}}_1(X),l_1),(\widehat{\pp{P}}_2(X),l_2),\ldots,(\widehat{\pp{P}}_n(X),l_n)\}$ where $\widehat{\pp{P}}_1(X)$ denotes an empirical distribution over a random variable $X$ and $l_i$ is the desired value of the estimate. For example, $l_i$ may be the entropy computed from $\widehat{\pp{P}}$. Thus, finding an estimator in this case can be cast as a supervised learning problem on probability distributions. Using the kernel mean embedding of $\widehat{\pp{P}}$, \citet{Zoltan15:DistReg} provides theoretical analysis of the distribution regression framework and demonstrates the framework on entropy estimation problem. \citet{Flaxman15:DistReg} employs distribution regression in ecological inference. Lastly, \citet{Lopez-Paz15:Towards} casts a bivariate causal inference problem as a classification problem on distributions. One of the future directions is to characterize the \emph{learned} estimators and to understand their connection to most of the classical estimators in statistics. 

Many statistical properties of a probability distribution are independent of the input space on which it is defined. For example, independence implies $p(\x,\y)=p(\x)p(\y)$ regardless of $\inx$ and $\iny$. Therefore, there is a need to develop an invariant representation for distributions which will allow us to deal with such distributions simultaneously across different domains. This kind of knowledge is known as \emph{domain-general knowledge} in cognitive science \citep{Goodman11:ThyCau}.


\chapter{Conclusions} 
\label{sec:conclusions} 

A kernel mean embedding of distributions has emerged as a powerful tool for probabilistic modeling, statistical inference, machine learning, and causal discovery. The basic idea is to map distributions into a reproducing kernel Hilbert space (RKHS) where the whole arsenal of kernel methods can be extended to probability measures. It has given rise to a great deal of research and novel applications of positive definite kernels in both statistics and machine learning. In this survey, we gave a comprehensive review of existing work and recent advances in this research area. In the course of the survey, we also discussed some of the most challenging issues and open problems that could potentially lead to new research directions.

The survey began with a brief introduction to the reproducing kernel
Hilbert space (RKHS) which forms the backbone of this survey (Section \ref{sec:background}). As the
name suggests, the kernel mean embedding owes its success to the
powerful concept of a positive definite function commonly known as the 
\emph{kernel function}. We provided a short review of its
theoretical properties and classical applications. Then, we
reviewed the Hilbert-Schmidt operators which are vital ingredients in modern applications of kernel mean embedding.

Next, we provided a thorough discussion of the Hilbert space embedding
of marginal distributions, theoretical guarantees, and a review of its
applications (Section \ref{sec:marginal-embedding}). To summarize, the Hilbert space embedding of distributions allows us to apply RKHS methods to probability measures. This extension prompts a wide range of applications such as kernel two-sample testing, independence testing, group anomaly detection, MCMC methods, predictive learning on distributions, and causal inference.
 
The survey then generalized the idea of embedding to conditional
distributions, gave theoretical insights, and reviewed some
applications (Section \ref{sec:conditional-embedding}). The
conditional mean embedding allows us to perform sum, product, and
Bayes' rules which are ubiquitous in graphical models, probabilistic
inference, and reinforcement learning, for example, in a 
non-parametric way. Furthermore, the conditional mean embedding admits
a natural interpretation as a solution to a vector-valued regression
problem. Then, we mentioned some relationships between the kernel mean 
embedding and other related areas (Section
\ref{sec:connections}). Lastly, we gave some suggestions for future
research directions (Section \ref{sec:future}). 
 
We hope that this survey will become a useful reference for graduate
students and researchers in machine learning and statistics who are
interested in the theory and applications of the kernel mean embedding of distributions.
  
\subsubsection*{Acknowledgement} 

KM would like to thank Anant Raj for reading the first draft of this survey and providing several helpful comments. We thank Wittawat Jitkrittum, Sebastian Nowozin, and Dino Sejdinovic for detailed and thoughtful comments about the final draft. We also thank Motonobu Kanagawa for pointing out the mistake in one of his paper, and gave detailed explanation of the mistake, which we have included in the survey. Last but not least, we are in debt to the editor and anonymous reviewers who provided insightful reviews that helped to improve the survey significantly.
 
\bibliographystyle{abbrvnat}
\bibliography{kme-review} 

\end{document}